\documentclass[12pt]{article}
\usepackage{amsmath}
\usepackage{graphicx,psfrag,epsf}
\usepackage{enumerate}
\usepackage{natbib}

\newcommand{\blind}{1}

\usepackage[margin=1in]{geometry}

\usepackage{authblk, subcaption}
\usepackage{colortbl}
\usepackage{hhline}
\usepackage{calc}
\usepackage{tabularx}
\usepackage{amsmath, bm, mathtools, mathrsfs, amsthm}
\usepackage{comment}
\usepackage{enumitem}
\usepackage{xcolor}
\newtheorem{theorem}{Theorem}

\newtheorem{lemma}{Lemma}%
\newtheorem{corollary}{Corollary}%

\graphicspath{{sim_figures/}} 
\allowdisplaybreaks
\usepackage{amsfonts}

\newcommand{\I}{\mathbb{I}}

\newcommand{\E}{\mathbb{E}}

\newcommand{\M}{\mathcal{M}}

\newcommand{\calL}{\mathcal{L}}

\DeclareMathOperator*{\argmin}{\arg\!\min}

\newcommand{\hilbert}{\hbox{${\rm I\kern-.2em H}$}}
\newcommand{\tangent}{\mathbb{T}}
\newcommand{\openr}{\mathbb{R}}
\newcommand{\openn}{\hbox{${\rm I\kern-.2em N}$}}
\newcommand{\opend}{\hbox{${\rm I\kern-.2em D}$}}
\newcommand{\smallO}{o_P}

\definecolor{babyblue}{rgb}{0.54, 0.81, 0.94}
\newcolumntype{C}[1]{>{ \arraybackslash}p{#1}} 

\usepackage{hyperref}
\usepackage{xr-hyper}
\makeatletter
\newcommand*{\addFileDependency}[1]{
  \typeout{(#1)}
  \@addtofilelist{#1}
  \IfFileExists{#1}{}{\typeout{No file #1.}}
}
\makeatother
\newcommand*{\myexternaldocument}[1]{%
    \externaldocument{#1}%
    \addFileDependency{#1.tex}%
    \addFileDependency{#1.aux}%
}
\myexternaldocument{supp}

\begin{document}

\def\spacingset#1{\renewcommand{\baselinestretch}%
{#1}\small\normalsize} \spacingset{1}


\if1\blind
{
  \title{\bf 
  Statistical learning for constrained functional parameters in infinite-dimensional models 
  }
  
    
  \author[1]{Razieh Nabi}
  \author[2]{Nima S. Hejazi}
  \author[3]{Mark J. van der Laan}
  \author[1]{David Benkeser}
  
  \affil[1]{Department of Biostatistics and Bioinformatics, Rollins School of Public Health, Emory University, Atlanta, GA, USA}
  \affil[2]{Department of Biostatistics, T.H. Chan School of Public Health, Harvard University, Boston, MA, USA}
  \affil[3]{Division of Biostatistics, School of Public Health, University of California, Berkeley, Berkeley, CA, USA}
  
  \date{}
  \maketitle
} \fi

\if0\blind
{
  \bigskip
  \bigskip
  \bigskip
  \begin{center}
    {\LARGE\bf 
    Statistical learning for constrained functional parameters in infinite-dimensional models
    }
\end{center}
  \medskip
} \fi

\begin{abstract} 
We develop a general framework for estimating function-valued parameters under equality or inequality constraints in infinite-dimensional statistical models. Such constrained learning problems are common across many areas of statistics and machine learning, where estimated parameters must satisfy structural requirements such as moment restrictions, policy benchmarks, calibration criteria, or fairness considerations. To address these problems, we characterize the solution as the minimizer of a penalized population risk using a Lagrange-type formulation, and analyze it through a statistical functional lens. Central to our approach is a \textit{constraint-specific path} through the unconstrained parameter space that defines the constrained solutions. For a broad class of constraint-risk pairs, this path admits closed-form expressions and reveals how constraints shape optimal adjustments. When closed forms are unavailable, we derive recursive representations that support tractable estimation. Our results also suggest natural estimators of the constrained parameter, constructed by combining estimates of unconstrained components of the data-generating distribution. Thus, our procedure can be integrated with any statistical learning approach and implemented using standard software. We provide general conditions under which the resulting estimators achieve optimal risk and constraint satisfaction, and we demonstrate the flexibility and effectiveness of the proposed method through various examples, simulations, and real-data applications.

\end{abstract}

\noindent%
\vfill

\newpage
\spacingset{1.9} 

\section{Introduction} 
\label{sec:intro}

Constrained statistical learning involves estimating parameters subject to structural restrictions, often motivated by domain-specific requirements that ensure compatibility with known properties of the system being modeled. Such constraints arise across a wide range of applications and reflect diverse goals. In some settings, constraints help preserve the external validity of causal findings, for example, by aligning model outputs with known effects from prior trials or enforcing policy benchmarks in intervention analysis \citep{dehejia1999causal, cole2010generalizing, kitagawa2018should, athey2021policy, nabi2019learning, parikh2022validating, liu2024learning, sun2024empirical}. In other cases, they aim to reduce disparities in predictive performance across subpopulations, as in algorithmic fairness frameworks that impose group-level constraints on model behavior \citep{mitchell2018prediction, plecko2022causal, barocas2023fairness}. Elsewhere, constraints serve to satisfy operational or regulatory criteria, such as moment restrictions in density estimation \citep{newey199316, hall1999density, chernozhukov2023constrained} or bias correction under model misspecification \citep{scharfstein1999adjusting, van2006targeted, vdl2011targeted}. 
Despite these differences in motivation, such problems share a common structure: they require estimating a function-valued parameter that satisfies one or more equality or inequality constraints derived from the data-generating distribution.

A variety of methods have been developed for incorporating constraints into statistical estimation procedures. In classical parametric settings, constrained estimation has a long history, including maximum likelihood under linear or nonlinear constraints, the use of Lagrange multipliers to enforce equality or inequality constraints, and constrained versions of the generalized method of moments for incorporating moment restrictions motivated by theory or policy, particularly in econometrics \citep{aitchison1958maximum, geyer1994asymptotics, gourieroux1995statistics, McFadden1999GMM, chernozhukov2023constrained}. While many of these classical approaches operate in finite-dimensional or parametric models, modern applications increasingly call for solutions in richer, nonparametric or infinite-dimensional settings \citep{bickel1993efficient, tsiatis2007semiparametric, vdl2011targeted}. This nonparametric regime presents distinct theoretical and algorithmic challenges, particularly when the constraint depends on the unknown data-generating distribution. 

In modern statistics and machine learning, constraints are often implemented at the algorithmic level, where validation, fairness, robustness, or performance-related constraints are enforced through penalty terms, regularizers, or feasibility restrictions within a selected optimization objective \citep{donini2018empirical, zafar2019fairness, agarwal2019fair}. A broad literature on algorithmic fairness distinguishes between pre-processing the data \citep{kamiran2012data, zemel2013learning, calmon2017optimized}, in-processing via constrained or regularized objectives \citep{zafar2019fairness, donini2018empirical}, and post-processing model outputs to satisfy fairness criteria \citep{hardt2016equality, woodworth2017learning, wang2023adjusting}. More recent work has also explored influence-function-based corrections for constrained optimization problems in causal and reinforcement learning \citep{jordan2022data}. While effective in their respective domains, these approaches are typically tailored to specific estimands, constraint types, or finite-dimensional parameterizations. Many of these methods are built around particular statistical models, often with a focus on classification, and are not readily adaptable to a broad range of constraints or learning algorithms. As a result, there remains a lack of general-purpose methodology for estimating function-valued parameters under equality or inequality constraints, particularly in infinite-dimensional nonparametric models where the constraint may itself depend on the unknown data-generating distribution. 

This paper addresses the gap identified above by developing a general framework for learning constrained functional parameters in infinite-dimensional statistical models. The target parameter is a function-valued object, such as a conditional expectation or probability density, defined as the minimizer of a population risk subject to one or more equality or inequality constraints. We characterize the solution by establishing Lagrange-type methods for infinite-dimensional statistical functionals, and we leverage tools from functional analysis to derive representations of the optimal constrained parameter. In many settings, these representations admit closed-form expressions involving the unconstrained solution and other components of the data-generating distribution. This enables plug-in estimation using standard learning procedures applied to unconstrained quantities, making the approach compatible with modern machine learning pipelines and implementable without custom solvers or parametric assumptions. Related but distinct work by \citet{chamon2020probably, chamon2022constrained} considers empirical Lagrangian duality in this context. 

The major contributions of this work are as follows:
(i) \textit{A general framework for constrained estimation in infinite-dimensional models with pathwise differentiable constraints.} We introduce estimators of function-valued parameters subject to one or more equality or inequality constraints, formulated via a Lagrange-type penalization. Our methodology leverages tools from functional analysis to yield general characterizations applicable across a broad class of constrained learning problems.
(ii) \textit{Closed-form and recursive solutions for a broad class of constraints.} We identify a rich set of constraint-function and risk-function pairs for which our framework yields closed-form solutions for the optimal constrained parameter. When closed forms are unavailable, we derive recursive characterizations that support plug-in estimation via standard learners and grid search over constraint paths. These results include detailed examples involving causal effect constraints, risk calibration, and moment restrictions.
(iii) \textit{Model-agnostic, theoretically justified plug-in estimation procedures.} Our framework enables estimation using standard learning tools applied to unconstrained nuisance parameters, without requiring custom solvers or parametric assumptions. We provide conditions under which the resulting estimators achieve optimal penalized risk and constraint satisfaction, and demonstrate the practical utility of our approach through simulations and real-data applications.

\section{Statistical estimation problem}
\label{sec:problem}

\textbf{\textit{Unconstrained functional parameter.}}  
We observe $n$ independent copies $O_1, \ldots, O_n$ of a random variable $O \sim P_0$, where $P_0$ is a member of a large, possibly unrestricted (infinite-dimensional), statistical model $\M$. In all of our examples, we let $\M$ be the nonparametric model, the set of all potential probability distributions compatible with $O$; however, our approach generalizes equally well to other infinite-dimensional semiparametric models. We use $P$ to denote an arbitrary distribution in $\M$.

Let $\Psi: \mathcal{M} \rightarrow {\bf \Psi}$ be the \emph{unconstrained} function-valued parameter of interest. This parameter maps a given distribution $P \in \mathcal{M}$ of $O$ to a parameter space ${\bf \Psi}$. We assume that $\Psi(P)$ can be defined as the minimizer of an expectation of a loss function:
\begin{align}
  \Psi(P) = \argmin_{\psi \in {\bf \Psi}} \ \int L(\psi)(o) \ dP(o) \ .
  \label{eq:target_unconstrained}
\end{align}%
We introduce the shorthand notation $R_P(\psi) = \int L(\psi)(o) \ \! dP(o)$ to describe the \emph{risk} of a given $\psi \in {\bf \Psi}$ under sampling from $P$. We will adopt the notational convention that $\psi = \Psi(P)$ for an arbitrary distribution in $\mathcal{M}$, while $\psi_0 = \Psi(P_0)$.

\textbf{\textit{Constrained functional parameter.}} 
We assume the constraint can be expressed as a known functional of the data-generating distribution. To introduce the core ideas, we begin with the case of a single equality constraint. Specifically, for $P \in \mathcal{M}$, we write the constraint as $\Theta_{P}: {\bf \Psi} \rightarrow \openr$ and without loss of generality assume a constraint of the form $\Theta_{P}(\psi) = 0$. We assume the constraint is feasible, meaning the set $\{\psi \in \mathbf{\Psi} : \Theta_P(\psi) = 0\}$ is non-empty under $P$, ensuring the constraint is meaningful for the problem at hand. 

We define a \emph{constrained target parameter} as any mapping $\Psi^*: \mathcal{M} \rightarrow {\bf \Psi}$, that satisfies:
\begin{align}
  \Psi^*(P) = \argmin_{\psi \in {\bf \Psi}, \ \Theta_{P}(\psi) = 0}  \
    R_P(\psi) \ . 
  \label{eq:target}
\end{align}
We adopt the notational convention that $\psi^* = \Psi^*(P)$ for an arbitrary distribution in $\mathcal{M}$, while $\psi^*_0 = \Psi^*(P_0)$. Our aim is to develop a framework for characterizing $\psi^*_0$ and developing estimators thereof. We first focus on a single equality constraint, then extend the framework to handle inequality constraints. Additional generalizations to multiple constraints are presented in the supplementary material. 

We make the simplifying assumption that $\psi_0$ is a conditional mean or probability, as is common in many statistical learning problems. Specifically, we assume $O = (Z, Y)$ and that $\psi_0^*$ is a function of the predictor vector $Z$ used to predict the outcome $Y$. We denote by $P_{0,Z}$ the marginal distribution of $Z$ under $P_0$. Extensions to other functionals, such as conditional densities, are discussed in the appendix.

\section{Methods}\label{sec:method}

\subsection{Theoretical foundations}

Characterization and estimation of constrained target parameters with known constraints over finite-dimensional parameter spaces is a well-studied problem in optimization theory. Solutions have been proposed using Lagrange-type methods and linear programming among other approaches. The current setting is more challenging since (i) we are optimizing over an infinite-dimensional function space; and (ii) the constraint is an unknown functional of the data generating distribution. Thus, we require new theory to establish appropriate methods for the analysis of our problem. Here we introduce some of the fundamental ideas from functional analysis that we use to establish this theory.

For any $\psi \in {\bf \Psi}$, we introduce the idea of a \emph{path through} $\psi$ that has a particular \emph{direction} $h$, where $h$ is a function of $Z$. The \emph{path} is denoted by $\{\psi_{\delta, h} : \delta \in \openr\} \subset {\bf \Psi}$ and is indexed by the univariate parameter $\delta$. We say that this path is \emph{through} $\psi$ if $\psi_{\delta,h}|_{\delta = 0} = \psi$. The \emph{direction} of the path, $h$, is defined as $\frac{d}{d\delta} \psi_{\delta,h} |_{\delta = 0}$. We let the directions of the paths vary over a set $\mathcal{H}_{P}(\psi)$ chosen such that it captures all possible local directions within the parameter space ${\bf \Psi}$ in which one can move away from $\psi$. We embed $\mathcal{H}_{P}(\psi)$ in a Hilbert space $\hilbert_P(\psi)$ endowed with a covariance inner product of the form $\langle f,g \rangle = \int f(z) g(z) dP_{Z}(z)$. In our examples, we take $\hilbert_{P}(\psi) = L^2(P_{Z})$, the space of all bounded functions $h$ of $Z$ such that $\int h(z)^2 dP_{Z}(z) < \infty$. Let $\tangent_P(\psi) \subseteq \hilbert_P(\psi)$ be the closure in $\hilbert_P(\psi)$ of the linear span of $\mathcal{H}_P(\psi)$. 

By embedding $\mathcal{H}_P(\psi)$ in $\hilbert_P(\psi)$, we can derive an inner-product characterization of \emph{bounded linear functionals} on $\tangent_P(\psi)$ of the form $f_P(\psi_{\delta,\cdot}): \tangent_P(\psi) \rightarrow \mathbb{R}$. We recall that the function $h \mapsto f_P(\psi_{\delta, h})$ for $h \in \tangent_P(\psi)$, is called \emph{linear} on $\tangent_P(\psi)$ if it is such that $f_P(\psi_{\delta, \alpha h_1 +\beta  h_2}) = \alpha f_P(\psi_{\delta, h_1}) + \beta f_P(\psi_{\delta, h_2})$ for all $h_1, h_2 \in \tangent_P(\psi)$ and for all $\alpha, \beta \in \mathbb{R}.$ Further, we say that the function is \emph{bounded} if there exists an $M > 0$ such that $|f_P(\psi_{\delta, h})| \le M \langle h, h \rangle^{1/2}$ for all $h \in \tangent_P(\psi)$. Finally, we recall that a linear functional on a Hilbert space is bounded if and only if it is continuous.

By the Reisz representation theorem, any bounded linear functional on $\tangent_P(\psi)$ will have an inner product representation $f_P(\psi_{\delta,h}) = \langle h, D_P(\psi) \rangle$ for some unique element $D_P(\psi) \in \tangent_P(\psi)$. The object $D_P(\psi)$ is sometimes referred to as the \emph{canonical gradient} of the functional.

\subsection{Assumptions}
\label{subsec:assumptions}

In developing our methodology, we consider derivatives of the risk $R_P(\psi)$ and the constraint
$\Theta_{P}(\psi)$ along paths $\{ \psi_{\delta, h} : \delta \in \mathbb{R} \}$ for $h \in \tangent_P(\psi)$. We can view these derivatives (at $\delta = 0$) as a functional $f_P$ on $\tangent_P(\psi)$ of the form $h \mapsto \frac{d}{d\delta} f_P(\psi_{\delta, h})|_{\delta = 0}$. 

Our key assumption is that the pathwise derivatives of $R_P$ and $\Theta_P$ represent bounded (or equivalently, continuous) linear functionals on $\tangent_P(\psi)$ \citep{van1995efficient}. Specifically, we assume for every $h \in \tangent_P(\psi)$ and for every univariate path through $\psi$ with direction $h$,

\begin{equation}\tag{A1}
  \label{assmp:pathwise_diff}
\begin{aligned} 
  &h \mapsto \frac{d}{d\delta} R_P(\psi_{\delta,h})\Big|_{\delta = 0} \
  \mbox{ and } \ h \mapsto \frac{d}{d\delta}
  \Theta_P(\psi_{\delta,h})\Big|_{\delta = 0}  \mbox{are bounded linear
  functionals on $\tangent_P(\psi)$} \ . 
\end{aligned}
\end{equation}
Under \eqref{assmp:pathwise_diff}, the pathwise derivatives will have an inner-product representation,
\begin{align*}
  \frac{d}{d\delta} R_P(\psi_{\delta,h})\Big|_{\delta = 0} = \langle
  D_{R,P}(\psi), h \rangle \quad \mbox{and} \quad \frac{d}{d\delta}
  \Theta_P(\psi_{\delta, h}) \Big|_{\delta = 0} = \langle D_{\Theta,P}(\psi),
  h \rangle \ ,
\end{align*}
for some unique elements $D_{R,P}$ and $D_{\Theta,P}$ of the tangent space $\tangent_P(\psi)$. See Appendix~\ref{app:pathwise_differentiability} for a remark on pathwise differentiability and links to non/semiparametric efficiency theory. We also assume $D_{\Theta,P}$ is non-degenerate at a non-trivial solution $\psi_0^*$, i.e., for a $\psi_0^* \ne \psi_0$:
\begin{align}
  P_0 \{ D_{\Theta,P}(\psi_0^*)(O) = 0\} = 0  \ . \tag{A2}
  \label{assmp:licq}
\end{align}
Assumption \eqref{assmp:licq} is sufficient but not necessary for our developments and corresponds to the simplest form of the linear independence constraint qualification in our setting \citep{gould1971necessary,sundaram1996first,boltyanski1998geometric}.

\subsection{Defining and characterizing constraint-specific path}
\label{subsec:constraint-specific-path}

We first show that under \eqref{assmp:pathwise_diff} the constrained functional parameter $\Psi^*(P)$, defined via \eqref{eq:target}, can be characterized using Lagrange multipliers; see Appendix~\ref{app:proofs} for a proof.
\begin{lemma} Consider a vector-valued parameter $\Phi(P): \mathcal{M} \rightarrow {\bf \Psi} \times \openr$, defined as
\begin{align}
	\Phi(P) = \argmin_{\psi \in {\bf \Psi}, \ \lambda \in
		\openr} \ R_P(\psi) + \lambda \Theta_{P}(\psi) \ .
  \label{eq:lagrangedef}
\end{align}
We note that $\Phi$ maps a given distribution $P$ into a vector-valued parameter $(\tilde{\Psi}(P), \Lambda(P))$, where $\tilde{\Psi}(P) \in {\bf \Psi}$ and $\Lambda(P) \in \openr$. Under \eqref{assmp:pathwise_diff} and \eqref{assmp:licq}, for each $P \in \mathcal{M}$, $\Psi^*(P) = \tilde{\Psi}(P)$. \label{lem:lossmin}
\end{lemma}
For a given $\lambda$, we define $\psi_{0, \lambda} = \argmin_{\psi \in {\bf \Psi}} R_{P_0}(\psi) + \lambda \Theta_{P_0}(\psi)$ as the minimizer of penalized risk under sampling from $P_0$. Lemma~\ref{lem:lossmin} establishes that $\psi^*_0 = \psi_{0, \lambda_0}$, where $\lambda_0 = \Lambda(P_0)$. We note that $\{\psi_{0, \lambda} : \lambda \in \openr\} \subset {\bf \Psi}$ represents a \emph{path} through $\psi_0$. We refer to this path as the \textit{constraint-specific path}, noting that both $\psi_0$ and $\psi_{0}^*$ are part of this path. For each $\psi_{0, \lambda}$ along the constraint-specific path and for $\lambda \in \openr$, $\psi_{0, \lambda}$ minimizes the criterion $
  \psi \mapsto R_{P_0}(\psi) + \lambda \Theta_{P_0}(\psi)$. 
Thus, a derivative of this criterion along a path $\{\psi_{0, \lambda, \delta} : \delta \in \openr\}$ through $\psi_{0, \lambda}$ at $\delta = 0$ equals zero and for every $h \in \tangent_{P_0}(\psi)$, 
\begin{align*}
  0 &= \frac{d}{d \delta}R_{P_0}(\psi_{0, \lambda, \delta, h}) \Big
      |_{\delta = 0} + \lambda \frac{d}{d \delta} \Theta_{P_0}
      (\psi_{0, \lambda, \delta, h}) \Big|_{\delta = 0} \\
    &= \langle D_{R, P_0}(\psi_{0, \lambda}), h \rangle + \lambda \big\langle
      D_{\Theta,P_0}(\psi_{0, \lambda}), h  \big\rangle \\
    &= \langle D_{R, P_0}(\psi_{0, \lambda}) + \lambda D_{\Theta,P_0}
      (\psi_{0,\lambda}), h \rangle \ ,
\end{align*}%
where the second line follows from \eqref{assmp:pathwise_diff} and the last line follows because $\tangent_{P_0}(\psi)$ is a Hilbert space, which is by definition an inner product space. Since inner products are non-degenerate, the constraint-specific path $\{\psi_{0, \lambda}: \lambda \in \openr\}$ must satisfy
\begin{align}
  D_{R, P_0}(\psi_{0, \lambda})(o) + \lambda D_{\Theta, P_0}
    (\psi_{0, \lambda})(o) = 0 \ , \ \text{for all } \lambda \in \openr \text{ and } o \text{ in a support of } P_0 \ . 
\label{eq:lfmpathcondition}
\end{align}%
Thus, \eqref{eq:lfmpathcondition} and the constraint equation $\Theta_{P_0}(\psi_{0,\lambda_0}) = 0$ provide two equations with which we can attempt to solve for $\psi_{0,\lambda}$ and $\lambda_0$. We show that closed-form solutions for $\psi_{0,\lambda}$ are available for many constraints and risk criteria; see Section~\ref{sec:general_constraint}. These expressions are ultimately expressed in terms of parameters of $P_0$, thereby suggesting natural estimators; see Section~\ref{subsec:estimators}. Closed-form solutions for $\lambda_0$ are also often available. However, even in settings where no closed-form solution for $\lambda_0$ can be found, we show that we are often able to propose practical approaches to estimation via grid search procedures.

When a closed-form solution for $\psi_{0, \lambda}$ is not available, it is useful to develop an alternative characterization of the constraint-specific path that allows for a recursive estimation strategy to be used. To that end we define $D_{0, R + \lambda \Theta}(\psi) = D_{R, P_0}(\psi) + \lambda D_{\Theta, P_0}(\psi)$. Given (\ref{eq:lfmpathcondition}), we know that for a given datum $o$, $D_{0, R + \lambda \Theta}(\psi_{0, \lambda})(o) = 0$ for all $\lambda$, and thus  $\frac{d}{d \lambda} D_{0, R + \lambda \Theta}(\psi_{0, \lambda})(o) = 0$. The chain rule implies  
\begin{align}
	0 &= \frac{d}{d \psi_{0, \lambda}}  D_{R, P_0}(\psi_{0, \lambda})  \times 
	\frac{d}{d \lambda} \psi_{0, \lambda} + \lambda \times 	\frac{d}{d \psi_{0, \lambda}}  D_{\Theta, P_0}(\psi_{0, \lambda})  \times 
	\frac{d}{d \lambda} \psi_{0, \lambda}   + D_{\Theta, P_0}(\psi_{0, \lambda}) \nonumber \\
	&= \frac{d}{d \psi_{0, \lambda}}  D_{0, R + \lambda \Theta}(\psi_{0, \lambda})  \times 
	 \frac{d}{d \lambda} \psi_{0, \lambda} + D_{\Theta, P_0}(\psi_{0, \lambda}) \ .  
	 \label{eq:dir-deriv}
\end{align}
We define the operator $\dot{D}_{0, R+\lambda\Theta}(\psi)(h) = \frac{d}{d \psi}  D_{0, R + \lambda \Theta}(\psi)(h)$. Assuming $\dot{D}_{0, R + \lambda \Theta}(\psi)$ is invertible (see Appendix~\ref{app:invertible_operators} for details), \eqref{eq:dir-deriv} implies
\begin{align}
	\frac{d}{d \lambda} \psi_{0, \lambda} = - \big\{ \dot{D}_{0, R + \lambda \Theta}
	(\psi_{0, \lambda}) \big\}^{-1} D_{\Theta, P_0}(\psi_{0, \lambda}) \ .
	\label{differentialeqn}
\end{align}
We will see in Section~\ref{subsec:estimators} that this representation allows for recursively constructed estimators of $\psi_{0,\lambda}$, while $\lambda_0$ can be determined via grid search, as mentioned above.

We summarize our results on the constraint-specific path in the following theorem. 
\begin{theorem}	\label{thm:path-characterization}
	Let $D_{R, P_0}(\psi)$ and $D_{\Theta, P_0}(\psi) \in \tangent_{P_0}(\psi)$ be the canonical gradients of pathwise derivatives of the risk function, $\left . \frac{d}{d \delta}	R_{P_0}(\psi_{\delta, h}) \right |_{\delta = 0}$, and the constraint, $\left . \frac{d}{d \delta}\Theta_{P_0}(\psi_{\delta, h})
	\right |_{\delta = 0} $, respectively. 
	Let  $\{\psi_{0, \lambda}  = \argmin_{\psi \in {\bf \Psi}} R_{P_0}(\psi) + \lambda	\Theta_{P_0}(\psi) : \lambda \in \openr\}$ define a constraint-specific path through $\psi_0$ at $\lambda = 0$. Under (\ref{assmp:pathwise_diff}) and (\ref{assmp:licq}), the following two statements are true: 
	\begin{enumerate}[label=(\roman*)]
        \setlength{\itemindent}{0.25cm}
        \setlength{\itemsep}{0.1cm}
        
	    \item \ The constraint-specific path satisfies: 
    	\begin{align}
    		D_{R,P_0}(\psi_{0, \lambda}) + \lambda D_{\Theta, P_0}(\psi_{0, \lambda}) = 0 \ , \ \ \forall \lambda \in \openr \ . 
            \label{eq:pathcondition-a}
    	\end{align}%
    	For a given $\lambda$, the above characterizes a solution to the minimization problem\\ $\argmin_{\psi \in \bm{\Psi}} R_{P_0}(\psi) + \lambda \Theta_{P_0}(\psi)$ as a function of $P_0$ and $\lambda$. 

        \item \ Let $\dot{D}_{0, R+\lambda\Theta}(\psi) = \frac{d}{d \psi}  [D_{R, P_0}(\psi) + \lambda D_{\Theta, P_0}(\psi)]$ and assume it is invertible. Then, the following is an equivalent characterization of the constraint-specific path: 
    	\begin{align}
    		\frac{d}{d \lambda}\psi_{0, \lambda} = - \big\{ \dot{D}_{0, R + \lambda
    			\Theta}(\psi_{0, \lambda})\big\}^{-1} D_{ \Theta, P_0}(\psi_{0, \lambda}) \ . 
            \label{eq:pathcondition_b}
    	\end{align}%
	\end{enumerate}
\end{theorem}

The constraint-specific path can be considered as establishing a ``first-order'' condition for solutions to the constrained optimization problem. Additional ``second-order'' conditions are required to establish whether the solution constitutes a minimum. Discussions regarding these conditions are  deferred to Appendix~\ref{app:second-order}.

\subsection{Estimation of constrained functional parameter}
\label{subsec:estimators}

In general, solutions to \eqref{eq:pathcondition-a} can be expressed as $\psi_{0,\lambda} = r(\eta_0, \lambda)$ for some mapping $r$, where $\eta_0 = \eta(P_0)$ denotes a collection of real- or function-valued nuisance parameters of $P_0$. These nuisances typically index the gradients  $D_{R, P_0}$ and $D_{\Theta, P_0}$, and almost always include $\psi_0$, the unconstrained risk minimizer. The specific nuisance components beyond $\psi_0$ depend on the structure of the problem. This form suggests a natural plug-in estimator for any $\psi_{0,\lambda}$ along the constraint-specific path. 

Let $\psi_{n,\lambda} = r(\eta_n, \lambda)$, where $\eta_n$ is an estimate of $\eta_0$. In some scenarios, there is also an explicit solution for $\lambda_0 = \ell(\eta_0)$ for some mapping $\ell$. This again suggests a natural plug-in estimator $\lambda_n = \ell(\eta_n)$. Thus, an estimate of $\psi_{0,\lambda_0}$ is given by $\psi_{n,\lambda_n}$. If no explicit solution for $\lambda_0$ is available then it is possible to use empirical minimization to estimate $\lambda_0$. Let $\Theta_n(\psi)$ denote an estimate of $\Theta_{P_0}(\psi)$ for any $\psi \in {\bf \Psi}$ and define $\lambda_n = \mbox{argmin}_{\lambda \in [\delta_1, \delta_2]} |\Theta_n(\psi_{n,\lambda})|$, which can be computed using a grid search. See Section~\ref{subsec:examples} for an example, including details of how to appropriately choose $\delta_1$ and $\delta_2$.

On the other hand, if \eqref{eq:pathcondition-a} does not yield a closed-form solution for $\psi_{0, \lambda}$ then we may instead leverage \eqref{eq:pathcondition_b} to recursively build estimators. With a slight abuse of notation, we define $r(\eta_0, \lambda) = - \big\{ \dot{D}_{0, R + \lambda\Theta}(\psi_{0, \lambda})\big\}^{-1} D_{ \Theta, P_0}(\psi_{0, \lambda})$, the right-hand side of \eqref{eq:pathcondition_b}. We can now define a procedure that begins with an estimate $\psi_n$ of $\psi_0$, the unconstrained minimizer. For a numerically small step size $d\nu$  we can compute $\psi_{n,d\nu}$ as $\psi_n + r(\eta_n, 0)d\nu$. This procedure proceeds where at the $k$-th step, we define $\lambda_k = \lambda_{k-1}+d\nu$ and compute $\psi_{n,\lambda_k} = \psi_{0,\lambda_{k-1}} + r(\eta_n, \lambda_{k-1})d\nu$. Since we can compute an estimate of $\psi_{n,\lambda}$ for any $\lambda$, it is possible to utilize empirical minimization as above to generate an estimate of $\lambda_0$. 

\subsection{On inequality and multi-dimensional constraints}
\label{subsec:inequality}

In some settings, rather than enforcing the equality constraint $ \Theta_P(\psi)=0$, it suffices to impose the bound $ \Omega_P(\psi)\le0$. We then define the optimal functional parameter $ \Psi^*(P)$ as
\begin{align}
    \Psi^*(P) = \argmin_{\psi \in {\bf \Psi}, \ \Omega_P(\psi) \ \leq \ 0} \ R_P(\psi) \ . 
\end{align}%
Analogous to  \eqref{assmp:pathwise_diff}, we assume that for any $h \in \tangent_P(\psi)$, $h \mapsto \frac{d}{d\delta}
\Omega_P(\psi_{\delta,h})\Big|_{\delta = 0}$is a bounded linear functional on $\tangent_P(\psi)$. The gradient of this mapping is denoted by $D_{\Omega, P}(\psi)$ and we again assume this gradient is non-degenerate, $D_{\Omega,P}(\psi_0^*)(o) \ne 0$ for all $o$ in the support of $O$.

Consider the solution $\psi_{0,	\mu} = \argmin_{\psi \in {\bf \Psi}} R_{P_0}(\psi) + \mu \Omega_{P_0}(\psi)$, where $\mu \in \openr^{\geq 0}$. Similar to the equality constraint, $\{ \psi_{0, \mu} :  \mu \in \openr^{\geq 0} \}$ defines a constraint-specific path through $\psi_0$ at $\mu = 0.$ Drawing from the rationale in the proof of Lemma~\ref{lem:lossmin}, characterizing a solution $\psi_{0, \mu}$ for a particular $\mu$ value allows us to find the optimal solution via the so-called \emph{complementary slackness} condition, $\mu \Omega_{P_0}(\psi_{0, \mu}) = 0$. As in Theorem~\ref{thm:path-characterization}, the path $\{\psi_{0, \mu} : \mu \in \openr^{\geq 0}\}$  satisfies
\begin{align}
	D_{R, P_0}(\psi_{0, \mu}) + \mu D_{\Omega, P_0}(\psi_{0, \mu}) = 0  \ , \quad \text{ for all } \ \mu \in \openr^{\geq 0} \ ,
	\label{eq:lfmpathconditionb} 
\end{align}%
and $\frac{d}{d \mu} \psi_{0, \mu} = - \{ \dot{D}_{0, R + \mu \Omega} (\psi_{0, \mu}) \}^{-1} D_{\Omega, P_0}(\psi_{0, \mu})$, where $\dot{D}_{0, R+\mu\Omega}(\psi) = \frac{d}{d \psi}  D_{0, R + \mu \Omega}(\psi)$ and \\ $D_{0, R + \mu \Omega} = D_{R, P_0}(\psi) + \mu D_{\Omega, P_0}(\psi)$.  

Estimating $\psi_{0, \mu_0}$ follows a similar procedure to estimating $\psi_{0, \lambda_0}$ as discussed previously; however, there is now an initial step to verify the complementary slackness condition. We first check whether the estimate $\psi_{n, \mu=0}$ of the unconstrained parameter $\psi_0$ satisfies $\Omega_n(\psi_{n, \mu=0}) \leq 0$, If so, then we set $\mu_n = 0$ and $\psi_{n, \mu_n} = \psi_n$, i.e., our procedure returns the unconstrained parameter estimate. If the estimate of the unconstrained parameter does not satisfy the estimated constraint, $\Omega_n(\psi_{n, \mu=0}) \leq 0$, then we can use empirical minimization of the constraint to find $\mu_n = \mbox{arg min}_{\mu \in \mathbb{R}^{\ge 0}} |\Omega_n(\psi_{n, \mu_n})|$.

We generalize our methodology to settings with multiple equality and inequality constraints. Given $d$ equality constraints, denoted by $\Theta^{(j)}_P(\psi) = 0$ for $j= 1, \ldots, d$, and $m$ inequality constraints, denoted by $\Omega^{(i)}_P(\psi) \leq 0, \ i = 1, \ldots, m$, the constrained statistical learning task is to determine the optimal functional parameter, $\Psi^*(P)$, defined as 
\begin{equation}\label{eq:target_general}
\begin{aligned}
   &\hspace{1.85cm} \Psi^*(P) = \argmin_{\psi \in {\bf \Psi}} \
    R_P(\psi) \\
    \text{subject to:} \hspace{0.5cm}
    &\Theta^{(j)}_P(\psi) = 0, \ j = 1, \ldots, d \ , \ \text{ and } \ \Omega^{(i)}_P(\psi) \leq 0, \ i = 1, \ldots, m \ . 
\end{aligned}
\end{equation}
We defer methodological extensions for characterizing $\Psi^*(P)$ in \eqref{eq:target_general} to Appendix~\ref{app:methods-multi}.

\section{A class of constraints with closed-form solutions}
\label{sec:general_constraint}

We consider a broad class of equality constraints under which our method yields closed-form solutions to risk minimization problems under both mean squared error (MSE) and cross-entropy risk criteria. Specifically, we consider constraints of the form:
\begin{align}
    \Theta_P(\psi) = \E_{P}\{ \kappa_P(Z) \ \zeta_P(\psi)(Z) \} \ ,  
    \label{eq:general_constraint}
\end{align}%
where $\psi(Z)$ is the unconstrained functional parameter mapping covariates $Z$ to either $\mathbb{R}$ (under MSE risk with $L(\psi)(o) = \{y - \psi(z)\}^2$) or $[0,1]$ (under cross-entropy risk with $L(\psi)(o) = -y\log \psi (z) - (1-y)\log \{1-\psi(z)\}$). Here, $\kappa_P$ is a weighting function not depending on $\psi$, and $\zeta_P(\psi)$ is a differentiable function of $\psi$. We note that $\kappa_P$ acts as a weighting function, which may assume negative values, and our interest lies in minimizing predictive risk subject to a constraint on the weighted average of $\zeta_P$. We let $\kappa_0 = \kappa_{P_0}$ and $\zeta_0 = \zeta_{P_0}$. 


We now derive the canonical gradients for the two risks and constraint functional in \eqref{eq:general_constraint}, and show how closed-form solutions for the optimal constrained parameter can be obtained. Paths through $\psi \in \bm{\Psi}$ can be characterized via $\{\psi_{\delta, h} : h \in L^2(P_{Z})\}$, where $L^2(P_{Z})$ denotes the space of real-valued functions with finite second moment on the support of $Z$ implied by $P$. We take $\tangent_{P}(\psi) = L^2(P_Z)$.  

\vspace{-0.15cm}
\begin{theorem} 
    The canonical gradients for the mean squared error and cross-entropy risks are 
    \begin{align}
        D_{R, P_0}(\psi)(Z) &= 2\{\psi(Z) - \psi_0(Z)\} \ , \qquad (\textit{\small mean squared error risk}) \label{eq:R_gradient_mse} \\
        D_{R, P_0}(\psi)(Z) &= \displaystyle \frac{\psi(Z) -\psi_0(Z)}{\psi(Z) \ (1- \psi(Z))} \ . \qquad (\textit{\small cross-entropy risk})\label{eq:R_gradient_cross}
    \end{align}

    In addition, the canonical gradient of the constraint defined in \eqref{eq:general_constraint}, is: 
    \begin{align}
        D_{\Theta, P_0}(\psi)(Z) =  \kappa_0(Z) \ \dot{\zeta}_0(\psi)(Z) \ , \quad \mbox{where} \quad \dot{\zeta}_0(\psi) = {d\zeta_0(\psi)}/{d\psi} \ .
        \label{eq:theta_gradient}
    \end{align}  
    \label{thm:gradients}
\end{theorem}%

\vspace{-1.1cm}
By Theorem \ref{thm:gradients}, when $\zeta(\psi)=\psi$, we have $D_{\Theta, P_0}(\psi) = \kappa_0$, which does not depend on $\psi$. 

\begin{corollary} 
Given the MSE risk criterion and constraint in \eqref{eq:general_constraint} with $\zeta(\psi)=\psi$,  \eqref{eq:pathcondition-a} implies 
\begin{align}
   \psi_{0}^*(z) \equiv \psi_{0, \lambda_0}(z)  = \psi_0(z) - \lambda_0 \kappa_0(z)/2 \ , 
    \label{eq:closed_form_mse}
\end{align}
where $\lambda_0$ is obtained in closed form by solving $\Theta_{P_0}(\psi_{0, \lambda_0}) = 0$, yielding $\lambda_0 = 2\Theta_{P_0}(\psi_0)/P_0 \kappa_0^2$, where $P_0 \kappa_0^2 = \int \kappa_0(z)^2 dP_0(z)$. 

Additionally, under the cross-entropy risk, \eqref{eq:pathcondition-a} implies that $\psi_0^*$ satisfies the quadratic equation, $\lambda_0 \kappa_0 \psi_0^{*2} - \left\{1 + \lambda_0 \kappa_0\right\} \psi_0^* + \psi_0 = 0$, which yields the unique solution
\begin{align}
    \psi_{0}^*(z) \equiv \psi_{0, \lambda_0}(z) =  \frac{\lambda_0 + \kappa_{0}^{-1}(z) + (-1)^{\I(\kappa_0(z) > 0)}  [ \{ \lambda_0 + \kappa_{0}^{-1}(z) \}^2 - 4\lambda_0 \kappa_{0}^{-1}(z)\psi_0(z) ]^{1/2}}{2\lambda_0} \ , 
    \label{eq:closed_form_cross_etropy}
\end{align}
where $\lambda_0$ is determined implicitly by solving $\Theta_{P_0}(\psi_{0, \lambda_0}) = 0$. 
\label{corollary:closed_forms}
\end{corollary}

\subsection{Conditions for optimal risk and constraint satisfaction} 

The results of the previous subsection provide a  pathway for incorporating constraints into learning procedures with minimal computational overhead, allowing the use of any off-the-shelf learner. Not only is this approach computationally convenient, but it also allows for straightforward characterization of the asymptotic behavior of the constrained estimator in terms of the performance of its constituent nuisance estimates, as we now demonstrate.

Consider the mean squared error risk criterion and the constraint in \eqref{eq:general_constraint} with $\zeta(\psi)=\psi$. A plug-in estimator of $\psi^*_0$ in \eqref{eq:closed_form_mse}, denoted by $\psi^*_n$, is given by
\begin{align}
    \psi_{n}^*(z) = \psi_n(z) -  \Theta_{n}(\psi_n) \frac{\kappa_n(z)}{P_n \kappa^2_n} \ , 
    \label{eq:psi*n}
\end{align}%
where $\psi_n$, $\Theta_{n}$, $\kappa_n$ denote estimates of $\psi_0$, $\Theta_{P_0}$, $\kappa_0$, respectively, and $P_n \kappa^2_n = \frac{1}{n} \sum_{i = 1}^n \kappa^2_n(Z_i).$ 

We now formalize sufficient conditions under which the estimator $\psi^*_n$:
(i) achieves the optimal penalized risk such that $R_{P_0}(\psi^*_n) - R_{P_0}(\psi^*_0) = o_{P_0}(1)$,
and (ii) satisfies the constraint such that $\Theta_{P_0}(\psi^*_n) = o_{P_0}(1)$.  Let $|| f ||_1 = P_0(|f|)$ denote the $L^1(P_0)$-norm, and $|| f ||_2 = (P_0f^2)^{1/2}$ the $L^2(P_0)$-norm of the $P_0$-measurable function $f$.

\begin{theorem}
    Under regularity conditions provided in Appendix~\ref{app:proofs}, $\psi^*_n$ in \eqref{eq:psi*n} achieves the optimal penalized risk, i.e., $R_{P_0}(\psi^*_n) - R_{P_0}(\psi^*_0) = o_{P_0}(1)$, if the following conditions are met: 

    \vspace{-0.4cm}
    \begin{enumerate}
        \setlength{\itemsep}{-0.2cm}
        \item[C1.] $L^2(P_0)$-consistency for the estimator $\psi_n$: $|| \psi_n - \psi_0 ||_2 = o_{P_0}(1)$, 
        \item[C2.] $L^2(P_0)$-consistency for the estimator $\kappa_n$: $|| \kappa_n - \kappa_0 ||_2 = o_{P_0}(1)$, 
        \item[C3.] Consistency of the estimate $\Theta_n$: $\Theta_n - \Theta_0 = o_{P_0}(1)$, and 
        \item[C4.] Consistency of the estimate $P_n \kappa_n^2$: $P_n \kappa_n^2- P_0 \kappa_0^2 = o_{P_0}(1)$. 
    \end{enumerate}

    \vspace{-0.4cm}
    Further, assume $\psi_n$ converges in probability to $\tilde\psi\in \bf \Psi$, where $\tilde\psi$ may differ from $\psi_0$. $\psi^*_n$ satisfies the constraint, i.e., $\Theta_{P_0}(\psi^*_n) = o_{P_0}(1)$, if (C4) and the following conditions are met: 

    \vspace{-0.4cm}
    \begin{enumerate}
        \setlength{\itemsep}{-0.2cm}
        \item[C5.] $L^1(P_0)$-consistency for the estimator $\kappa_n$: $|| \kappa_n - \kappa_0 ||_1 = o_{P_0}(1)$, and
        \item[C6.] Consistency of the estimate $\Theta_n$: $\Theta_n(\tilde{\psi}) - \Theta_{0}(\tilde{\psi}) = o_{P_0}(1)$. 
    \end{enumerate}
    \label{thm:risk_and_constraint_at_P0}
\end{theorem} 

\vspace{-0.35cm}
Theorem~\ref{thm:risk_and_constraint_at_P0} provides conditions under which the plug-in constrained estimator achieves optimal penalized risk and constraint satisfaction at the true distribution $P_0$. For this to be practically attainable, we typically will require consistent estimation of the key components of $P_0$ used to construct the estimator, including: the unconstrained parameter $\psi_0$, the constraint gradient $\kappa_0$, and any nuisance parameters involved in evaluating $\Theta_{P_0}(\psi_0)$. However, in some applications, such as fair machine learning based on counterfactual-valued constraints (e.g., \citep{zhang2017causal, nabi2018fair, chiappa2019path}), it may be possible to leverage theory for doubly- and multiply-robust estimators in order to achieve optimal risk and constraint satisfaction even under some forms of misspecification, as we will see below. 




\subsection{Examples} 
\label{subsec:examples}

We consider three representative constraints—the average total effect, natural direct effect, and equalized risk among cases—under both mean squared error and cross-entropy risk criteria. For both risks, the unconstrained minimizer is a conditional mean (or a conditional probability when $Y$ is binary). Throughout, we assume $\mathcal{M}$ is a nonparametric model, $X$ is a binary variable, and $O = (Z, Y) \sim P_0 \in \mathcal{M}$. In each example, we express the constraint in the form of \eqref{eq:general_constraint} by identify the corresponding $\kappa_0$ and ${\zeta}_0(\psi)$, derive an estimator for the optimal constrained parameter, and discuss convergence and robustness conditions. Detailed derivations and second-order condition analyses are provided in the supplementary materials. 


\subsubsection{Average total effect}
\label{subsec:ate}

In some applications, it is desirable to constrain the \textit{average treatment effect} (ATE) when learning a predictive model, either to reflect known causal relationships or to promote fairness. From a causal validity perspective, such constraints can enforce alignment with prior evidence, such as results from clinical trials or expert-driven benchmarks. From a fairness perspective, treating the exposure as sensitive, ATE constraints can promote fairness by ensuring that average predictions remain stable under counterfactual changes to group membership. This is especially relevant in settings like risk assessment or policy evaluation, where equitable treatment is a priority.  

Let $Z = (X, W)$, where $W$ contains all confounders between $X$ and $Y$. Define $\Theta_{P_0}(\psi) = \int \{\psi(1, w) - \psi(0, w)\} dP_{0,W}(w)$, where $P_{0,W}$ is the marginal distribution of $W$ under $P_0$. Under standard causal identification assumptions, $\Theta_{P_0}(\psi) = 0$ corresponds to imposing a constraint that sets the ATE of $X$ on $Y$ to zero. This constraint fits the general form in \eqref{eq:general_constraint} by letting $\zeta_0(\psi) = \psi$ and $ \kappa_0(z) = (2x-1)/\pi_0(x \mid w)$, where $\pi_{0}(x \! \mid \! w) = P_0(X = x \! \mid \! W = w)$. In this case, $P_0 \kappa_0^2 = \int \left\{ \pi_0(1 \mid w)^{-1} + \pi_0(0 \mid w)^{-1} \right\} dP_{0, W}(w)$.  

Assuming the \textbf{mean squared error risk criterion}, $\psi^*_0(x, w)$ under the null ATE constraint is given in a closed form by \eqref{eq:closed_form_mse}, with an estimator given in \eqref{eq:psi*n}. The key nuisance parameters indexing $\kappa_0$ and $\lambda_0$ are $\eta_0 = (\psi_0, \pi_0, P_{0,W})$. The parameters $\psi_0$ and $\pi_0$ can be estimated using any appropriate statistical learning technique for conditional means/probabilities, while a natural choice for $P_{0,W}$ is its empirical analog. Thus, we consider $\kappa_n(x,w) = (2x - 1)/\pi_n(x \mid w)$, $P_n \kappa_n^2 =\frac{1}{n} \sum_{i=1}^n \left\{ \pi_n(1 \mid W_i)^{-1} + \pi_n(0 \mid W_i)^{-1} \right\}$, while $\Theta_n(\psi_n)$ can be obtained via a plug-in g-formula as $\frac{1}{n}\sum_{i=1}^n \psi_n(1,W_i) - \psi_n(0,W_i)$. 

Under the assumptions $|| \pi_n - \pi_0 ||_2 = o_{P_0}(n^{-1/a})$ and $|| \psi_n - \psi_0 ||_2 = o_{P_0}(n^{-1/b})$, and under regularity conditions detailed in Appendix~\ref{app:ate}, the $L^2(P_0)$-consistency rate of the constrained estimator $\psi^*_n$ is dominated by the slower of the two nuisance rates, i.e., $|| \psi^*_{n} - \psi^*_{0} ||_2 = o_{P_0}(n^{\max\{-{1}/{a}, - {1}/{b}\}})$. By Theorem~\ref{thm:risk_and_constraint_at_P0}, $\psi^*_n$ achieves the optimal penalized risk whenever both $\psi_n$ and $\pi_n$ are consistent. In contrast, constraint satisfaction does not require consistency of $\psi_n$. If $\Theta_n$ is constructed using a doubly robust method---such as augmented inverse probability weighting (AIPW) or targeted minimum loss-based estimation (TMLE)---then the constraint is satisfied as long as $\pi_n$ is consistent, regardless of the specification of $\psi_n$.

Assuming the \textbf{cross-entropy risk criterion}, $\psi^*_0(x,w)$ under the null ATE constraint is given by \eqref{eq:closed_form_cross_etropy},  where a closed-form expression for $\lambda_0$ is not readily available. Thus, we propose to build an estimator by generating estimates $\eta_n = (\psi_n, \pi_n)$ of $\eta_0 = (\psi_0, \pi_0)$. For a given $\lambda$, we propose to estimate $\psi_{0,\lambda}$ using a plug-in estimate $\psi_{n,\lambda}$ based on \eqref{eq:closed_form_cross_etropy}, where $\psi_0$ and $\pi_0$ are replaced with estimates $\psi_n$ and $\pi_n$, respectively. An estimate of $\lambda$ can then be obtained as $\lambda_n = \mbox{arg min}_{\lambda \in \mathbb{R}} \left| \sum_{i=1}^n \{ \psi_{n,\lambda}(1, W_i) - \psi_{n,\lambda}(0,W_i)\} \right|,$
where the minimization can be solved via grid search. Note that the objective function for obtaining $\lambda_n$ may also incorporate a robust estimator of the ATE in place of the plug-in estimator used here. We observe the same $L^2(P_0)$-consistency behavior as in the MSE risk criterion.

For derivations and additional details on the ATE‐constraint example, see Appendix~\ref{app:ate}.

\subsubsection{Natural direct effect} 
\label{subsec:nde}

We now consider a constraint on the \textit{natural direct effect} (NDE), which captures the effect of a treatment on the outcome not mediated by an intermediate variable. Such constraints are useful for preserving known causal structure or promoting fairness, particularly when the goal is to control for mediation effects in prediction or decision-making.

Let $Z = (X, M, W)$, where $M$ is the set of mediators between $X$ and $Y$. Define $\Theta_{P_0}(\psi) = \int \left\{ \psi(1, m, w) - \psi(0, m, w) \right\} dP_{0, M|X,W}(m \! \mid 
\! 0, w)  dP_{0,W}(w)$, where $P_{0,M|X,W}(m \! \mid \! x, w)$ is the conditional probability distribution of $M$ under $P_0$, which we denote by $f_{0,M}(m \mid x, w)$. Under causal identification assumptions, $\Theta_{P_0}(\psi) = 0$ implies a null NDE of $X$ on $Y$. 

This constraint is a special case of \eqref{eq:general_constraint}, with $\zeta_0(\psi) = \psi$ and $\kappa_0(z) = \{ (2x-1)/\pi_0(x \! \mid  \! w)\} \gamma_0(m \! \mid \! x, w)$, where $\gamma_0(m \! \mid \! x, w) = f_{0, M}(m \! \mid \! 0, w)/f_{0, M}(m \! \mid \! x, w)$. In this case, $P_0 \kappa_0^2 = \iint \left\{ \pi_0(1 \! \mid  \! w)^{-1} \gamma_0(m \! \mid  \! x, w) + \pi_0(0 \! \mid \! w)^{-1} \right\} dP_{0, M|X,W}(m \! \mid \! 0, w) \ \! dP_{0, W}(w)$. 

Consequently, assuming the \textbf{mean squared error risk criterion}, $\psi^*_0(x, m, w)$ under the null NDE constraint takes the form of \eqref{eq:closed_form_mse}, with a corresponding estimator in \eqref{eq:psi*n}. The key nuisance parameters are $\eta_0 = (\psi_0, \pi_0, f_{0, M}, P_{0,W})$. Using the empirical distribution for $P_{0, W}$ and $(\psi_n, \pi_n, f_{n, M})$, we get $\gamma_n(m | x, w) = f_{n, M}(m | x, w)/f_{n, M}(m | x, w)$, $\kappa_n(x,m, w) = \{(2x - 1)/\pi_n(x | w)\} \gamma_n(m | x, w)$, $\Theta_n(\psi_n) = \frac{1}{n}\sum_{i=1}^n \sum_{m} \{\psi_n(1,m, W_i) - \psi_n(0,m, W_i)\}f_{n, M}(m | X_i, W_i)$, and $P_n \kappa_n^2 =\frac{1}{n} \sum_{i=1}^n \sum_{m} \left\{ \pi_n(1 | W_i)^{-1} \gamma_n(m | X_i, W_i) + \pi_n(0 | W_i)^{-1} \right\} f_{n, M}(m | X_i, W_i)$.
When $M$ is continuous valued, marginalization may involve numeric or Monte Carlo integration. 

Given the $L^2(P_0)$-consistency rates for $\psi_n$ and $\pi_n$ established in the previous example, and assuming $|| f_{n,M} - f_{0,M} ||_2 = o_{P_0}(n^{-1/c})$, the $L^2(P_0)$-consistency rate of $\psi^*_n$ is governed by the slowest of the three nuisance rates. Under regularity conditions detailed in Appendix~\ref{app:nde}, we have $|| \psi^*_{n} - \psi^*_{0} ||_2 = o_{P_0}(n^{\max\{-{1}/{a}, - {1}/{b}, -1/c\}})$. By Theorem~\ref{thm:risk_and_constraint_at_P0}, $\psi^*_n$ achieves the optimal penalized risk whenever $\psi_n$, $\pi_n$, $f_{n, M}$ are consistent. However, if $\Theta_n$ is constructed using a robust estimator---such as AIPW or TMLE---then the constraint can be satisfied without requiring consistency of $\psi_n$. 

When $M$ is multivariate or continuously valued, direct estimation of $f_{0,M}$ can be challenging. In such cases, we may reparameterize $\kappa_0$ and $\Theta_{P_0}$ to avoid density estimation by leveraging sequential regressions and Bayes’ rule. For example, $\gamma_0(m \mid x, w)$ can be rewritten as $\{\alpha_0(0 \mid w, m)/\alpha_0(x \mid w, m)\} \times \{\pi_0(x \mid w)/\pi_0(0 \mid w)\}$, where $\alpha_0(x \mid w, m) = P_0(X = x \mid W = w, M = m)$, and can be estimated using binary regression, including parametric models such as logistic regression or flexible machine learning methods. With this reparameterization, the requirement for $L^2(P_0)$-consistency of $f_{n,M}$ is replaced by that of $(\pi_n, \alpha_n)$.

Under the \textbf{cross-entropy risk criterion}, we can obtain $\psi^*_n$ using \eqref{eq:closed_form_cross_etropy}, where $\lambda_0$ is estimated by $\lambda_n = \underset{\lambda \in \mathbb{R}}{\mbox{arg min}} | \sum_{i=1}^n \sum_{m} \{ \psi_{n,\lambda}(1, m, W_i) - \psi_{n,\lambda}(0, m, W_i)\} f_{n, M}(m \mid 0, W_i)|$. When $M$ is continuous, the summation over $m$ is replaced with numerical integration. The minimization can be carried out via grid search, and the objective function may incorporate a robust estimator of the NDE in place of the plug-in version. The $L^2(P_0)$-consistency behaviors matches that of the MSE case. 

For derivations and additional details on the NDE‐constraint example, see Appendix~\ref{app:nde}.

\subsubsection{Equalized risk} 
\label{subsec:erica}

Predictive systems are often evaluated not only by overall performance but also by how performance varies across groups. In many settings, it is important to prevent predictive risk from concentrating disproportionately within subpopulations defined by sensitive attributes or outcomes. Concepts such as distributionally robust optimization, invariant risk minimization, equality of opportunity, and risk parity in finance reflect this goal, promoting uniformity in predictive accuracy or error rates across groups or environments \citep{denuit2007actuarial, maillard2010properties, roncalli2013introduction, hardt2016equality, sagawa2019distributionally}. 

As a concrete example, we consider a constraint that enforces equality of expected predictive risk among individuals with $Y = 1$; i.e., the ``cases'' in a binary outcome setting. We refer to this as the \emph{equalized risk in cases} (ERIC) constraint. Suppose $Z = (W, X)$, and let $\psi(Z)$ map $Z$ to $[0,1]$. We define the constraint as:
$\Theta_{P_0}(\psi) = \E_{P_0}\{L(\psi)(O) \mid X = 1, Y = 1\} - \E_{P_0}\{L(\psi)(O) \mid X = 0, Y = 1\}$, where $L(\psi)$ is the cross-entropy (negative log-likelihood) loss. This constraint targets group-level equality in risk among the positive outcome class, and highlights distinct technical features relative to the causal constraints discussed earlier. 

The ERIC constraint fits within the general form of \eqref{eq:general_constraint} with $\zeta_0(\psi) = \log(\psi)$ and $\kappa_0(z) =  (1-2x)\psi_0(x,w)/p_{0,1}(x)$, where $p_{0,1}(x) = P_0(X=x, Y=1)$. By Theorem~\ref{thm:gradients}, $D_{\Theta, P_0}(\psi)(o) = - (2x-1)/p_{0,1}(x) \{ \psi_0(x,w)/\psi(x,w)\}$. Combined with the risk gradient, Condition~\eqref{eq:pathcondition-a} yields a closed-form solution for the constraint-specific path: 
\begin{align}
    \psi_{0, \lambda}(x, w) = \psi_0(x, w) \ \{1 + \lambda \ C_0(x) \}/\{1 + \lambda \ C_0(x) \ \psi_0(x, w)\} \ ,
	\label{eq:ex_erica}
\end{align}
where $C_0(x) = (2x-1)/p_{0,1}(x)$. A closed-form expression for $\lambda_0$ is not readily available. Thus, for estimation of $\psi_{0, \lambda_0}$, we first generate estimates $\psi_n$ of $\psi_0$. We then compute a parsimonious estimate of $p_{0,1}(x)$ by setting $p_{n,1}(x) = \sum_{i=1}^n \{I(X_i = x) \psi_n(X_i, W_i)\} / \sum_{i=1}^n I(X_i = x)$. An estimate of $\lambda_0$ can then be obtained via empirical minimization of the constraint,
\begin{align}
    \lambda_n = \underset{-p_{n,1}(1) < \lambda < p_{n,1}(0)}{\mbox{arg min}} \left| \sum_{i=1}^n  Y_i C_n(X_i) \log \psi_{n, \lambda}(X_i, W_i)   \right| \ , \label{eq:emp_mini_lambda_est}
\end{align}
where $C_n(x) = (2x - 1)/p_{n,1}(x)$ and $\psi_{n, \lambda}(x,w) = \psi_n(x,w) (1 + \lambda C_n(x))/(1 + \lambda C_n(x) \psi_n(x,w))$. The restriction on the range of $\lambda$ in \eqref{eq:emp_mini_lambda_est} is motivated by second-order condition analysis (see Appendix~\ref{app:er_cases}), and $\lambda_n$ can be computed via grid search. Additional derivations and asymptotic details for the ERIC constraint are provided in Appendix~\ref{app:er_cases}. 

\subsubsection{Additional examples} 

Additional examples, illustrating our framework's broader applicability, are provided in the supplementary materials. These include:
(i) causal constraints including the conditional average treatment effect, effect on the treated, and effect on the controls (Appendix~\ref{app:cate});
(ii) overall equalized risk and two-dimensional equalized risk stratified by outcome (Appendix~\ref{app:er_overall});
(iii) density estimation under moment constraints (Appendix~\ref{app:moment_rest}); and
(iv) bias corrected estimation under model misspecification (Appendix~\ref{app:misspecified}).


\section{Simulations}
\label{sec:sims}

We evaluated the asymptotic performance of our estimators in several settings. For each bound on the constraint, we study (i) the convergence of the risk of our estimator $R_{P_0}(\psi_{n,\lambda_n})$ to the optimal constrained risk $R_{P_0}(\psi_{0,\lambda_0})$ and (ii) that the true fairness constraint under $\psi_{n,\lambda_n}$, $\Theta_{P_0}(\psi_{n,\lambda_n})$, converges appropriately. In all simulations, we evaluated our proposed estimators when built based on both equality and inequality constraints. Part of our results are relegated to Appendix~S12.

For each setting, we generated 1000 simulated data sets of sample sizes $n$ = 100, 200, 400, 800, 1600 and used an independent test set of 1$e$6 observations to numerically approximate the value of the true risk and the true constraint.

Our first set of simulations considered a setting where the true underlying functional parameters fall into known finite-dimensional parametric models. We considered two data generating mechanisms. Both data generating mechanisms shared the same joint distribution of $(X, W)$. The vector $W$ consisted of six components $W_1, \dots, W_6$ drawn from the following distribution: $W_1 \sim \mbox{Bern}(1/4), W_2 \mid W_1 \sim \mbox{Bern}(\mbox{expit}(W_1))$, where \mbox{expit} is the inverse logit function, $W_3 \mid W_1, W_2 \sim \mbox{Bern}(\mbox{expit}(-W_2 + W_1))$, $W_4 \sim \mbox{Norm}(0,1)$, $W_5 \sim \mbox{Unif}(0,1)$, $W_6 \sim \mbox{Exp}(1)$. Given $W$, $X$ was generated from a $\mbox{Bern}(\mbox{expit}(W_1 - W_2/3 - W_6 / 10))$. We then considered separate means of generating (i) a continuous outcome to study the ATE and NDE constraints in Sections~\ref{subsec:ate} and \ref{subsec:nde} and (ii) a binary outcome $Y$ to study the equal risk in the cases constraint in Section~\ref{subsec:erica}.

\subsection{Average treatment effect and natural direct effect} \label{subsec:ate_nde_sim}

In this setting, we simulated a binary mediator $M$ such that $f_{0, M}(1 \mid x, w) = \mbox{expit}(-1 - x - w_1 + w_2/2 - w_5/2)$ and simulated the outcome $Y$ from a Normal distribution with mean given $(X,W,M)$ equal to $-X + 2M + 2W_1 - W_3 - W_4 + 2W_5$ and variance equal to 4. We separately considered the prediction problem of minimizing mean squared error of constrained predictions of $Y$ given $(X,W)$-only under the ATE constraint (Section~\ref{subsec:ate}) and given $(X,W,M)$ under the NDE constraint (Section~\ref{subsec:nde}). Note that the conditional distribution of $Y$ implies that linear regression can be used to correctly model both the conditional mean given $(X,W)$ (i.e., $\psi_0$ for the ATE constraint problem), as well as the conditional mean given $(X,W,M)$ (i.e., $\psi_0$ for the NDE constraint problem). Both these problems also require estimates of $\pi_0$, which was estimated using maximum likelihood based on a correctly-specified logistic regression model.

We compared our proposed method to an alternative proposal based on constrained maximum likelihood estimation \citep{nabi2018fair}. This approach utilizes the Constrained Optimization BY Linear Approximations (COBYLA) approach to maximizing the mean squared error risk criteria over a linear regression model constrained so that either the implied ATE or NDE was less than the specified constraint bound \citep{powell1994direct}. 

\begin{figure}
\centering
\includegraphics[width=6.5cm, height=5.5cm]{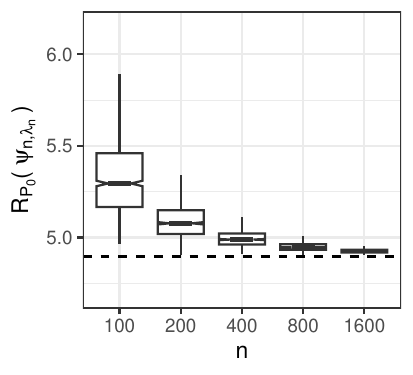}
\includegraphics[width=6.5cm, height=5.5cm]{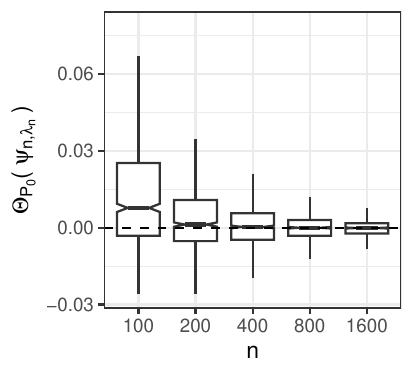}
\includegraphics[width=6.5cm, height=5.5cm]{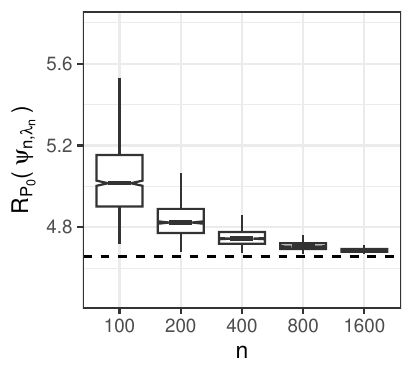}
\includegraphics[width=6.5cm, height=5.5cm]{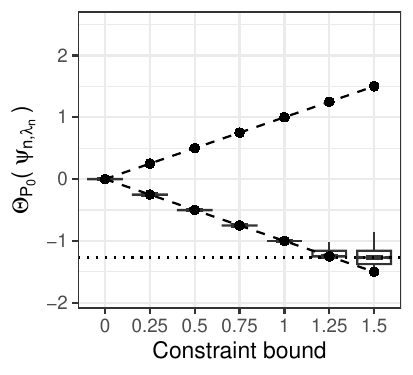}
\caption{\textbf{Average treatment effect constraint and mean squared error.} \underline{Top left:} Distribution of risk of $\psi_{n,\lambda_n}$ over 1000 realizations for each sample size for the equality constraint $\Theta_{P_0}(\psi) = 0$. The dashed line indicates the optimal risk $R_{P_0}(\psi_0^*)$. \underline{Top right:} Distribution of the true constraint over 1000 realizations for each sample size. The dashed line indicates the equality constraint value of zero. The constraint value under the unconstrained $\psi_0$, $\Theta_{P_0}(\psi_0) = -1.27$ and is not shown due to the scale of the figure. \underline{Bottom left:} Distribution of risk of $\psi_{n,\lambda_n}$ over 1000 realizations for each sample size for the inequality constraint $|\Theta_{P_0}(\psi)| \le 0.5$. The dashed line indicates the optimal risk $R_{P_0}(\psi_0^*)$ under this constraint. \underline{Bottom right:} Distribution of the true constraint for estimators built using the equality constraint (constraint bound = 0) and inequality constraints with varying bounds at $n = 800$. The dotted line shows the value of the constraint under $\psi_0$, $\Theta_{P_0}(\psi_0)$. The dashed lines shows the positive and negative bounds on the constraint.
}
\label{fig:ate_sim}
\end{figure}

Our proposed estimator performed as expected in terms of achieving optimal risk and controlling the constraint. Figure \ref{fig:ate_sim} shows results for the ATE constraint. The top left panel indicates that our proposed approach nears the optimal risk as sample size increases, while still controlling the constraint with high probability (top right panel). When we relax the constraint bound to only enforce that the ATE is less than 1/2 (bottom right), we find that (i) as expected, the optimal value of the risk (indicated by the dashed line) lowers and our estimator again achieves this optimal risk as sample size increases. Finally, we observed that for each fixed sample size, our estimator appropriately respects the nature of inequality constraints when the bound on the constraint is set larger than the value of $\Theta_{P_0}(\psi_0)$. Results displaying the distribution of the constraint for sample size $n = 800$ are shown in the bottom right panel. We find that once the bound on the constraint is larger than the true ATE (around -1.27), our procedure shifts to using $\psi_n$, the unconstrained estimate of $\psi_0$. Accordingly, the true value of the constraint for the estimate levels off at $\Theta_{P_0}(\psi_0)$. Similar results were seen for the NDE constraint (Appendix~S12).

Comparing our method to constrained maximum likelihood, we found that our method resulted in improved predictive performance with similar control of the constraint (Figure \ref{fig:sim_comp_to_mle}). Similar results were observed for the NDE, but improvements for our proposed methodology over constrained maximum likelihood were considerably more pronounced (Appendix~S12).

\begin{figure}[!t]
    \centering
    \includegraphics[width=0.46\linewidth]{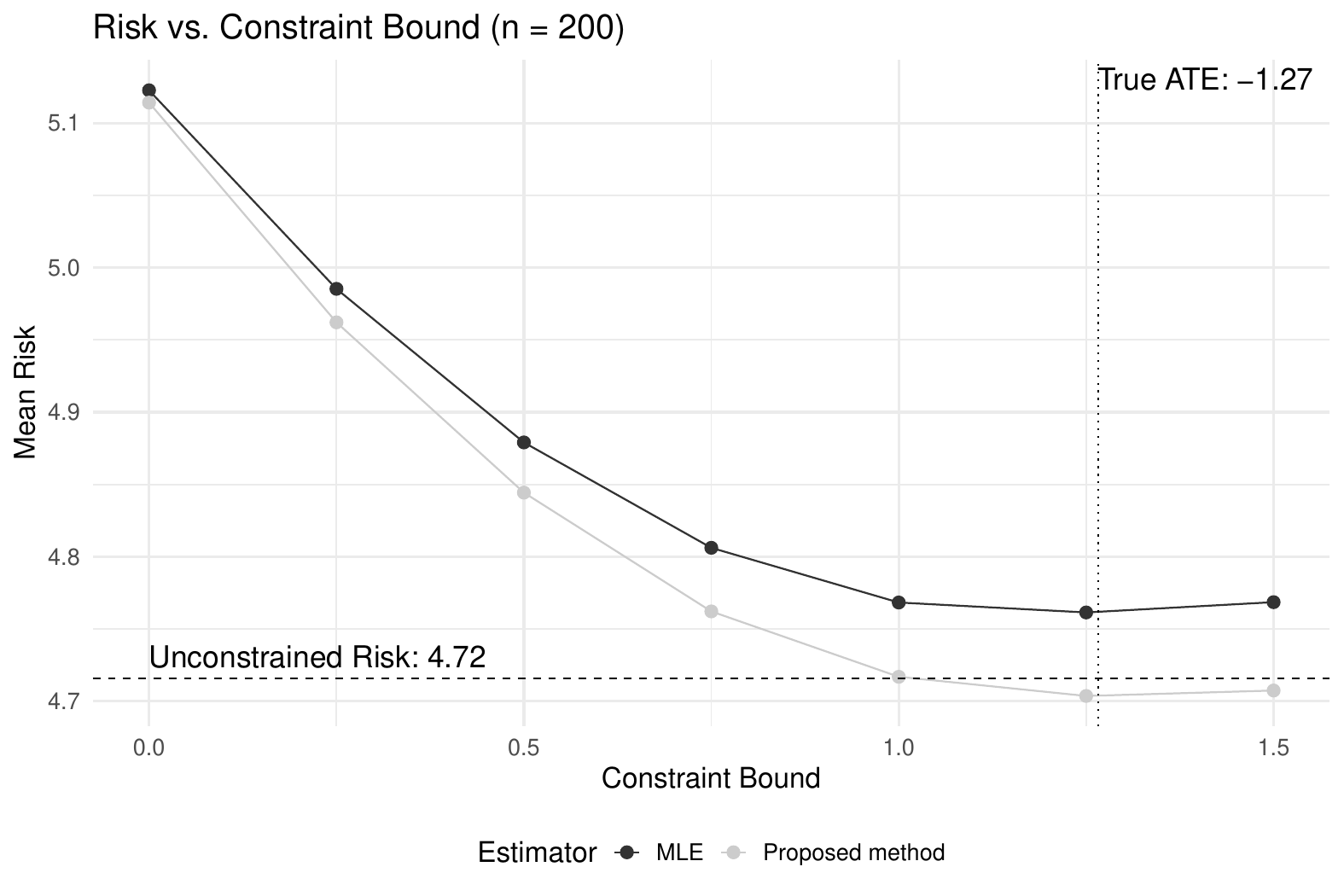}
    \includegraphics[width=0.46\linewidth]{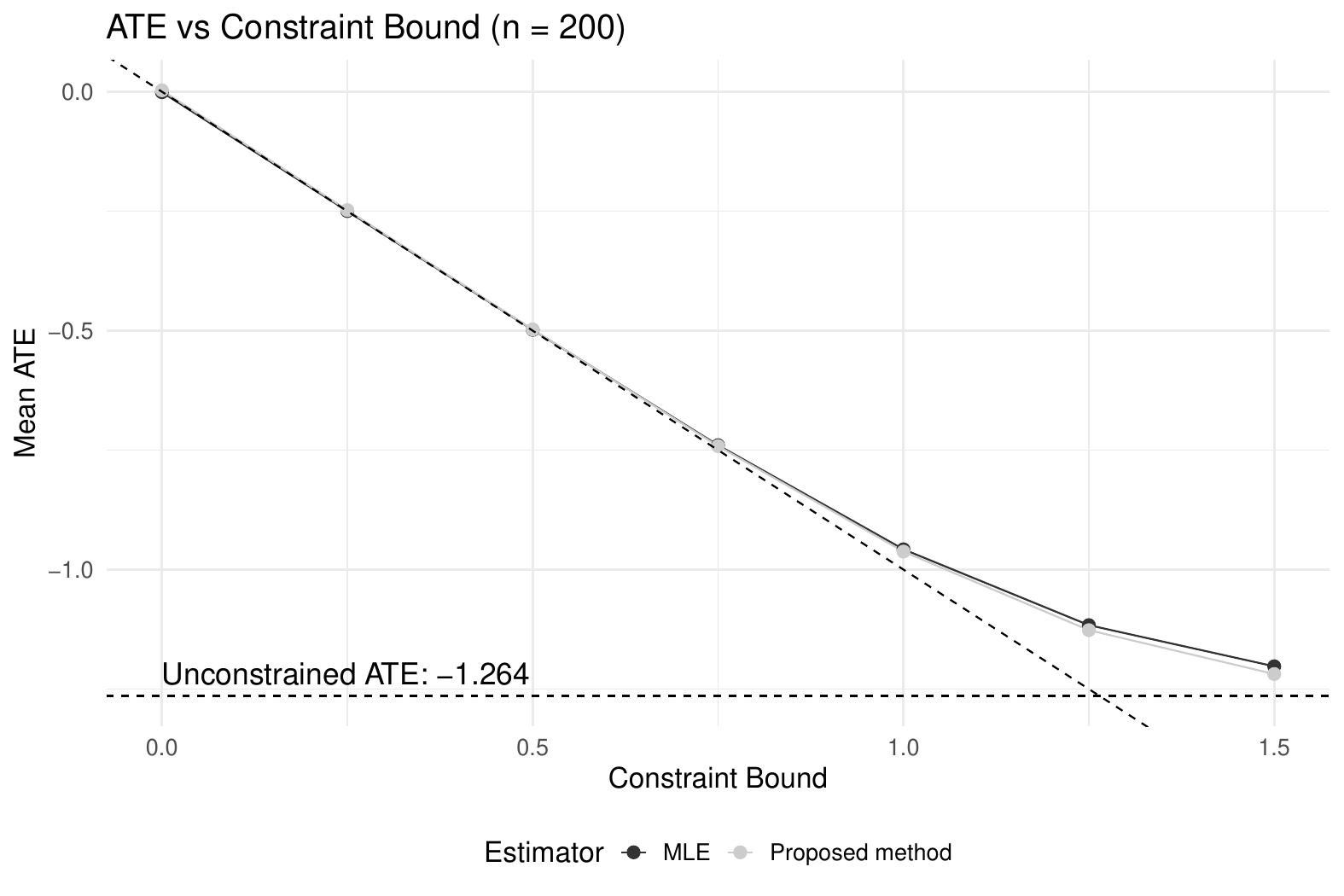}
    \includegraphics[width=0.46\linewidth]{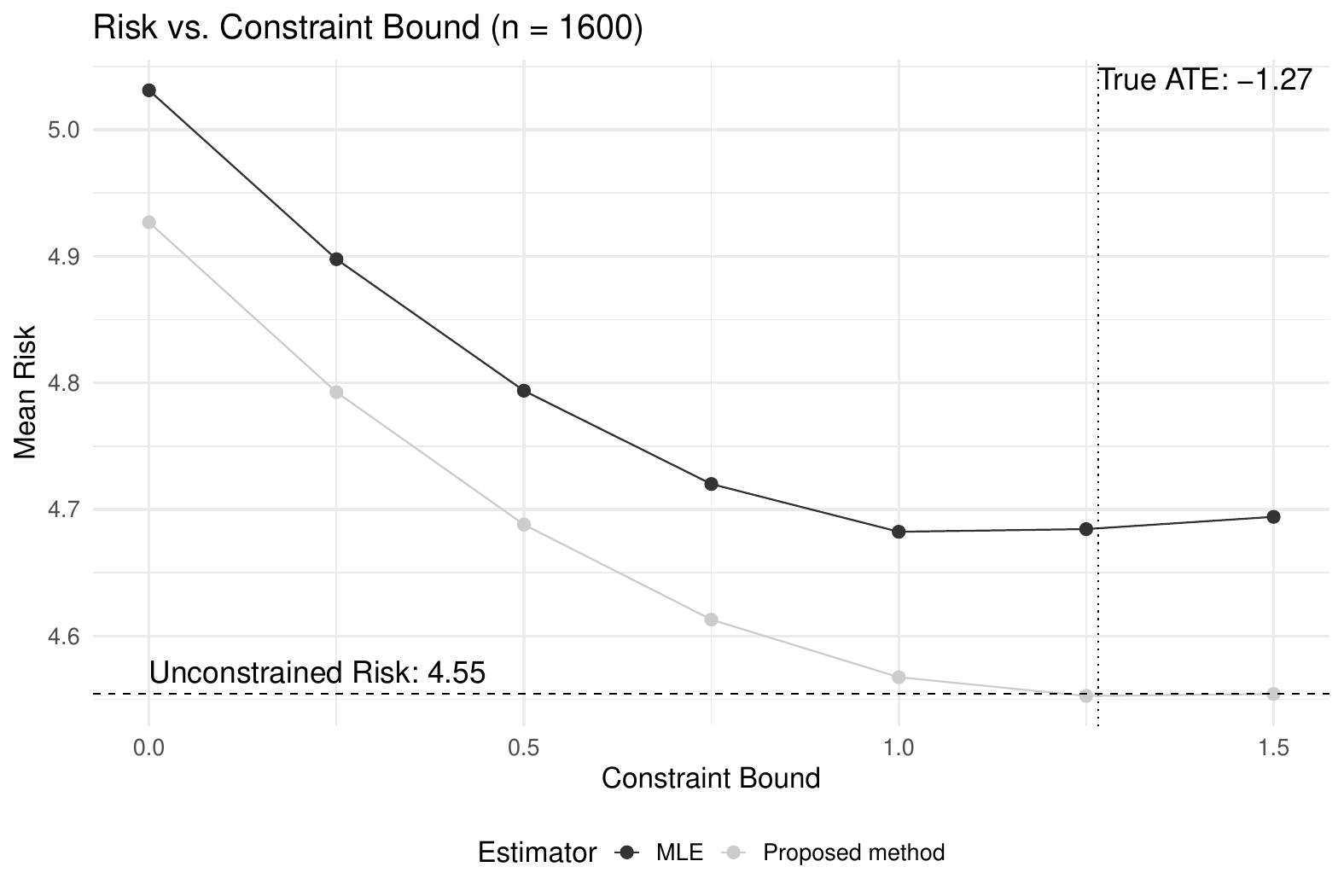}
    \includegraphics[width=0.46\linewidth]{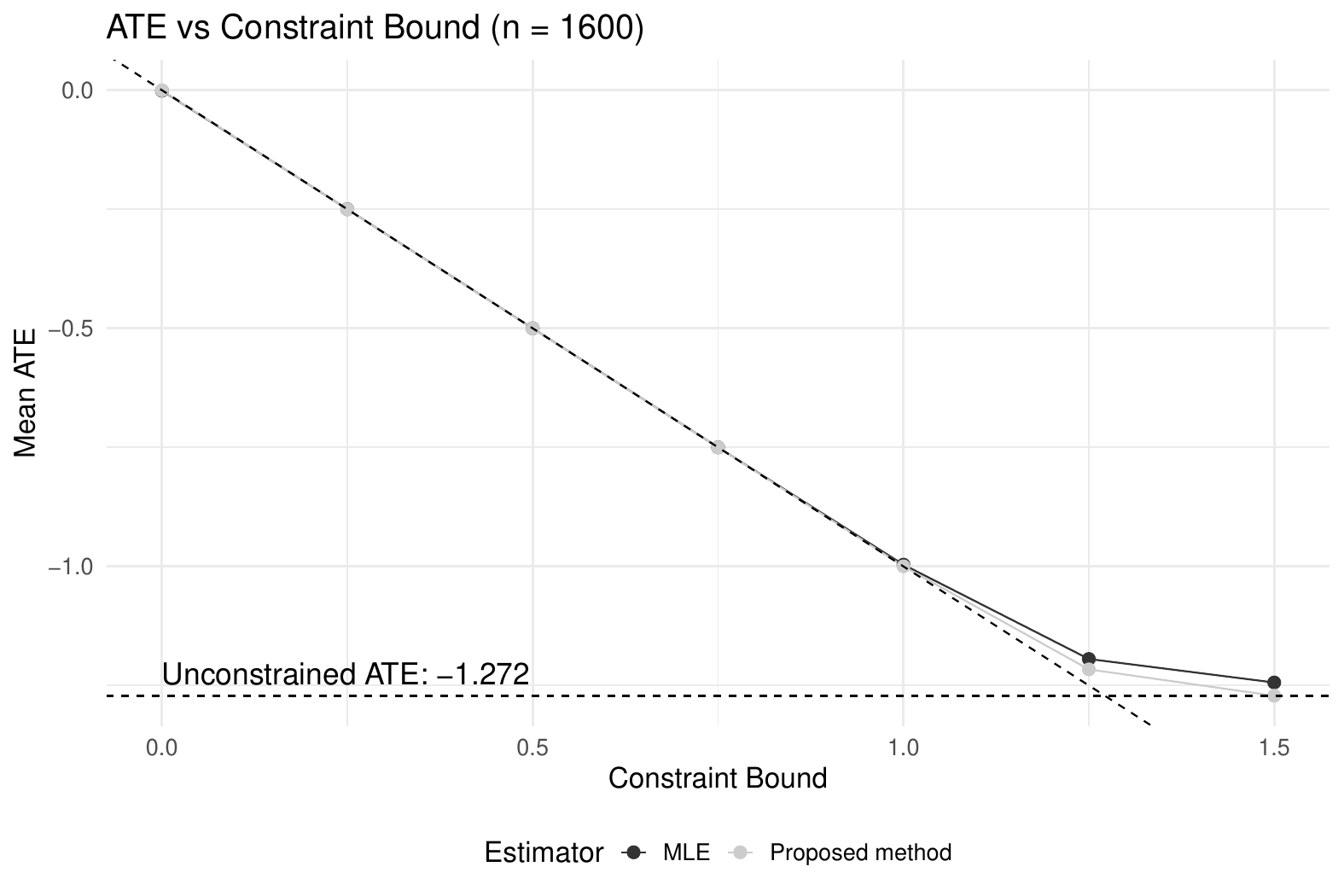}
    \caption{\textbf{Comparison of proposed method vs. constrained MLE.} Average risk (left) and constraint (right) of $\psi_{n,\lambda_n}$ (gray) vs. constrained MLE (black) at two sample sizes. The horizontal dashed line displays the risk of  the unconstrained estimate of $\psi_0$.}
    \label{fig:sim_comp_to_mle}
\end{figure}

Additional simulations in Appendix~S12 explore the impact of misspecification of working models for nuisance parameters. In general, we find that often the constraint remains well controlled, but optimal risk is not achieved under inconsistent nuisance parameter estimation. We also explore the extent to which other estimators of causal effects (e.g., inverse probability weighting and doulby-/ multiply robust estimators) can be leveraged to improve estimation. In general, our results show that robust estimators can better control the constraint value under misspecification, while singly robust, inverse weighted estimators tend to have worse predictive performance.


\subsection{Equalized risk in the cases}

In this setting, given $(X,W)$, $Y$ was generated from a $\mbox{Bern}(\psi_0(X,W))$ distribution, where $\psi_0(X,W) = \mbox{expit}(-X/2 - W_1 - W_2 - W_3 + 2W_5)$. For this data generating distribution, the true difference in risk in the cases is approximately $\Theta_{P_0}(\psi_0) = 0.12$. We implemented our estimator as described in Section~\ref{subsec:erica}. The estimate of $\lambda$ in \eqref{eq:emp_mini_lambda_est} was found using a grid search over 1$e$5 equally spaced values of $\lambda$.

\begin{figure}
\centering
\includegraphics[width=6.5cm, height=5.5cm]{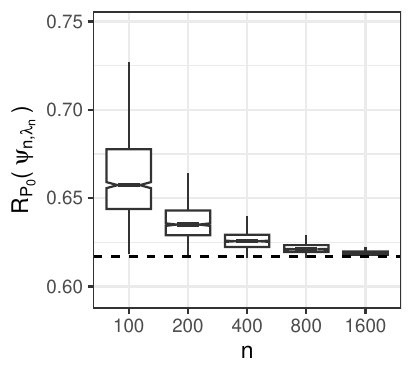} 
\includegraphics[width=6.5cm, height=5.5cm]{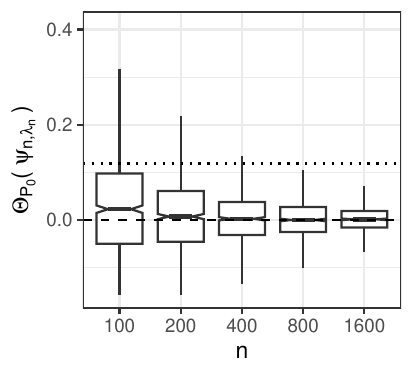}
\includegraphics[width=6.5cm, height=5.5cm]{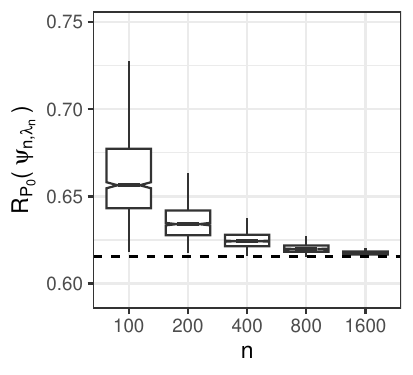}
\includegraphics[width=6.5cm, height=5.5cm]{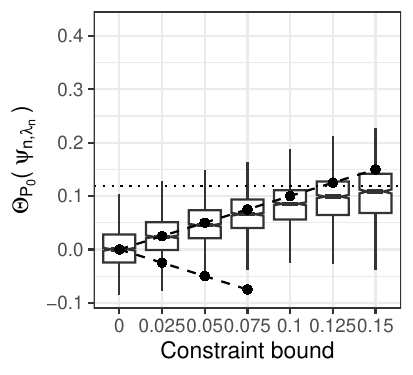}
\caption{\textbf{Equalized risk in the cases constraint and cross entropy risk.} \underline{Top left:} Distribution of risk of $\psi_{n,\lambda_n}$ over 1000 realizations for each sample size for the equality constraint $\Theta_{P_0}(\psi) = 0$. The dashed line indicates the optimal risk $R_{P_0}(\psi_0^*)$.  \underline{Top right:} Distribution of the true constraint over 1000 realizations for each sample size. The dashed line indicates the equality constraint value of zero. The constraint value under the unconstrained $\psi_0$, $\Theta_{P_0}(\psi_0) = 0.12$, is shown with a dotted line. \underline{Bottom left:} Distribution of risk of $\psi_{n,\lambda_n}$ over 1000 realizations for each sample size for the inequality constraint $|\Theta_{P_0}(\psi)| \le 0.05$. The dashed line indicates the optimal risk $R_{P_0}(\psi_0^*)$ under this constraint. \underline{Bottom right:} Distribution of the true constraint for estimators built using the equality constraint (constraint bound = 0) and inequality constraints with varying bounds at $n = 800$. The dotted line shows the value of the constraint under $\psi_0$, $\Theta_{P_0}(\psi_0)$. The dashed lines shows the positive and negative bounds on the constraint.}
\label{fig:erica_sim}
\end{figure}

We found that the risk of our proposed estimator approached the optimal risk and the constraint was approximately solved in large samples (Figure \ref{fig:erica_sim}). In this setting, there was more variability in terms of control of the constraint in small sample sizes. In some instances the true constraint was worse than the true value of the constraint for the unconstrained $\psi_0$ (dotted line, top right panel). We also saw this variability in constraint across different constraint bounds (bottom right panel). However, with sufficiently large samples this variability diminished and the optimal risk was achieved while appropriately controlling the constraint.

\subsection{Other machine learning algorithms} 
\label{sims:ml}

We also evaluated our proposal using more data-driven approaches to estimation of $\psi_0$ and other functional parameters. We first considered a setting with high-dimensional covariates, where penalization is necessary to achieve adequate performance of linear models. In this simulation, we separately considered vectors $W$ consisting of 20, 50, and 100 independent standard Normal covariates. We used the LASSO \citep{tibshirani1996regression} to estimate functional parameters under the ATE and NDE equality constraints at the same sample sizes as in previous simulations. Details of the data generating process can be found in Appendix~S12. 
We present results for the ATE constraint here; results for the NDE constraint were similar. 

\begin{figure}[!t]
\centering
\includegraphics[scale=0.75]{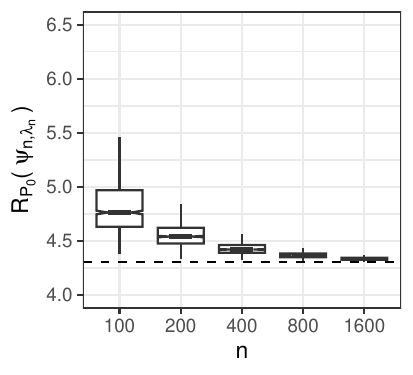}
\includegraphics[scale=0.75]{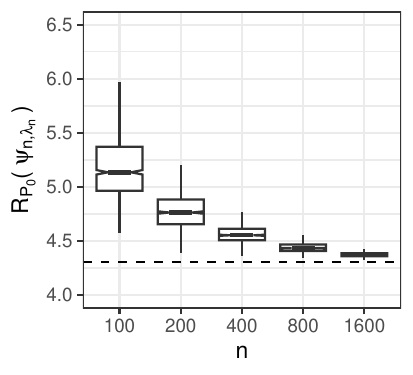}
\includegraphics[scale=0.75]{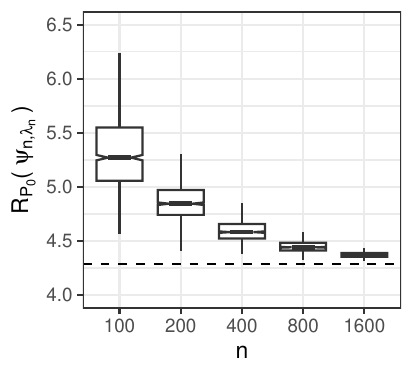}
\includegraphics[scale=0.75]{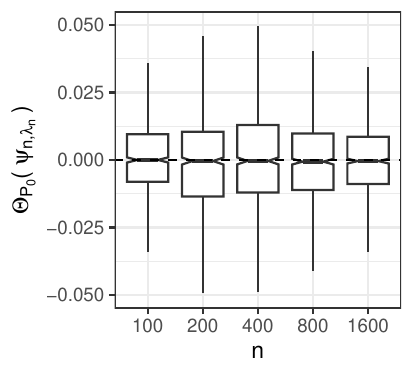}
\includegraphics[scale=0.75]{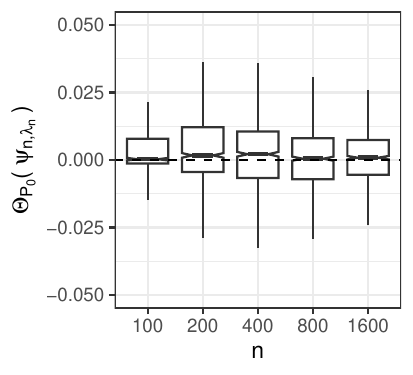}
\includegraphics[scale=0.75]{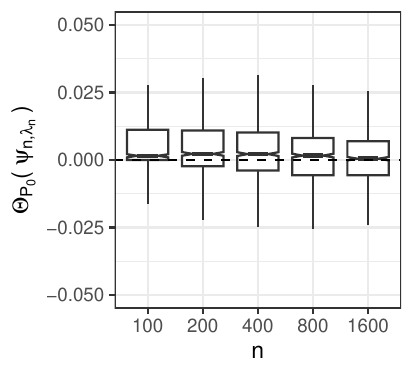}
\caption{\underline{Top row:} Performance of the proposed approach for the average treatment effect constraint and mean squared error in the presence of high-dimensional covariates. Left to right: Distributions of risk of $\psi_{n,\lambda_n}$ over 1000 realizations for each sample size for 10 (left), 50 (middle), 100 (right) covariates, respectively. The dashed line indicates the optimal risk $R_{P_0}(\psi_{0,\lambda_0})$. \underline{Bottom row:} Performance of the proposed approach for the average treatment effect constraint in the presence of high-dimensional covariates. Left to right: Distributions of the constraint $\Theta_{P_0}(\psi_{n,\lambda_n})$ over 1000 realizations for each sample size for 10 (left), 50 (center), 100 (right) covariates, respectively. The dashed line indicates the desired constraint of zero. Under the unconstrained $\psi_0$, $\Theta_{P_0}(\psi_0) = 0.82$ and is not shown due to the figure's scale.}
\label{fig:highdim_sim_risks}
\end{figure}


Our procedure behaved as expected across all settings, although more covariates slowed convergence to optimal risk (Figure \ref{fig:highdim_sim_risks}, top row); nonetheless, the constraint was approximately satisfied at every sample size (bottom row).

We ran a comparable simulation with the highly adaptive LASSO (HAL) \citep{benkeser2016highly, van2017generally} on a univariate covariate $W$ to assess how the variation norm of $\psi_0$ affects our results. 
As HAL’s convergence rate slows with larger variation norms, we expected—and observed—slower convergence of the risk of $\psi_{n,\lambda_n}$ to the optimal risk, although our method still approached optimal risk in moderate samples even at high variation norms. See Appendix~S12 for details.

\section{Data application}
\label{sec:data}


We used the Adult census dataset (48,842 records) to predict the binary outcome $Y=\I({\text{income}> 50\mathrm{k}}$ based on demographic and occupational covariates. Let $X=$ sex, $W=$ {race, native country}, and $M=$ {occupation, work class}. We imposed the constraint that the natural direct effect of sex on prediction lies below a specified threshold. After restricting to complete cases we retained $n=44,869$ observations. 


We generated 100 random splits (70\% train, 30\% validation) and used parametric models with AIPW to estimate the unconstrained NDE at 0.161, indicating a substantial direct effect of sex on income earnings that is not mediated through sex differences in type of work.  We then trained income‐prediction algorithms constraining the NDE to 0 and to thresholds of 0.05, 0.10, 0.15, 0.20, and 0.25. 
We used either constrained maximum likelihood estimation in a logistic regression model for the outcome regression or our proposed methodology for developing a nonparametric constrained algorithm. The constrained maximum likelihood utilized the Sequential Least Squares Quadratic Programming (SLSQP) method as implemented in the NLOPT optimization library \citep{kraft1994algorithm,nloptr}. Our proposed methodology utilized working logistic regression models for all nuisance parameters. We compared the validation cross-entropy, area under the receiver operating characteristics curve (AUC), and the value of the constraint for each method (estimated in the validation sample using AIPW).

\begin{figure}[t]
  \centering
  \begin{subfigure}[b]{0.32\textwidth}
    \includegraphics[width=\textwidth]{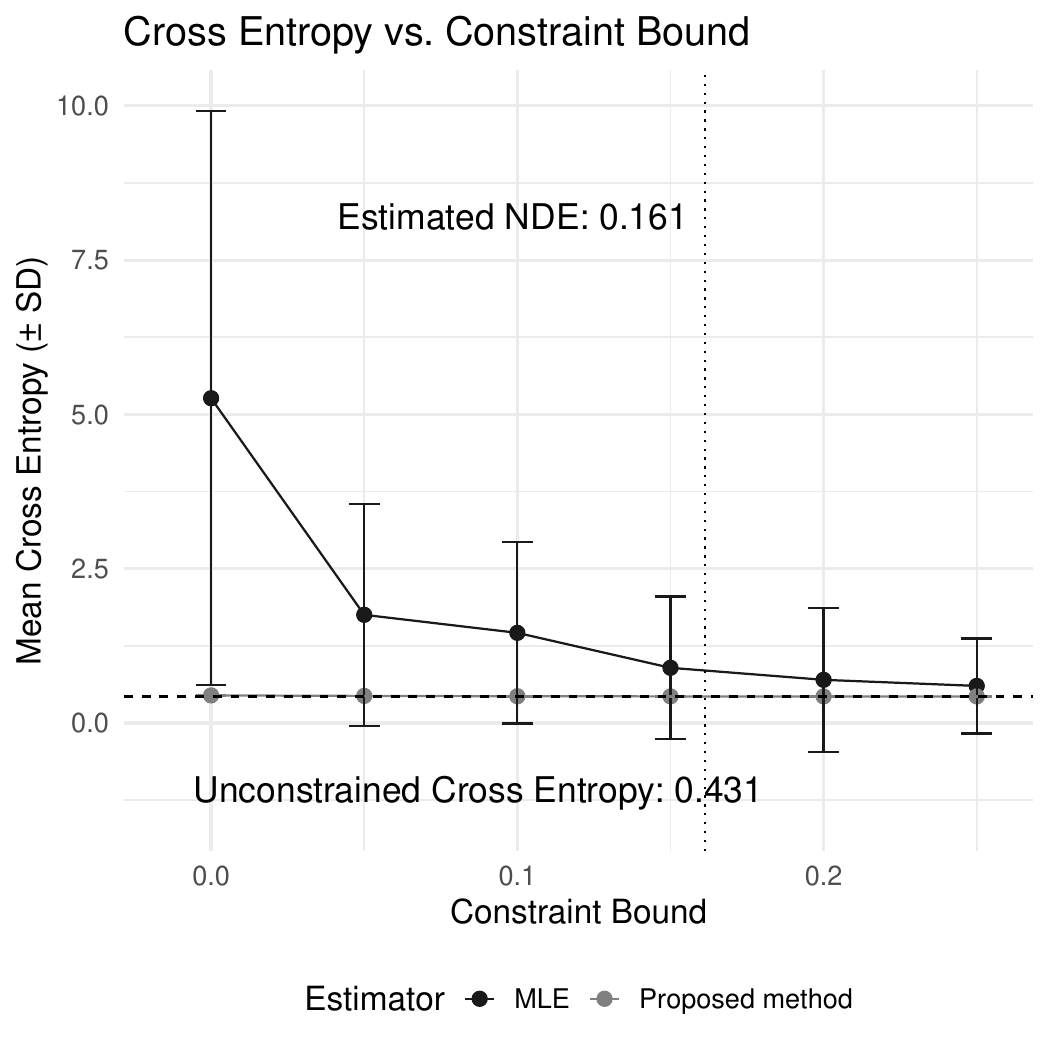}
  \end{subfigure}
  \hfill
  \begin{subfigure}[b]{0.32\textwidth}
    \includegraphics[width=\textwidth]{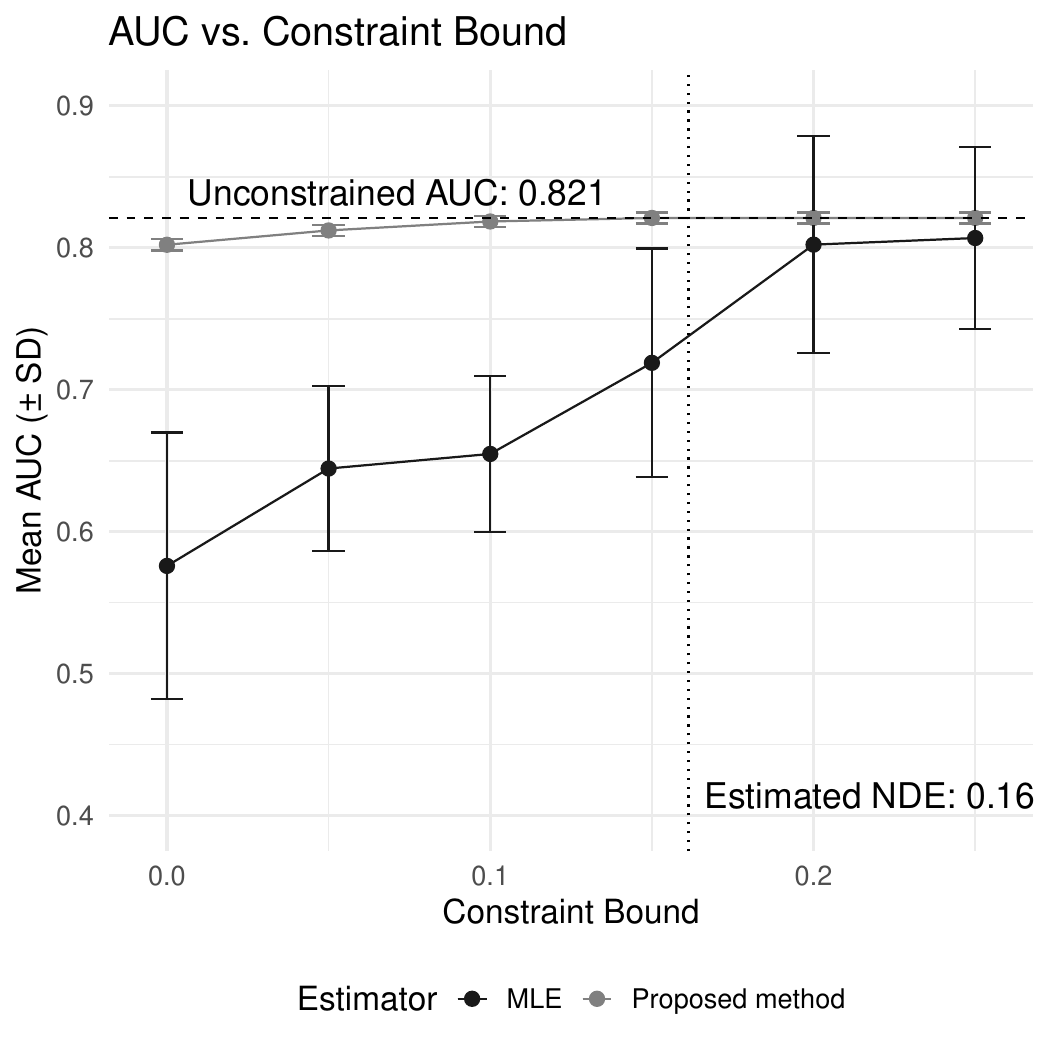}
  \end{subfigure}
  \hfill
  \begin{subfigure}[b]{0.32\textwidth}
    \includegraphics[width=\textwidth]{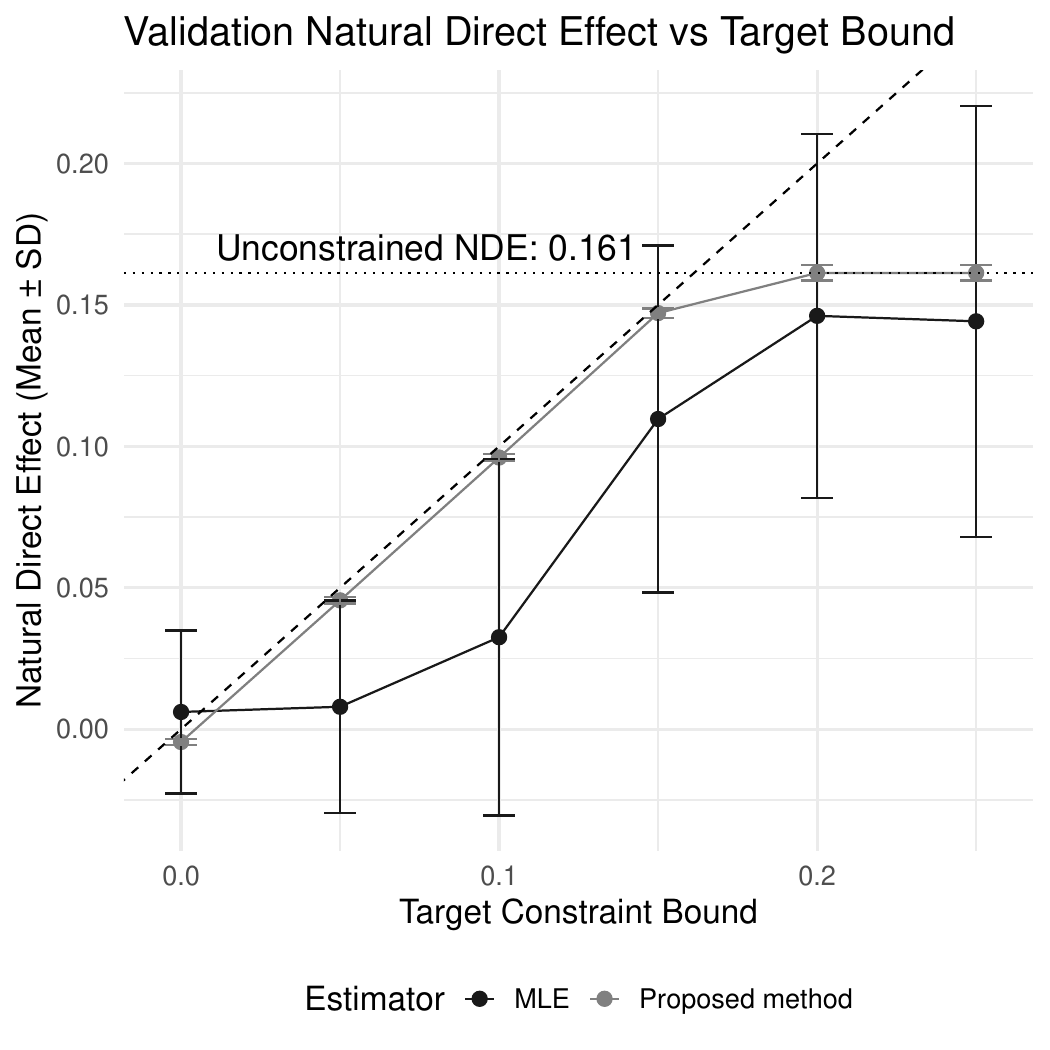}
  \end{subfigure}
  \caption{Comparison of the average validation sample performance of constrained maximum likelihood estimator versus proposed method.}
  \label{fig:adult_results}
\end{figure}

The maximum likelihood estimator had highly variable performance across repeated validation samples, particularly when trying to tightly control the NDE near 0 (Figure \ref{fig:adult_results}). Not only was the result more variable, but also tended to have considerably worse predictive performance when compared to the proposed method. On the other hand, the proposed method achieves nearly the same performance as the unconstrained estimator, even for tight control of the NDE. Moreover, we see that the proposed method is able to appropriately control the bound when it is necessary and for constraint bounds larger than the estimated effect, the proposed method appropriately has the same behavior as the unconstrained model. Overall, these results demonstrate an empirical application where the proposed method demonstrates superior stability and performance over a constrained maximum likelihood method. These results are further substantiated by an additional simulation study with a data generating process mimicking the structure of the Adult data set (Appendix~S12).

We also analyzed the COMPAS data comparing our constrained learning method to other methods for predicting recidivism from demographic data. We again found improvements in predictive performance and constraint control relative to other approaches, though the differences were more modest than in the Adult data (see Appendix~S13). 

\section{Discussion}
\label{sec:disc}

This work lays the foundation for constrained estimation of function-valued parameters in infinite-dimensional models, with applications in fairness, external validity, and risk control. Several directions for future research emerge naturally from our framework. First, the current approach relies on pathwise differentiability to define constraint-specific paths and apply the Riesz representation theorem. In practice, constraints may be nondifferentiable or discontinuous, such as threshold-based fairness metrics or discrete decision rules. Extending the framework to accommodate such settings, for example through smoothing techniques or surrogate losses, is a promising avenue. While our results provide asymptotic guarantees for penalized risk minimization and constraint satisfaction, it would also be valuable to study finite-sample performance and guarantees. In particular, non-asymptotic regret bounds could help quantify the gap between the estimated and oracle constrained solutions in decision-focused applications. In addition, a deeper theoretical analysis of the constraint-specific path may offer further insight. Studying its geometric properties, including conditions for uniqueness, smoothness, and connections to canonical least favorable submodels, could clarify when the path admits closed-form representations, when it aligns with universal least favorable directions, and how its structure influences estimation stability under misspecification. We view these directions as important steps toward a more comprehensive theory of constrained statistical learning in high-dimensional and policy-relevant settings.



\vspace{0.75cm}
\begingroup
\renewcommand{\baselinestretch}{0.92}
\selectfont  
\setlength{\bibsep}{10pt}    
\bibliographystyle{abbrvnat}
\bibliography{refs}
\endgroup


\pagebreak 

\appendix 

\begin{center}
	 \textbf{\Large Supplementary Materials}
\end{center}

\vspace{0.5cm}
The supplementary materials are organized as follows. 

\vspace{0.2cm}
\noindent Appendix~\ref{app:notation} provides a summary of the notations utilized throughout the manuscript, facilitating understanding and reference. 

\vspace{0.2cm}
\noindent Appendix~\ref{app:details} contains additional details on the notion of pathwise differentiability \eqref{app:pathwise_differentiability}, discussions on second-order conditions to ascertain whether the constrained minimizer indeed represents a valid minimum \eqref{app:second-order}, and an extension of our methodology to incorporate multiple equality and inequality constraints \eqref{app:methods-multi}. 

\vspace{0.2cm}
\noindent Appendix~\ref{app:proofs} contains all the proofs. 

\vspace{0.2cm}
\noindent Appendix~\ref{app:related_work} provides deeper insights into fairness notions, constraint formulations, and related methodological perspectives that complement our main results. 

\vspace{0.2cm}
\noindent Appendices~\ref{app:ate}, \ref{app:nde}, and \ref{app:er_cases} offer detailed derivations of the canonical gradients, analyses of second-order conditions, and examinations of the large-sample properties of the proposed estimators for the three examples discussed in the main manuscript---average total effect, natural direct effect, and equalized risk among cases, respectively.  

\vspace{0.2cm}
\noindent Appendices~\ref{app:cate}, \ref{app:er_overall}, \ref{app:moment_rest}, \ref{app:misspecified} provide supplementary examples extending our methodology to additional constraints. Appendix \ref{app:cate} covers causal constraints (conditional average treatment effect, effect on the treated and effect on controls). Appendix \ref{app:er_overall} presents overall equalized risk and two-dimensional equalized risk stratified by outcome (cases and controls). Appendix \ref{app:moment_rest} develops moment based constraints in density estimation. Appendix \ref{app:misspecified} addresses bias corrected estimation under model misspecification.

\vspace{0.2cm}
\noindent Appendix~\ref{app:sims_details} presents additional simulation details and results.  

\vspace{0.2cm}
\noindent Appendix~\ref{app:real_data} contains additional real data analysis. 

\pagebreak 
\section{Glossary of terms and notations} 
\label{app:notation} 

For ease of navigation through notations used in the manuscript, we provide a comprehensive list in Table~\ref{tab:notations}. 

\begin{table}[!h]
	\begin{center}
		\caption{\centering Glossary of terms and notations}
		\label{tab:notations}
		\addtolength{\tabcolsep}{0pt}
		{\small
			\begin{tabular}{ ll | ll} 
				\hline  
				\textbf{Symbol} & \textbf{Definition} & \textbf{Symbol} & \textbf{Definition}  
				\\ \hline 
				$X$  & Sensitive/treatment variable  &  $Y$  & Outcome variable    \\ 
				$W$ & Baseline covariates  & $M$ & Mediator variables  \\ 
				$O$ & Observed variables   & $Z$ & $O \setminus Y$  \\ 
				$ \mathcal{M}$ & Statistical model & $P_0 \in \mathcal{M}$ & True data distribution \\ 
				$\psi \equiv \Psi(P)$ & Risk minimizer under $P \in \mathcal{M}$  & $\bf \Psi$ & Parameter space \\ 
				$L(\psi)$ & Loss function & $R_P(\psi)$ & Risk function, $\int L(\psi)(o)dP(o)$ \\
				$\Theta_P(\psi)$ & Equality constraint & $\Omega_P(\psi)$ & Inequality constraint \\
				$\psi^*_0 \equiv \psi^*(P_0)$ & Constrained risk minimizer & $\lambda, \mu$ & Lagrange multipliers \\
				$\{\psi_{\delta, h}: \delta \in \openr\}$ & \multicolumn{3}{l}{Path through $\psi$ at $\delta = 0$ with direction $h$, where $h = \{d\psi_{\delta, h}/d\delta\}|_{\delta =0 }$} \\
				$\mathcal{H}_P(\psi)$ & Collection of paths & $\hilbert_P(\psi),\langle.,\rangle $ & Hilbert space, inner product \\
				$\tangent_P(\psi)$ & \multicolumn{3}{l}{Closure in $\hilbert_P(\psi)$ of linear span of  $\mathcal{H}_P(\psi)$} \\
				$L^2(P_Z)$ & \multicolumn{3}{l}{Space of functions with finite 2nd moment on the support of $Z$ under $P$} \\
				\multicolumn{4}{l}{$D_{R, P}(\psi), D_{\Theta, P}(\psi), D_{\Omega, P}(\psi) \quad $ Gradients of risk, equality constraint, inequality constraint} \\
				\multicolumn{4}{l}{$P_{0, M|X,W}(m \mid x, w) \equiv f_{0, M}(m \mid x, w) \quad P_0(M = m \mid X = x, W = w)$} \\
				$P_{0, W}$ & Marginal distribution of $W$ & $\pi_0(x \mid w)$ & $P_0(X = x \mid W = w)$ \\
				$\gamma_0(m \mid x, w)$ & $f_{0, M}(m \mid 0, w)/f_{0, M}(m \mid x, w)$ & $\alpha_0(x \mid w, m)$ & $P_0(X = x \mid W = w, M = m)$ \\ 
				$p_{0,1}(x)$ & $P_0(X = x, Y=1)$ & $p_{0}(x)$ & $P_0(x)$ \\
				$\eta_0$ & Collection of nuisances & $\eta_n$ & Estimates of nuisances \\
				$P_0 f$ & $\int f(o) dP_0(o)$ & $P_n f$ & $\frac{1}{n} \sum_{i=1}^n f(O_i) $ \\ 
				$|| f ||_1$ & $L^1(P_0)$-norm of $f$, $P_0(|f|)$ & $|| f ||_2$ & $L^2(P_0)$-norm of $f$, $P_0(f^2)$  
			\end{tabular}
		}
	\end{center}
\end{table}

\section{Additional methodological details and extensions}
\label{app:details}

\subsection{Pathwise differentiability and links to efficiency theory}
\label{app:pathwise_differentiability}

Assumption \eqref{assmp:pathwise_diff} is similar to the assumption of \emph{pathwise differentiability} that is commonly invoked in nonparametric and semiparametric efficiency theory (see e.g., \cite{van1995efficient}, Definition 1.6). In these applications, it is common to consider paths $\{ P_{\delta,h} : \delta \in \mathbb{R} \}$ through $P \in \mathcal{M}$ where $h \in L_0^2(P)$, the Hilbert space of real-valued functions defined on the support of $P$ with mean zero and finite second moment endowed with inner product $\langle f, g \rangle = \int f(o)g(o)dP(o)$. In these settings pathwise differentiability of a given parameter $\Gamma: \mathcal{M} \rightarrow \mathbb{R}$ is typically defined by the existence of a continuous linear map $\dot{\Gamma}_P: L_0^2(P) \rightarrow \mathbb{R}$ that can be used to represent the pathwise derivative, i.e., we say $\Gamma$ is pathwise differentiable if $\frac{d}{d\delta}\Gamma(P_{\delta,h})|_{\delta = 0} = \dot{\Gamma}_P(h)$.

The spirit of assumption (\ref{assmp:pathwise_diff}) is the same as this classic definition of pathwise differentiability: the assumption requires functionals that are \emph{smooth enough} for their derivatives to be well behaved. However, the specific details in terms of constructing the respective Hilbert spaces differ across the two settings. For example, we consider paths through $\psi$, a functional of $P$ rather than through $P$ itself. Our notation indexing vector spaces by $P$, e.g., $\hilbert_P(\psi)$, appropriately reflects this distinction. In spite of the subtle distinctions, it should still be considered appropriate to state that assumption (\ref{assmp:pathwise_diff}) is one requiring pathwise differentiability of $R_P$ and $\Theta_P$.

\subsection{Invertible operators}
\label{app:invertible_operators}

The constraint-specific path holds for both $\lambda$ and $\lambda + \delta$. That is:  
\begin{align*}
	D_{R, P_0}(\psi_{0, \lambda}) + \lambda D_{\Theta, P_0}(\psi_{0, \lambda}) &= 0  \ , 
	\\
	D_{R, P_0}(\psi_{0, \lambda + \delta }) + (\lambda + \delta) D_{\Theta, P_0}(\psi_{0, \lambda + \delta}) &= 0  \ .  
\end{align*}%
The difference yields: 
\begin{align*}
	\frac{ \Big\{ D_{R, P_0}(\psi_{0, \lambda + \delta }) + \lambda D_{\Theta, P_0}(\psi_{0, \lambda + \delta}) \Big\} - \Big\{ D_{R, P_0}(\psi_{0, \lambda}) + \lambda D_{\Theta, P_0}(\psi_{0, \lambda}) \Big\} }{\delta} = - D_{\Theta, P_0}(\psi_{0, \lambda + \delta}) \ , 
\end{align*}
which simplifies to: 
\begin{align*}
	\frac{ D_{R, P_0}(\psi_{0, \lambda + \delta }) - D_{R, P_0}(\psi_{0, \lambda })}{\delta} +  \lambda  \frac{ D_{\Theta, P_0}(\psi_{0, \lambda + \delta}) - D_{\Theta, P_0}(\psi_{0, \lambda})  }{\delta} = - D_{\Theta, P_0}(\psi_{0, \lambda + \delta}) \ . 
\end{align*}%
Taking the limit as $\delta \rightarrow 0$ results in: 
\begin{align*}
	\frac{d}{d \lambda} D_{R, P_0}(\psi_{0, \lambda})  +  \lambda \frac{d}{d \lambda} D_{\Theta, P_0}(\psi_{0, \lambda})  = - D_{\Theta, P_0}(\psi_{0, \lambda}) \ , 
\end{align*}%
which, by the chain rule of differentiation, is equivalent to: 
\begin{align}
	\frac{d}{d \psi_{0, \lambda}} D_{R, P_0}(\psi_{0, \lambda}) \frac{d \psi_{0, \lambda}}{d \lambda}  +  \lambda \frac{d}{d \psi_{0, \lambda}}  D_{\Theta, P_0}(\psi_{0, \lambda}) \frac{d \psi_{0, \lambda}}{d \lambda}  = - D_{\Theta, P_0}(\psi_{0, \lambda}) \ .
	\label{eq:differentiation}
\end{align}%
Here, $\frac{d}{d \psi_{0, \lambda}}$ represents the functional derivative with respect to $\psi_{0, \lambda}$, and $\frac{d \psi_{0, \lambda}}{d \lambda}$ represents the directional derivative of $\psi_{0, \lambda}$ in the direction of $\lambda.$ 

The left hand side of \eqref{eq:differentiation} is simply the functional derivative operator acting on $\frac{d \psi_{0, \lambda}}{d \lambda}.$ To solve for $\frac{d \psi_{0, \lambda}}{d \lambda}$, we need to invert that operator. We define the operator $\dot{D}_{R + \lambda \Theta}(\psi)(h)$ to be
\begin{align*}
	\dot{D}_{R + \lambda \Theta}(\psi)(h) \coloneqq \frac{d}{d \psi} D_{R, P_0}(\psi) h +  \lambda \frac{d}{d \psi} D_{\Theta, P_0}(\psi) h \ . 
\end{align*}

Assuming this operator is invertible, we can solve for $\frac{d \psi_{0, \lambda}}{d \lambda}$ as follows: 
\begin{align*}
	\frac{d \psi_{0, \lambda}}{d \lambda} =  \{\dot{D}_{R + \lambda \Theta}(\psi_{0, \lambda}) \}^{-1}(- D_{\Theta, P_0}(\psi_{0, \lambda}) ) \ . 
\end{align*} 

\subsection{Second-order conditions}
\label{app:second-order}

Assume $\Theta_P(\psi) = 0$ defines a non-empty feasible region in the parameter space $\bf \Psi$. Given Assumption~\eqref{assmp:pathwise_diff}, let $\psi^*_0 = \psi_{0, \lambda_0}$ denote the functional parameter which fulfills the ``first-order'' condition as outlined in \eqref{eq:pathcondition-a}, with $\lambda_0$ selected to ensure $\Theta_{P_0}(\psi_{0, \lambda_0}) = 0.$ Typically, $\psi^*_0$ is recognized as an extremum point, which could be either a minimum or a maximum. To determine if $\psi^*_0$ represents a minimum, the following arguments on second-order conditions are considered. 

Assume $X$ represents a sensitive characteristic with $K$ categories, and let $Z = O \setminus X$. We can decompose the functional parameter $\psi(O)$ into $K$ distinct components, where each component corresponds to units with a specific level of $X$. For $i = 0, \ldots, K-1$, we denote the component by $\psi^{i}(Z)$. Similarly, we can partition $\psi^*_0(Z)$ into $\psi^{0, *}_0(Z), \ldots, \psi^{K-1, *}_0(Z)$. For a given data point $z$, our objective is to check whether the vector $(\psi^{0, *}_0(z), \ldots, \psi^{K-1, *}_0(z))$ is a minimum point. For this objective, we can examine the sufficient second-order conditions which involve constructing the bordered Hessian matrix; i.e., a modified version of the Hessian that incorporates the constraints of the problem via the Lagrange multiplier(s). For a given $\psi^i \in \Psi (i = 0, \ldots, K - 1)$ and $\lambda \in \openr$, let 
\begin{itemize}
	\setlength{\itemsep}{0.25cm}
	
	\item \ $\calL_P(\psi^i, \lambda) \coloneqq R_P(\psi) + \lambda \Theta_P(\psi)\big|_{X=i}$ define the Lagrangian function evaluated at $X=i$, 
	
	\item \ $\dot{\calL}_P(\psi^i, \lambda)  = \frac{\partial}{\partial \delta} \calL_P(\psi_{\delta, h}, \lambda)\big|_{\delta=0, X=i}$ define the  the canonical function evaluated at $X=i$. In other words, $\dot{\calL}_P(\psi^i, \lambda) \coloneqq D_{R, P}(\psi^i) + \lambda D_{\Theta, P}(\psi^i)$, where $D_{R, P}(\psi^i) = D_{R, P}(\psi)\big|_{X=i}$. 
	
	\item \ $\ddot{\calL}_P(\psi^i, \lambda) = \frac{\partial}{\partial \delta}\dot{\calL}_P(\psi_{\delta, h}, \lambda)\big|_{\delta = 0, X=i}$ define the Hessian function evaluated at $X=i$. In other words, $\ddot{\calL}_P(\psi^i, \lambda) \coloneqq \dot{D}_{R, P}(\psi^i) + \lambda \dot{D}_{\Theta, P}(\psi^i)$, where $\dot{D}_{R, P}(\psi^i) = \frac{\partial}{\partial \delta}D_{R, P}(\psi^i_{\delta, h})\big|_{\delta = 0}$. 
\end{itemize}

For a given $\psi$ and $\lambda$, the bordered Hessian evaluated at data point $z$ is a $(K + d) \times (K + d)$ matrix, denoted by $\bar{H}_z(\psi, \lambda)$ and defined as follows: 
\begin{align}
	\bar{H}_z(\psi(z), \lambda) 
	&= \begin{bmatrix}
		0 & \bar{D}_{\Theta, P}(\psi)(z)
		\\ \\ 
		\bar{D}^T_{\Theta, P}(\psi)(z) & \bar{\ddot{\calL}}_P(\psi(z), \lambda)
	\end{bmatrix}  
	\ , \label{eq:borderedH}
\end{align}%
where $\frac{\partial^2 }{\partial \lambda^2}\calL_P(\psi, \lambda) = 0$, $\bar{D}_{\Theta, P}(\psi)$ denotes a vector of length $K$  with the $j$-th element being $D_{\Theta, P}(\psi^j)$, and $\bar{\ddot{\calL}}_P(\psi, \lambda)$ denotes a $K \times K$ matrix with the $ij$-th element ($i \not=j$) being $\frac{\partial}{\partial \delta} \Big\{ \frac{\partial }{\partial \delta}\calL_P(\psi_{\delta, h})\big|_{X=i} \Big\} \Big|_{\delta=0, X=j}$, and the $i$-th element on the diagonal being $\ddot{\calL}_P(\psi^i, \lambda)$. 

To determine whether $\psi^{*}_0 \equiv \psi_{0, \lambda_0}$ is a minimizer, we then need to examine the leading principal minors of the bordered Hessian evaluated at $(\psi^{0, *}_0, \ldots, \psi^{K - 1, *}_0)$. For simplicity, we consider the sufficient conditions required for our provided examples in Section~\ref{subsec:examples}, where $K=2$. In these scenarios, the bordered Hessian in \eqref{eq:borderedH} simplifies to:  
\begin{align}
	\bar{H}_z(\psi^0(z), \psi^1(z), \lambda) = 
	\begin{bmatrix}
		0 &  D_{\Theta, P}(\psi^0)(z) &  D_{\Theta, P}(\psi^1)(z) 
		\\ \\ 
		D_{\Theta, P}(\psi^0)(z)  & \ddot{\calL}_P(\psi^0(z), \lambda) & 0 
		\\ \\ 
		D_{\Theta, P}(\psi^1)(z) & 0 & \ddot{\calL}_P(\psi^1(z), \lambda)
	\end{bmatrix} 
	\ . 
	\label{eq:borderedH_single}
\end{align}
Then to have a  minimum, it suffices that the determinant of the $3 \times 3$ bordered Hessian, evaluated at $(\psi^{0, *}_0, \psi^{1, *}_0)$ be negative. That is:
\begin{align}
	- D^2_{\Theta, P}(\psi^{1, *}_0) \times \ddot{\calL}_P(\psi^{0, *}_0, \lambda)
	- D^2_{\Theta, P}(\psi^{0, *}_0) \times \ddot{\calL}_P(\psi^{1, *}_0, \lambda) < 0 \ . 
	\label{eq:borderedH_det_single}
\end{align}

In order for \eqref{eq:borderedH_det_single} to hold, it is sufficient for the Hessian function $\ddot{\calL}_P(\psi^x, \lambda)$ to be positive. This positivity can be linked to the convexity of both the risk and constraint functions, as well as the sign of the Lagrange multiplier. A risk function $R_P(\psi)$ is termed convex if, for any $\psi_1, \psi_2 \in {\bf \Psi}$ and for any $\alpha$ within the closed unit interval, $R_P(\alpha \psi_1 + (1-\alpha) \psi_2) \leq \alpha R_P(\psi_1) + (1-\alpha) R_P(\psi_2)$. The risk function is considered strictly convex when this inequality is strict. Convexity of the constraint is defined in a similar manner. When the parameter space $\bf \Psi$, the risk function $R_P(\psi)$, and the constraint $\Theta_P(\psi)$ are convex, and with a positive $\lambda$, it is assured that the extremum point will be a minimum.

\subsection{General multi-dimensional constraints}
\label{app:methods-multi}

Suppose there are $d$ equality constraints, denoted by $\Theta^{(j)}_P(\psi) = 0$ for $j= 1, \ldots, d$, along with $m$ inequality constraints, denoted by $\Omega^{(i)}_P(\psi) \leq 0, \ i = 1, \ldots, m$. Let $\Theta_P(\psi)$ collect all the equality constraints and $\Omega_P(\psi)$ for all the inequality constraints. The task is then to determine the optimal functional parameter, $\Psi^*(P)$, defined by minimizing $R_P(\psi)$ over $\psi$ in the parameter space ${\bf \Psi}$, subject to these equality and inequality constraints:
\begin{equation}\label{eq:supp_target_general}
	\begin{aligned}
		&\hspace{1.85cm} \Psi^*(P) = \argmin_{\psi \in {\bf \Psi}} \
		R_P(\psi) \\
		\text{subject to:} \hspace{0.5cm}
		&\Theta^{(j)}_P(\psi) = 0, \ j = 1, \ldots, d \ , \ \text{ and } \ \Omega^{(i)}_P(\psi) \leq 0, \ i = 1, \ldots, m \ . 
	\end{aligned}
\end{equation}

To extend our method from the one-dimensional constraint to accommodate multi-dimensional constraints, we assume that the element-wise derivatives of the constraints along paths $\{\psi_{\delta, h} : \delta \in \openr\}$ for $h \in \tangent_P(\psi)$ satisfy that 
\begin{align}
	&\hspace{2cm} h \mapsto \frac{d}{d\delta}
	\Theta^{(j)}_P(\psi_{\delta,h})\Big|_{\delta = 0} \quad  \text{and} \quad  h \mapsto \frac{d}{d\delta}
	\Omega^{(i)}_P(\psi_{\delta,h})\Big|_{\delta = 0}  \ , \tag{S-A5} 
	\label{assmp:pathwise_diff-multi} 
\end{align} 
are bounded linear functionals on $\tangent_P(\psi)$ for $j = 1, \ldots, d$ and $i = 1, \ldots, m$. 

For each equality constraint $\Theta^{(j)}_P$ and inequality constraint $\Omega^{(i)}_P$, we define the canonical gradient of their respective pathwise derivatives at $\delta = 0$ as $D^{(j)}_{\Theta, P}(\psi)$ and $D^{(i)}_{\Omega, P}(\psi)$. These gradients are aggregated into vector valued functions $D_{\Theta, P}(\psi)$ and $D_{\Omega, P}(\psi)$ for the equality and inequality constraints, respectively. We also assume the gradients $D_{\Omega,P}$ and $D_{\Theta, P}$ are non-degenerate at a solution $\psi_0^*$. That is 
\begin{align}
	D^{(j)}_{\Theta,P}(\psi_0^*)(o) \ \text{ and }   D^{(i)}_{\Omega,P}(\psi_0^*)(o) \quad  \text{are linearly independent} \ ,  \tag{S-A6}
\end{align}
for all $j = 1, \ldots, d$ and $i = 1, \ldots, m$, and for all $o$ in the support of $O$. This the general form of the linear independence constraint qualification for this problem \citep{boltyanski1998geometric, sundaram1996first, gould1971necessary}. 

Consider the solution $\psi_{0,	\lambda, \mu} = \argmin_{\psi \in {\bf \Psi}} R_{P_0}(\psi) + \lambda^{\top} \Theta_{P_0}(\psi) + \mu^{\top} \Omega_{P_0}(\psi)$, where $\Theta_{P_0}(\psi) \in \openr^d$, $\Omega_{P_0}(\psi) \in \openr^m$, $\lambda \in \openr^d$, and $\mu \in (\openr^{\geq 0})^m$. Similar to the one-dimensional constraint, $\{ \psi_{0, \lambda, \mu} : \lambda \in \openr^d, \mu \in (\openr^{\geq 0})^m \}$ defines a constraint-specific path through $\psi_0$ at $\lambda=[0]^d, \mu = [0]^m.$ 
Drawing from the rationale in the proof of Lemma~\ref{lem:lossmin}, finding the solution $\psi_{0, \lambda, \mu}$ for particular $\lambda$ and $\mu$ values allows us to find the optimal solution $\psi^*_0 \equiv \Psi^*(P_0)$ in \eqref{eq:supp_target_general}: it is the solution $\psi_{0, \lambda_0, \mu_0}$ where $\lambda_0$ and $\mu_0$ are determined by the conditions $\Theta_{P_0}(\psi_{0, \lambda, \mu}) = [0]^d$ and $\mu \Omega_{P_0}(\psi_{0, \lambda, \mu}) = [0]^m$, where $\mu \Omega_{P_0}(\psi_{0, \lambda, \mu})$ denotes pointwise multiplication of the elements of the vectors. This latter condition is often referred to in non-linear optimization as \textit{complementary slackness} conditions. Following similar logic preceding Theorem~\ref{thm:path-characterization}, the path $\{\psi_{0, \lambda, \mu} : \lambda \in \openr^d, \mu \in (\openr^{\geq 0})^m\}$ must satisfy 
\begin{align}
	D_{R, P_0}(\psi_{0, \lambda, \mu}) + \lambda^\top D_{\Theta, P_0}(\psi_{0, \lambda, \mu}) + \mu^\top D_{\Omega, P_0}(\psi_{0, \lambda, \mu}) = 0  \ , 
	\label{appeq:lfmpathconditionb} 
\end{align}%
for $\lambda \in \openr^d, \ \mu \in (\openr^{\geq 0})^m$. The above characterizes $\psi_{0, \lambda, \mu}$ as the solution. 

When a closed-form solution for $\psi_{0, \lambda, \mu}$, as described via \eqref{appeq:lfmpathconditionb}, is not available, we propose an alternative characterization of the constraint-specific path that allows for a recursive estimation strategy to be used; similar to \eqref{eq:pathcondition_b} for the one-dimensional constraint. Let $\dot{D}_{0, R + \lambda^\top \Theta + \mu^\top \Omega}(\psi) = \frac{d}{d \psi}  [D_{R, P_0}(\psi) + \lambda^\top D_{\Theta, P_0}(\psi) + \mu^\top D_{\Omega, P_0}(\psi)]$ and assume it is invertible. Mirroring the analysis applied in the one-dimensional case, the path defined by $\{\psi_{0, \lambda, \mu}: \lambda \in \mathbb{R}^d, \mu \in (\mathbb{R}^{\geq 0})^m\}$ is determined to satisfy the following differential equations:
\begin{equation}\label{multdiffeqn}
	\begin{aligned}
		\frac{d}{d \lambda_j} \psi_{0, \lambda, \mu} &= - \big\{ \dot{D}_{0, R + \lambda^\top
			\Theta + \mu^\top \Omega}(\psi_{0, \lambda, \mu})\big\}^{-1} D^{(j)}_{\Theta, P_0}(\psi_{0, \lambda, \mu}) \ , 
		\quad \text{for } j = 1, \ldots, d \ ,
		\\
		\frac{d}{d \mu_i} \psi_{0, \lambda, \mu} &= - \big\{ \dot{D}_{0, R + \lambda^\top
			\Theta + \mu^\top \Omega}(\psi_{0, \lambda, \mu})\big\}^{-1} D^{(i)}_{\Omega, P_0}(\psi_{0, \lambda, \mu}) \ , 
		\quad \text{for } i = 1, \ldots, m \ . 
	\end{aligned}
\end{equation}

\subsubsection{Estimation via \eqref{appeq:lfmpathconditionb}} 
Assume there exists a closed-form solution for \eqref{appeq:lfmpathconditionb} that writes as $\psi_{0, \lambda, \mu} = r(\eta_0, \lambda, \mu)$ for some mapping $r$, where $\eta_0 = \eta(P_0)$ denote the collection of nuisance parameters indexing the gradients $D_{R, P_0}, D_{\Theta, P_0},$ and $D_{\Omega, P_0}$. The plug-in estimator $\psi_{n, \lambda, \mu} = r(\eta_n, \lambda, \mu)$ can be used to estimate any $\psi_{0, \lambda, \mu}$ on the constraint-specific path. A plug-in estimate of $\psi_{0, \lambda_0, \mu_0}$ is given by $\psi_{n, \lambda_n, \mu_n}$, where $\lambda_n$ and $\mu_n$ are obtained either through a simple plug-in of their corresponding closed-form solutions or empirical minimizations solved via a (d+m)-dimensional grid search or other more sophisticated minimization approaches. 

We describe the grid search procedure in the following. Let $\Theta^{(j)}_n(\psi)$ and $\Omega^{(i)}_n(\psi)$ denote estimates of $\Theta^{(j)}_{P_0}(\psi)$ and $\Omega^{(i)}_{P_0}(\psi)$, respectively.  Before proceeding with the grid search, we first evaluate the complementary slackness condition outlined below. This involves initializing $\mu_n = [0]^m$ and searching for $\lambda^*_n$ that satisfies $\Theta_{n}(\psi_{n, \lambda^*_n, \mu_n=0}) = [0]^d$ and $\Omega_n(\psi_{n, \lambda^*_n, \mu_n=0}) \leq [0]^m$. If these conditions are met, the process terminates early; otherwise, the search continues in an iterative exclusion and search process described below: 
\begin{itemize}
	\item \ Begin by excluding one element at a time from the vector $\mu_{n}$, specifically $\mu_{i, n}$, while maintaining the rest at zero. We use $\mu_{\preceq i, n}$ to represent the set of elements in $\mu_n$ that have been excluded up to and including $\mu_{i, n}$. Conversely, $\mu_{\succ i, n}$ denotes the elements that have not been excluded.
	\item \ For each specific configuration identified by $\mu_{i, n}$, we aim to find the vectors $\lambda^*_n$ and $\mu^*_{\preceq i, n}$ that satisfy the condition: $\Omega^{(\succ i)}_n (\psi_{n, \lambda^*_n, \mu^*_{\preceq i, n}}) \leq [0]^{|\succ i|}$, and 
	\begin{align*}
		&\lambda^*_{n} = \arg\min_{\lambda_n \in [-\delta, \delta]^d} \Theta_n (\psi_{n, \lambda_n, \mu_{\preceq i, n}, \mu_{\succ i, n} = 0}) \ , \\
		&\mu^*_{\preceq i, n} = \arg\min_{\mu_{s} \in (0, \delta]} \Omega^{(\preceq i)}_n (\psi_{n, \lambda_n, \mu_{\preceq i, n}, \mu_{\succ i, n} = 0}) = [0]^{|\preceq i|} \ .
	\end{align*}
	Here, $|\preceq i|$ and $|\succ i|$ represent the number of elements excluded and included, respectively, in $\mu_n$ relative to $\mu_{i, n}$. The functions $\Omega^{(\preceq i)}_n$  and $\Omega^{(\succ i)}_n$ aggregate the constraints from $\Omega_n$ for the elements in $\mu_{\preceq i, n}$ and $\mu_{\succ i, n}$, respectively. 
	\item \ The search terminates if such vectors are identified; otherwise, it proceeds by excluding an additional element from $\mu$ and repeating the evaluation.
\end{itemize}

If the procedure has not terminated after exploring all possible combinations of exclusions from $\mu_n = [0]^m$, we seek to find the optimal estimates of $\lambda^*_{n}$ and $\mu^*_{n}$, such that 
\begin{align*}
	\lambda^*_{n}, \mu^*_n = \argmin_{\lambda \in [-\delta, \delta]^d, \mu \in (0, \delta]^m} |\Theta_n (\psi_{n, \lambda, \mu}) |  + | \Omega_n (\psi_{n, \lambda, \mu}) | \  . 
\end{align*}

\subsubsection{Recursive estimation via \eqref{multdiffeqn}}
Contrary to the one-dimensional constraint scenario, the construction of a recursive estimator for $\psi_{0, \lambda, \mu}$ via \eqref{multdiffeqn} is more involved. This complexity arises from the necessity to perform computations over  a (d+m)-dimensional grid to determine $\psi_{0, \lambda, \mu}$ across all values of $\lambda$ and $\mu$. 

As an example, suppose we have two equality constraints. Then a recursive solution for $\psi_{0, \lambda} = \psi_{0, \lambda_1, \lambda_2}$ using (\ref{multdiffeqn}) requires a 2-dimensional grid search and a quantification of $\psi_{0, \lambda_1 + d\nu, \lambda_2}$, $\psi_{0, \lambda_1, \lambda_2 + d\nu}$, and $\psi_{0, \lambda_1 + d\nu, \lambda_2 + d\nu}$. Let $r_{j}(\eta_0, \lambda) = - \big\{ \dot{D}_{0, R + \lambda^\top \Theta}(\psi_{0, \lambda})\big\}^{-1} D^{(j)}_{\Theta, P_0}(\psi_{0, \lambda})$, for $j = 1, 2$. The characterization in (\ref{multdiffeqn}) describes $\frac{d}{d \lambda_1} \psi_{0, \lambda}$ as function $r_1(\eta_0, \lambda)$ and $\frac{d}{d \lambda_2} \psi_{0, \lambda}$ as function $r_2(\eta_0, \lambda)$. It also implies $\frac{d^2}{d \lambda_1 d \lambda_2} \psi_{0, \lambda} = \frac{d}{d \lambda_2} r_1(\eta_0, \lambda)$, which can be represented as
{\small 
	\begin{align*}
		\frac{d^2}{d \lambda_1 d \lambda_2} \psi_{0, \lambda} 
		&= \left\{ \frac{d}{d \psi_{0, \lambda}} r_1(\eta_0, \lambda)  \right\} \times \left\{\frac{d}{d \lambda_2} \psi_{0, \lambda} \right\} = \left\{ \frac{d}{d \psi_{0, \lambda}} r_1(\eta_0, \lambda) \right\} \times r_2(\eta_0, \lambda) 
		\equiv r_{12}(\eta_0, \lambda) \ .
	\end{align*}
}

Equivalently, $r_{12}(\eta_0, \lambda)$ can be defined as $\left\{\frac{d}{d \psi_{0, \lambda}} r_2(\eta_0, \lambda) \right\} \times r_1(\eta_0, \lambda)$. 
We have now succeeded in expressing the two first-order derivatives of $\psi_{0,\lambda}$ w.r.t.~$\lambda_1$, $\lambda_2$ and the second-order derivative w.r.t.~both $\lambda_1, \lambda_2$. Using numerical derivatives, we can derive the following pieces required for the recursive estimation of $\psi_{0, \lambda}$ for all $\lambda$ in an arbitrary $d\nu$-grid:
{\small 
	\begin{align*}
		\psi_{0, \lambda_1 + d\nu, \lambda_2} &= \psi_{0, \lambda_1, \lambda_2} + d\nu
		\times r_1(\eta_0, \lambda_1, \lambda_2), \\
		\psi_{0, \lambda_1, \lambda_2 + d\nu} &= \psi_{0, \lambda_1, \lambda_2} + d\nu
		\times r_2(\eta_0, \lambda_1, \lambda_2), \\
		\psi_{0, \lambda_1 + d\nu, \lambda_2 + d\nu} &= \psi_{0, \lambda_1, \lambda_2} + d\nu
		\times r_1(\eta_0, \lambda_1, \lambda_2) + d\nu \times r_2(\eta_0, \lambda_1, \lambda_2) + d\nu^2 \times r_{12}(\eta_0, \lambda_1, \lambda_2) \ .
	\end{align*}%
}

The above set of differential equations can be used to \textit{recursively} estimate $\psi_{0, \lambda_1, \lambda_2}$.
We start with $\psi_n$, an estimate of the unconstrained minimizer $\psi_0$. For a numerically small step size $d\nu$, we can compute $\psi_{n,d\nu, 0} = \psi_n + r_1(\eta_n, 0, 0)d\nu$, $\psi_{n,0, d\nu} = \psi_n + r_2(\eta_n, 0, 0)d\nu$, and $\psi_{n,d\nu, d\nu} = \psi_n + r_1(\eta_n, 0, 0)d\nu + r_2(\eta_n, 0, 0)d\nu + r_{12}(\eta_n, 0, 0)d\nu^2$. This procedure proceeds where at the $k$-th step, we define $\lambda_{j, k} = \lambda_{j, k-1}+d\nu$, for $j = 1, 2$, and compute $\psi_{n,\lambda_{1, k}, \lambda_{2, k-1}} = \psi_{0,\lambda_{1, k-1}, \lambda_{2, k-1}} + r_1(\eta_n, \lambda_{1, k-1}, \lambda_{2, k-1})d\nu$, $\psi_{n,\lambda_{1, k-1}, \lambda_{2, k}} = \psi_{0,\lambda_{1, k-1}, \lambda_{2, k-1}} + r_2(\eta_n, \lambda_{1, k-1}, \lambda_{2, k-1})d\nu$, and $\psi_{n,\lambda_{1, k}, \lambda_{2,k}}$ as
\begin{align*}
	\psi_{n, \lambda_{1, k-1}, \lambda_{2, k-1}} + r_1(\eta_n, \lambda_{1, k-1}, \lambda_{2, k-1})d\nu + r_2(\eta_n, \lambda_{1, k-1}, \lambda_{2, k-1})d\nu + r_{12}(\eta_n, \lambda_{1, k-1}, \lambda_{2, k-1})d\nu^2.   
\end{align*}
Since we can compute an estimate of $\psi_{n,\lambda_1, \lambda_2}$ for any $\lambda_1, \lambda_2$, it is possible to utilize empirical minimization as above to generate an estimate of $\lambda_0 = (\lambda_{1, 0}, \lambda_{2, 0})$.  

\vspace{0.15cm}
For examples of multi-dimensional constraints, see Appendices~\ref{app:cate_2dim} and \ref{app:er_cases_controls}. 

\clearpage
\section{Proofs}
\label{app:proofs}

\subsection{Proof of Lemma~\ref{lem:lossmin}} 

For a given $P \in \mathcal{M}$, define $\psi_{\lambda'} = \argmin_{\psi \in {\bf \Psi}} R_P(\psi) + \lambda' \Theta_{P}(\psi)$ as the optimizer of  penalized risk for a given penalty $\lambda' \in \openr$. Define a submodel through $\psi_{\lambda'}$, $\{ \psi_{\lambda', \delta} : \delta \in \openr \} \subseteq {\bf \Psi}$. Our differentiability assumptions and the fact that $\psi_{\lambda'}$ minimizes penalized risk, allows us to write for an arbitrary path
\begin{align}
	\frac{d}{d \delta} 
	R_P(\psi_{\lambda', \delta})
	\Big|_{\delta = 0} + \ \lambda'  \frac{d}{d \delta}
	\Theta_{P}(\psi_{\lambda', \delta})
	\Big|_{\delta = 0} = 0 \ , \quad \forall \lambda' \in \openr \ .
	\label{eq:scorepsif}
\end{align}%
Let $\lambda = \argmin_{\lambda' \in \openr} R_P(\psi_{\lambda'}) + \lambda' \Theta_{P}(\psi_{\lambda'})$. Then, $\lambda = \Lambda(P)$ while $\psi_{\lambda} = \Psi^*(P)$. Since $\lambda$ minimizes this profiled risk, it is also true that
\begin{equation}
	\begin{aligned}
		0 &= \frac{d}{d \lambda} \left[R_P(\psi_\lambda)
		+ \lambda  \Theta_P(\psi_{\lambda}) \right] \\
		&= \frac{d}{d \lambda} R_P(\psi_\lambda)
		+ \lambda  \frac{d}{d \lambda} \Theta_{P}(\psi_{\lambda}) + 
		\Theta_{P}(\psi_{\lambda}) \ .
	\end{aligned}
	\label{eq:scorelambda}
\end{equation}%
Next, note that $\{ \psi_{\epsilon} : \epsilon = \lambda + \delta, \delta \in \openr \}$ also defines a path through $\psi_{\lambda}$ so that \eqref{eq:scorepsif} at $\lambda' = \lambda$ implies:
\begin{align}
	0 &= \frac{d}{d \delta} \left[ R_P(\psi_\epsilon) + \lambda
	\Theta_{P}(\psi_\epsilon) \right] \Big|_{\delta = 0} 
	=  \frac{d}{d \epsilon} \left[ R_P(\psi_\epsilon)
	+ \lambda \Theta_{P}(\psi_\epsilon) \right] \frac{d \epsilon}{d \delta} \Big|_{\delta = 0} \notag \\
	&= \frac{d}{d \epsilon } R_P(\psi_{\epsilon})
	+ \lambda \frac{d}{d \epsilon } \Theta_{P}(\psi_\epsilon) \Big|_{\delta = 0} 
	= \frac{d}{d\lambda} R_P(\psi_\lambda)
	+ \lambda \frac{d}{d \lambda} \Theta_P(\psi_\lambda)
	\ . 
	\label{eq:scorelambda2}
\end{align}
Combining (\ref{eq:scorelambda}) and (\ref{eq:scorelambda2}) implies that $\Theta_{P}(\psi_{\lambda}) = 0$. 

\subsection{Proof of Theorem~\ref{thm:gradients}}

Consider paths through $\psi \in \bm{\Psi}$ of the form $\{\psi_{\delta, h}(z) = \psi(z) + \delta h(z): \delta \in \openr\}$. These paths are indexed by direction $\frac{d}{d\delta} \psi_{\delta,h} |_{\delta = 0} = h(z)$, which we can allow to vary over $\tangent_{P}(\psi) = L^2(P_{Z})$, the space of bounded real-valued functions of $z$ defined on the support of $Z$ implied by $P$. 

Let $L(\psi)(o) = (y - \psi(z))^2$ be the squared error loss. For any $h \in \tangent_P(\psi)$, we have
\begin{align*}
	\frac{d}{d \delta} P_0L(\psi_{\delta, h})(O)
	\Big|_{\delta = 0} 
	&= -\E_{P_0}\left[ 2 \{Y - \psi(Z)\} h(Z) \right] \\
	&= \E_{P_0} \left[2\{\psi(Z) - \psi_0(Z)\}
	h(Z)\right] \ . 
\end{align*}
Thus, the canonical gradient for the MSE risk is $D_{R, P_0}(\psi)(o) = 2\{\psi(z) - \psi_0(z)\}$.

For any $h \in \tangent_P(\psi)$, we have 
\begin{align*}
	\left. \frac{d}{d \delta} \Theta_{P_0}(\psi_{\delta, h}) \right |_{\delta = 0} = \int \kappa_0(z) \dot{\zeta}_0(z) h(z) dP_{0,Z}(z) \ . 
\end{align*}
Thus, the canonical gradient for the constraint is $D_{\Theta,P_0}(\psi) = \kappa_0 \dot{\zeta}_0$, where $\dot{\zeta}_0 = {d\zeta_0}/{d\psi}$.

In the cross-entropy risk, we have $L(\psi)(o) = -y\log \psi (z) - (1-y)\log \{1-\psi(z)\}$ as the negative log loss. To derive the canonical gradient of the cross-entropy risk, we consider paths through $\psi(z) \in {\bf \Psi}$ at $\delta=0$ of the form 
\begin{align*}
	\big\{ \psi_{\delta, h} = \text{expit}\left[ \text{logit}\{\psi(z)\} + \delta h(z)  \right]: \delta \in \openr \big\} \ , 
\end{align*}
with direction $\frac{d}{d\delta} \psi_{\delta,h}|_{\delta = 0} =  h(z)\psi(z)(1-\psi(z))$. We note that although the path direction $\left.\frac{d}{d\delta}\psi_{\delta,h}\right|_{\delta=0}$ is indexed by $\psi$, one may suppress this dependence and carry out the derivations using a direction indexed solely by $h$. We can explore all possible local directions at $\psi$ within the parameter space $\bf \Psi$ by letting $h(z)$ vary over real-valued function of $z$. Thus, $\tangent_{P}(\psi)$ could be defined as a subspace of $L^2(P_{Z}),$ that is the space of real-valued functions defined on the support of $Z$ implied by $P$. For any $h \in \tangent_P(\psi)$, we have
\begin{align*}
	\frac{d}{d \delta} P_0L(\psi_{\delta, h})
	\Big|_{\delta = 0} 
	&= \int \left\{\frac{-y}{\psi(z)} + \frac{1-y}{1- \psi(z)} \right\} h(z) \ dP_0(o) \\
	&= \int \left\{\frac{-\psi_0(z)}{\psi(z)} + \frac{1-\psi_0(z)}{1- \psi(z)} \right\} h(z) \ dP_0(z) \\ 
	&= \int  \frac{\psi(z) - \psi_0(z)}{\psi(z) \ (1-\psi(z))} \ h(z) \ dP_0(z) \ . 
\end{align*}
Thus, the canonical gradient for the cross-entropy risk is $D_{R, P_0}(\psi)(o) = \displaystyle \frac{\psi(z) - \psi_0(z)}{\psi(z) \ (1-\psi(z))}$. Under the above defined paths through $\psi$, the canonical gradient for the constraint remains as $D_{\Theta,P_0}(\psi) = \kappa_0 \dot{\zeta}_0$, where $\dot{\zeta}_0 = {d\zeta_0}/{d\psi}$.

We note that if we considered the path direction  indexed by $\psi$, i.e., $\left.\frac{d}{d\delta}\psi_{\delta,h}\right|_{\delta=0} = h\psi(1-\psi)$, then both the risk and constraint gradients would be scaled by $\psi(1 - \psi)$. What matters for the constraint-specific path is that these gradients are scaled proportionally, since the path is determined by the first-order condition: (risk gradient) $+ \lambda \times$(constraint gradient) $= 0$.

\subsection{Proof of Corollary\ref{corollary:closed_forms}} 
\label{app:proofs_corollary}

\noindent \textit{\textbf{Solution under the MSE risk:}} 
By Theorem \ref{thm:gradients}, $D_{R, P_0}(\psi)(o) = 2\{\psi(z) - \psi_0(z)\}$, and $D_{\Theta, P_0}(\psi) = \kappa_0$ (when $\zeta(\psi)=\psi$). Condition~\eqref{eq:pathcondition-a} implies:  
\begin{align*}
	2\{\psi_{0, \lambda_0}(z) - \psi_0(z)\} + \lambda_0 \kappa_0(z) = 0 \ \Rightarrow \  \psi_{0, \lambda_0}(z) = \psi_0(z) - 0.5 \lambda_0 \kappa_0(z) \ , 
\end{align*}
where $\lambda_0$ solves the constraint equation: $\Theta_{P_0}(\psi_{0, \lambda_0}) = 0$. That is, 
\begin{align*}
	0 
	&= \E_{P_0}\{ \kappa_0(Z) \ \psi_{0, \lambda_0}(Z) \} \\
	&= \E_{P_0}\Big\{ \kappa_0(Z) \ \big\{ \psi_0(Z) - 0.5 \lambda_0 \kappa_0(Z) \big\} \Big\} \\
	&= \Theta_{P_0}(\psi_0) - 0.5 \lambda_0 \E_{P_0}\{ \kappa_0(Z)^2\} \ , 
\end{align*}
which yields: $ \lambda_0 = \frac{2 \ \Theta_{P_0}(\psi_0)}{\E_{P_0}\{ \kappa_0(Z)^2\} }$. In the main manuscript, we write $\E_{P_0}\{ \kappa_0(Z)^2\}$ as $P_0 \kappa_0^2$. 

\vspace{0.5cm}
\noindent \textit{\textbf{Solution under the cross-entropy risk:}} 
By Theorem \ref{thm:gradients}, $D_{R, P_0}(\psi)(o) = \frac{\psi(z) - \psi_0(z)}{\psi(z) \ (1-\psi(z))}$, and $D_{\Theta, P_0}(\psi) = \kappa_0$ (when $\zeta(\psi)=\psi$). Condition~\eqref{eq:pathcondition-a} implies:  
\begin{align*}
	\frac{\psi(z) - \psi_0(z)}{\psi(z) \ (1-\psi(z))} + \lambda_0 \kappa_0(z) = 0 \ \Rightarrow \ \lambda_0 \kappa_0(z) \psi^2_{0, \lambda}(z) - (1 + \lambda_0 \kappa_0(z)) \psi_{0, \lambda}(z) + \psi_0(z)  = 0 \ .  
\end{align*}

We assert that a solution for $\psi_{0, \lambda}$ is assured to exist within the unit interval $(0,1)$. By applying the quadratic formula, we find the roots as:
\begin{equation*}
	\frac{1 + \lambda \kappa_0(z) \pm \{ 1 + \lambda^2 C_0^2(z) + 2 \lambda \kappa_0(z) - 4\lambda \kappa_0(z) \psi_0(z) \}^{1/2}}{2 \lambda \kappa_0(z)} \ .
\end{equation*}
This simplifies to: 
\begin{align*}
	\frac{ \{ \lambda + \kappa_0(z)^{-1} \} \pm  \big\{ ( \lambda + \kappa_0(z)^{-1})^2 - 4\lambda \kappa_0(z)^{-1}\psi_0(z)   \big\}^{1/2}  }{2\lambda} \ . 
\end{align*}

The presence of two solutions is guaranteed by the non-negativity of the discriminant, as evidenced by the following arguments. 
\begin{itemize}
	\item For $\lambda > 0$ and $\kappa_0(z) > 0$: $(\lambda + \kappa_0(z)^{-1})^2 - 4\lambda \kappa_0(z)^{-1} \psi_0(1, w) \geq  (\lambda - \kappa_0(z)^{-1})^2$.   
	
	\item For $\lambda > 0$ and $\kappa_0(z) < 0$: $(\lambda + \kappa_0(z)^{-1})^2 - 4\lambda \kappa_0(z)^{-1} \psi_0(0, w) \geq 0$.  
	
	\item For $\lambda < 0$ and $\kappa_0(z) > 0$: $(\lambda + \kappa_0(z)^{-1})^2 - 4\lambda \kappa_0(z)^{-1} \psi_0(1, w) \geq 0$ 
	
	\item For $\lambda < 0$ and $\kappa_0(z) < 0$: $(\lambda + \kappa_0(z)^{-1})^2 - 4\lambda \kappa_0(z)^{-1} \psi_0(0, w) \geq (\lambda - \kappa_0(z)^{-1})^2$.  
\end{itemize}

Using these inequalities, it is straightforward to show that the following solution for $\psi_{0, \lambda}(z)$ always lies within the unit interval: 
\begin{align*}
	\frac{ \{ \lambda + \kappa_0(z)^{-1} \} + (-1)^{\I(\kappa_0(z) > 0)}  \big\{ ( \lambda + \kappa_0(z)^{-1} )^2 - 4\lambda \kappa_0(z)^{-1}\psi_0(z)   \big\}^{1/2}  }{2\lambda} \ . 
\end{align*}

\subsection{Proof of Theorem~\ref{thm:risk_and_constraint_at_P0}} 

\vspace{0.25cm}
\noindent \underline{\textit{\textbf{Risk condition: $R_{P_0}(\psi^*_n) - R_{P_0}(\psi^*_0) = o_{P_0}(1)$}}}

\vspace{0.5cm} \noindent 
Note that under Corollary~\ref{corollary:closed_forms}, we have $\psi^*_0(z) = \psi_0(z) - \Theta_{P_0}(\psi_0) \displaystyle \frac{\kappa_0(z)}{\sigma^2_0(\kappa_0)}$, where $\sigma^2_0(\kappa_0) = P_0 \kappa_0^2$. A plug-in estimator is given by $\psi^*_n(z) = \psi_n(z) - \Theta_{n} \displaystyle \frac{\kappa_n(z)}{\sigma^2_n}$. We expand on the MSE risk expression under $\psi^*_n$ as follows, where  

{\small
	\begin{align*}
		&R_{P_0}(\psi^*_n) = \int \left\{y - \psi^*_n(z)\right\}^2 dP_0(o) \\ 
		&= \int  \left\{y - \psi_0(z) + \psi_0(z) - \psi^*_n(z)\right\}^2 dP_0(o) \\
		&= \int \left\{\left\{y - \psi_0(z)\right\}^2  + \left\{\psi_0(z) - \psi^*_n(z)\right\}^2 \right\} dP_0(o) \\
		&= \int \left\{\left\{y - \psi_0(z)\right\}^2  +  \left\{ \psi_0(z) - \psi_n(z) +  \frac{\Theta_n}{\sigma^2_n} \kappa_n(z)\right\}^2 \right\} dP_0(o) \\
		&= \int \left\{ \left\{y - \psi_0(z)\right\}^2 + \left\{\psi_0(z) - \psi_n(z)\right\}^2 + \left\{\frac{\Theta_n}{\sigma^2_n} \kappa_n(z) \right\}^2 + 2\frac{\Theta_n}{\sigma^2_n} \left\{\psi_0(z) - \psi_n(z)\right\} \kappa_n(z) \right\} dP_0(o) \\
		&= \int \left[ \left\{y - \psi_0(z)\right\}^2 + \left(\frac{\Theta_0}{\sigma^2_0}\right)^2 \kappa^2_0(z) \right] dP_0(o)  \\
		&\hspace{4em}
		+ P_0(\psi_0 - \psi_n)^2 + 
		\left(\frac{\Theta_n}{\sigma^2_n} \right)^2 P_0 \kappa_n^2
		- \left(\frac{\Theta_0}{\sigma^2_0}\right)^2 P_0 \kappa^2_0
		+ 2 \frac{\Theta_n}{\sigma^2_n} P_0 \left[ \left\{\psi_0 - \psi_n\right\} \kappa_n \right] \\
		&=R_{P_0}(\psi^*_0) + P_0(\psi_0 - \psi_n)^2 + 
		\left(\frac{\Theta_n}{\sigma^2_n} \right)^2 P_0 \kappa_n^2
		- \left(\frac{\Theta_0}{\sigma^2_0}\right)^2 P_0 \kappa^2_0
		+ 2 \frac{\Theta_n}{\sigma^2_n} P_0 \left[ \left\{\psi_0 - \psi_n\right\} \kappa_n \right] \ . 
	\end{align*}
}%
Therefore, 
{\small 
	\begin{equation} \label{eq:risk_diff_asymptotic}
		\begin{aligned}
			&R_{P_0}(\psi_n^*) - R_{P_0}(\psi^*_0) \\
			&\hspace{0.25cm} = 
			P_0 (\psi_n - \psi_0)^2 + \left\{ \left(\frac{\Theta_n}{\sigma^2_n}\right)^2 - \left(\frac{\Theta_0}{\sigma^2_0} \right)^2 \right\} P_0 \kappa_n^2 + \left(\frac{\Theta_0}{\sigma^2_0} \right)^2 \!\! P_0 (\kappa_n^2 - \kappa_0^2) + 2 \frac{\Theta_n}{\sigma^2_n} P_0 \left\{ (\psi_0 - \psi_n) \kappa_n \right\} \ .
		\end{aligned}
	\end{equation}
}%

\vspace{0.5cm}
\noindent $\bullet$ The first term on the right-hand side of (\ref{eq:risk_diff_asymptotic}) equals $o_{P_0}(1)$ by the assumption of $L^2(P_0)$-consistency of $\psi_n$.

\noindent $\bullet$ The second term on the right-hand side of (\ref{eq:risk_diff_asymptotic}) can be written as 
{\small 
	\begin{align*}
		&\left\{ \left(\frac{\Theta_n}{\sigma^2_n}\right)^2 - \left(\frac{\Theta_0}{\sigma^2_0} \right)^2 \right\} P_0 \kappa_n^2 \\
		&\hspace{1cm} = \left\{ \left(\frac{\Theta_n}{\sigma^2_n} - \frac{\Theta_0}{\sigma^2_0} \right)^2 + 2 \frac{\Theta_0}{\sigma^2_0} \left( \frac{\Theta_n}{\sigma_n^2} - \frac{\Theta_0}{\sigma^2_0} \right) \right\} P_0 \left[ (\kappa_n - \kappa_0)^2 + 2 \kappa_0(\kappa_n - \kappa_0) + \kappa^2_0 \right] \ ,
	\end{align*}
}%
indicating that this term consists only of second- and higher-order terms that will all equal to $o_{P_0}(1)$ under our assumptions.

\noindent $\bullet$ The third term on the right-hand side of (\ref{eq:risk_diff_asymptotic}) is \begin{align*}
	\left(\frac{\Theta_0}{\sigma^2_0}\right)^2 P_0 (\kappa_n^2 - \kappa_0^2) = \left(\frac{\Theta_0}{\sigma^2_0}\right)^2 P_0 \left\{(\kappa_n - \kappa_0)^2 + 2 \kappa_0(\kappa_n - \kappa_0) \right\} \ .
\end{align*}
$L^2(P_0)$-consistency of $\kappa_n$, along with the fact that $\Theta_0$ is finite and $\sigma_0^2$ is non-zero, implies that \[
\left(\frac{\Theta_0}{\sigma^2_0}\right)^2 P_0 (\kappa_n - \kappa_0)^2 = o_{P_0}(1) \ .
\]
These same assumptions combined with H{\"o}lder's inequality implies that
\[
\left(\frac{\Theta_0}{\sigma^2_0}\right)^2 P_0 \kappa_0 (\kappa_n - \kappa_0) = o_{P_0}(1) \ .
\]

\noindent $\bullet$ The fourth term on the right-hand side of (\ref{eq:risk_diff_asymptotic})
is \begin{align*}
	2 \frac{\Theta_n}{\sigma^2_n} P_0 \left\{ (\psi_0 - \psi_n) \kappa_n \right\} = 2 \frac{\Theta_n}{\sigma^2_n} P_0 \left\{ (\psi_0 - \psi_n) (\kappa_n - \kappa_0) + \kappa_0(\psi_n - \psi_0) \right\} \ . 
\end{align*}
The term $P_0 \left\{ (\psi_0 - \psi_n) (\kappa_n - \kappa_0)\right\}$ is second-order and will equal to $o_{P_0}(1)$ under the $L^2(P_0)$ consistency assumptions by the Cauchy-Schwarz inequality. On the other hand, boundedness of $\kappa_0$, $L^2(P_0)$-consistency, and H{\"o}lder's inequality imply that $P_0 \{ \kappa_0 (\psi_n - \psi_0) \} = o_{P_0}(1)$.

\vspace{0.75cm}
\noindent \underline{\textit{\textbf{Constraint condition: $\Theta_{P_0}(\psi^*_n) = o_{P_0}(1)$}}} 

\vspace{0.5cm} \noindent 
The constraint $\Theta_{P_0}(\psi^*_n)$ can be expanded on as follows:
\begin{equation*}
	\begin{aligned}
		\Theta_{P_0}(\psi^*_n) 
		&= \int \kappa_0(z) \psi^*_n(z) \  dP_0(o) 
		\\
		&= \int \kappa_0(z) \left\{ \psi_n(z) - \Theta_n \frac{\kappa_n(z)}{\sigma_n^2} \right\} dP_0(o)
		\\
		&= \int \kappa_0(z) \psi_n(z) dP_0(o) - \frac{ \Theta_n}{\sigma_n^2} \int \kappa_0(z)  \kappa_n(z) dP_0(o)
		\\
		&= \Theta_{P_0}(\psi_n) - \frac{\Theta_{n}}{\sigma^2_n} P_0 ( \kappa_0 \kappa_n ) \\
		&= \Theta_{P_0}(\psi_n) - \frac{\Theta_{n}}{\sigma^2_n} P_0 \{ \kappa_0 ( \kappa_n - \kappa_0 ) \} - \frac{\sigma^2_0}{\sigma^2_n} \Theta_{n} \ . 
	\end{aligned}
\end{equation*}

We note that $\Theta_{P_0}(f) = P_0 ( \kappa_0 f )$ for any $P_0$-measurable function $f$. Thus, we can write:
\begin{align*}
	&\Theta_{P_0}(\psi_n) - \frac{\Theta_{n}}{\sigma^2_n} P_0 \{ \kappa_0 ( \kappa_n - \kappa_0 ) \} - \Theta_{n} \frac{\sigma^2_0}{\sigma^2_n} \\
	&\hspace{2em} = P_0 \{ \kappa_0 (\psi_n - \psi_0) \} - \frac{\Theta_{n}}{\sigma^2_n} P_0 \{ \kappa_0 ( \kappa_n - \kappa_0 ) \} - \sigma^2_0 \left( \frac{\Theta_{n}}{\sigma^2_n} - \frac{\Theta_0}{\sigma^2_0} \right) \ . 
\end{align*}

In this case, sufficient conditions are: (i) bounded $\kappa_0$; (ii) $L^1(P_0)$ convergence of $\kappa_n$ and $\psi_n$; (iii) consistency of $\Theta_n, \sigma^2_n$ to $\Theta_0, \sigma^2_0$.

One could also analyze as follows. Let $\tilde{\psi}$ be the in-probability limit of $\psi_n$. Then 
\begin{align*}
	&\Theta_{P_0}(\psi_n) - \frac{\Theta_{n}}{\sigma^2_n} P_0 \{ \kappa_0 ( \kappa_n - \kappa_0 ) \} - \frac{\sigma^2_0}{\sigma^2_n} \Theta_{n} \\
	&\hspace{2em} = P_0 \{ \kappa_0 (\psi_n - \tilde{\psi}) \} - \frac{\Theta_{n}}{\sigma^2_n} P_0 \{ \kappa_0 ( \kappa_n - \kappa_0 )\} - \sigma^2_0 \left( \frac{\Theta_{n}}{\sigma^2_n} - \frac{\Theta_{P_0}(\tilde{\psi})}{\sigma^2_0}\right) \ . 
\end{align*}
In this case, the sufficient conditions for convergence are those given by the Theorem: (i) bounded $\kappa_0$, (ii) $L^1$-consistency of $\psi_n$ and $\kappa_n$ for $\tilde{\psi}$ and ${\kappa}_0$, respectively; and (iii) $\Theta_n, \sigma^2_n$ being consistent for $\Theta_{P_0}(\tilde{\psi}), \sigma^2_0$.

\clearpage
\section{Additional conceptual discussions}
\label{app:related_work}

\subsection{On various notions of fairness} 

As a motivating example, consider a setting in which a financial institution is tasked with providing loans to applicants, and it desires to use an automated system that utilizes an applicant's characteristics to make a binary loan approval decision or establish a specific loan amount. To construct this prediction system, the financial institution can leverage historical data on previous loans. For a given applicant, we can represent the data as $O = (W, X, Y)$, where $W$ denotes the set of applicant characteristics (such as income and credit rating), $X$ represents a potentially sensitive characteristic (such as race/ethnicity), and $Y$ denotes the outcome of interest (such as an indicator of whether the loan was repaid successfully). Traditional statistical learning methods aim to obtain the ``best'' prediction of $Y$ given $X$ and $W$, for example using empirical risk minimization techniques~\citep{vdl2003unifiedcv, vdl2004asymptotic, dudoit2005asymptotics}. However, such an approach may result in prediction systems that treat certain groups \emph{unfairly} across various values of the sensitive characteristic. Instead, financial institutions may seek to obtain the optimal prediction of $Y$, given $(X, W$, while preserving a predefined notion of fairness with respect to the sensitive characteristic $X$. 

For instance, for binary $X$ and $Y$, one might seek to incorporate into the loan approval system an optimal prediction function that also satisfies \textit{equal opportunity}~\citep{hardt2016equality, dwork2012fairness, fairmlbook}, a constraint ensuring that the true positive rate (of the prediction function) remains approximately fixed across the two values of $X$. Such a constraint requires that, among those who pay back their loans, an equal opportunity (with respect to $X$) exist of being granted the loan in the first place, without imposing any similar constraint upon those who eventually default on their loans. Mathematically, this corresponds to defining
{\small 
	\begin{align*}
		\Theta_{P_0}(\psi, c) = P_0(\I(\psi(X, W) > c) = 1 \mid Y = 1, X = 1) - P_0(\I(\psi(X, W) > c) = 1 \mid Y = 1, X = 0) \ , 
	\end{align*}
}
where $c$ is the classifier threshold. Similarly, other associative notions of fairness (such as demographic parity, risk parity, or equalized odds) arise by substituting the appropriate conditional probabilities into $\Theta_{P_0}(\psi,c)$.

Alternatively, the financial institution may wish to develop a system that, on average, approves loans in identical amounts for individuals under counterfactual scenarios in which the value of their sensitive characteristic is changed---a criterion termed \emph{counterfactual fairness}~\citep{kusner2017counterfactual}. A population-level variation of the counterfactual fairness aligns with restricting the overall effect of $X$ on $Y$. Mathematically, this corresponds to defining  
{\small 
	\begin{align*}
		\Theta_{P_0}(\psi) = \int \left\{ \psi(1, w) - \psi(0, w)  \right\} dP_{0, W}(w) \ . 
	\end{align*}
}%
Causal notions of fairness, such as the counterfactual fairness, have also been extended to mediation and path-specific counterfactual fairness, which restricts the effect of the sensitive characteristic on the outcome along certain impermissible mediating pathways~\citep{nabi2018fair, nabi2019learning, chiappa2019path}. In general, counterfactual and causal reasoning has emerged as an important framework for quantifying the notion of algorithmic fairness \citep{zhang2017causal, kusner2017counterfactual, zhang2018fairness, chiappa2019path, makhlouf2020survey, nabi2022optimal, nilforoshan2022causal}.

Our framework accommodates essentially any pathwise differentiable fairness constraint. When pathwise differentiability does not hold, we could potentially approximate the constraint using smooth surrogates. For example, the equal opportunity as defined above, is not pathwise differentiable due to the hard‐threshold indicator function, but can be approximated with an smooth approximation. For illustration purposes, we instead looked at the equalized risk in the main manuscript, by replacing $\I(\psi(X, W) > c) = 1$ with the loss function, $L(\psi)(X, W)$.

\subsection{On fairness adjustments under our methodology}  

Fairness in machine learning motivates a primary class of constrained learning problems. In this context, a notable strength of our framework is that we can often obtain closed-form solutions that express the optimal approach for fair prediction in terms of interpretable parameters of $P_0$. This is important for several reasons. First, it enables practitioners to integrate background scientific knowledge into the estimation process of these interpretable parameters, potentially leading to improved predictive performance. This approach stands in contrast to methods that aim to estimate the constrained parameter directly, which may not effectively leverage such background knowledge, possibly resulting in sub-optimal predictive performance. Furthermore, having closed-form solutions for optimal fair prediction facilitates an insight into the dynamics that result in unfair predictions, and outlines strategies for their mitigation. This insight could be particularly valuable for the development and maintenance of  online monitoring systems for machine learning models in practice, ensuring they remain fair under evolving data distributions. Taking the average treatment effect as an example, our results suggest that, while a population shift in $\psi_0$ and/or $\pi_0$ suggest a sub-optimal predictive model, a population shift in $P_{0,W}$ would imply that a predictive algorithm that was previously deemed ``fair'' may no longer yield fair predictions under the shifted distribution. 

\subsection{On equality vs. inequality constraints}  

In practice, there is an inherent trade-off between predictive performance and constraint satisfaction; particularly when fairness or policy goals are imposed as hard constraints. In many critical applications, such as parole hearings or organ allocation, guaranteed constraint satisfaction is often essential, and equality constraints (e.g., ensuring group-level parity in treatment or error rates) are essential in operational practice and regulatory frameworks. 

That said, our framework does not require the constraint to be set to a null value, and indeed supports a broad class of constraints that can be centered at application-specific targets. For instance, one may impose a constraint of the form $\Theta_P(\psi) \leq \tau$, with $\tau$ chosen to reflect domain knowledge or legal/policy thresholds. This flexibility allows users to explore the predictive performance–constraint trade-off curve by varying $\tau$, or to apply inequality constraints when strict fairness is impractical or overly costly. 

Selecting such thresholds remains a domain-specific challenge. This decision often lies beyond the statistician's expertise and should involve input from subject-matter experts, ethicists, or policymakers. In this sense, the fact that our framework supports general equality and inequality constraints (without requiring any fixed target) provides a flexible platform for aligning with diverse fairness or operational objectives. 

The discussion of fairness-aware learning under constraint uncertainty by \cite{viviano2024fair} supports the general idea that soft or flexible constraint formulations may be more appropriate in many practical settings. We see this as complementary to our framework, which can be readily adapted to enforce such inequality-based constraints (including multi-dimensional forms), and we have added a citation and brief discussion to this effect in the revised manuscript.

\subsection{On different approaches to fair constrained estimation}

Irrespective of the fairness criteria adopted, the objective of algorithmic fairness in practice involves using data to create a prediction system that adheres to a user-selected fairness constraint. The literature on constrained learning for fairness can broadly be categorized into three primary categories based on the stage of intervention to ensure fairness: 
(i) \textit{pre-processing} methods, which adjust the input data to reduce dependencies between sensitive attributes and class labels while preserving the information content between the class labels and other variables \citep{kamiran2012data, zemel2013learning, calmon2017optimized}; 
(ii) \textit{in-processing} methods, which involve incorporating fairness constraints or regularization terms directly into the optimization problem \citep{zafar2019fairness}, for example using constrained/regularized empirical risk minimization \citep{donini2018empirical}; 
and (iii) \textit{post-processing} methods, which modify the outputs of the model to achieve fairness without changing the training process \citep{hardt2016equality, woodworth2017learning}.  

As an example of a \emph{post‐processing} approach, \cite{wang2023adjusting} propose methods for adjusting any pre‐trained decision model to satisfy causal fairness definitions—specifically equal counterfactual opportunity (ECO) and counterfactual fairness (CF). Their methods: (i) use a fitted ML model without requiring retraining; (ii) construct ECO-fair decisions by marginalizing over the protected attribute while preserving dependence on other features; (iii) construct CF-fair decisions by adjusting ECO decisions using counterfactual reasoning via abduction within a causal model; and (iv) provide theoretical guarantees that the adjustments remain as close as possible, in terms of minimizing the Kullback-Leibler divergence, to the original ML predictions while satisfying fairness criteria. 

The proposed methods, however, only target ECO and CF fairness and do not generalize to other fairness notions. They also do not address enforcing multiple fairness constraints simultaneously, which may be needed in practice. Additionally, the functional model space is constrained indirectly through the assumed structural causal model and the fairness definitions (ECO/CF), which define subsets of decision functions satisfying these fairness criteria under the SCM. While the methods do not explicitly constrain the functional form of ML predictors, in practice they use additive-error models and linear/logistic models to enable tractable counterfactual estimation and fairness adjustments. 

Our framework, by contrast, supports arbitrary pathwise differentiable constraints, including those motivated by fairness, causal transportability, risk control, and structural restrictions—providing a unified and extensible formulation.   

More recent work by \cite{jordan2022data} advances automated bias adjustment in causal inference and offline reinforcement learning by using finite-difference-based, data-driven influence functions, enabling flexible bias correction without custom analytical derivations. While they do not propose a new optimization solver in the traditional sense, they introduce a structured, dual-based perturbation analysis framework for bias-adjusted, statistically efficient estimation within optimization-based causal inference problems. This leverages existing solvers while enabling systematic bias correction via finite-difference methods informed by the problem’s dual structure. The paper also characterizes when these numerical approximations retain desirable statistical properties, such as double robustness, under appropriate perturbation and smoothing rates. 

Their procedure estimates optimization-based causal estimands (e.g., constrained policy values, sensitivity bounds) through four key steps:
(i) compute the plug-in estimate using existing solvers;
(ii) perturb the empirical data and re-solve to approximate the influence function via finite differences, leveraging dual variables and structured perturbation analysis;
(iii) apply a one-step bias correction using these influence estimates;
(iv) generalize this procedure across different constrained optimization-based estimands with minimal additional analytic effort. 

This approach automates influence-function-based bias adjustment without requiring closed-form derivations for each new constraint or estimand. While the provides a practical and automated approach for bias adjustment in optimization-based causal inference, the method has several limitations. It requires careful tuning of perturbation sizes and smoothing bandwidths, which can be challenging in practice, and suffers from the curse of dimensionality when applied to high-dimensional continuous covariates. The approach can also be computationally intensive, as it requires repeated calls to optimization solvers for each data perturbation, which may be prohibitive for large-scale or complex problems. Additionally, it assumes stability of the optimization solution under perturbations, which may not hold in non-convex or degenerate settings, potentially leading to instability in constrained or complex optimization formulations. Finally, they do not explicitly analyze or provide specialized methods for handling multiple constraints simultaneously, beyond the general statement that their framework applies to constrained optimization problems.  

In contrast, our approach focuses on first-order characterizations of constrained functional parameters via gradient conditions in Hilbert spaces. This leads to closed-form or recursive plug-in estimators that do not require solving perturbed optimization problems, and can be readily paired with modern learners and cross-fitting. 

Alternative approaches for fair machine learning include \textit{meta algorithms} \citep{agarwal2019fair} and \textit{Bayesian inference} \citep{gardner2014bayesian, dimitrakakis2019bayesian, chiappa2019causal, foulds2020bayesian, perrone2021fair}.

\clearpage
\section{Constraint: average total effect (ATE)}
\label{app:ate}


\subsection{ATE: canonical gradients}
\label{app:ate_gradients}

We can write the ATE under $\psi$ as: 
\begin{align*}
	\Theta_{P_0}(\psi) &= \E_{P_0}\left[ \displaystyle \frac{2X-1}{\pi_0(X \mid W)} \psi(X, W) \right] \ . 
\end{align*}
According to the general constraint form in \eqref{eq:general_constraint}: 
\begin{align*}
	\kappa_0(x, w) = \frac{2x-1}{\pi_0(x \mid w)}  \quad  \mbox{and} \quad \zeta_0(\psi) = \psi \ .
\end{align*}

\noindent Therefore, the canonical gradients, provided by Theorem~\ref{thm:gradients}, are as follows:
\begin{itemize}
	\item MSE risk gradient: $D_{R, P_0}(\psi)(o) = 2\{\psi(x,w) - \psi_0(x,w)\}$, 
	\item Cross-entropy risk gradient: $D_{R, P_0}(\psi)(o) = \displaystyle \frac{\psi(x, w)-\psi_0(x, w)}{\psi(x, w) ( 1- \psi(x, w))}$, 
	\item Constraint gradient: $D_{\Theta, P_0}(\psi)(o) = \kappa_0(x,w)$. 
\end{itemize}

\subsection{ATE: second-order conditions}
\label{app:ate_2nd}

For verifying second-order conditions, we follow the discussion in Appendix~\ref{app:second-order}. 

\subsubsection*{2nd-order conditions under MSE risk: }

Given the canonical gradients $D_{R, P_0}(\psi)$ and $D_{\Theta, P_0}(\psi)$, we can write the following:
\begin{align*}
	\mathcal{L}_{P_0}(\psi, \lambda) &=  \int \{\psi_0(x, w) - \psi(x, w)\}^2 \ dP_0(x, w)  + \lambda \int \{\psi(1, w) - \psi(0, w)\} \ dP_0(w) \ , \\ 
	\dot{\mathcal{L}}_{P_0}(\psi, \lambda) &=  2\{\psi(x,w) - \psi_0(x,w)\} + \lambda \kappa_0(x,w) \ , \\
	\ddot{\mathcal{L}}_{P_0}(\psi, \lambda) &= 2 \ .
\end{align*} %
For any given $\psi = (\psi^1, \psi^0) \in {\bf \Psi}$ and $\lambda \in \openr$, the criterion \eqref{eq:borderedH_det_single} simplifies to: 
\begin{align*}
	- D^2_{\Theta, P}(\psi^1) \times \ddot{\calL}_P(\psi^0, \lambda)
	- D^2_{\Theta, P}(\psi^0) \times \ddot{\calL}_P(\psi^1, \lambda) 
	= - 2 (\kappa_0(1, w)^2 + \kappa_0(0, w)^2) < 0 \ . 
\end{align*}
This concludes that $\psi_{0, \lambda_0} = (\psi^1_{0, \lambda_0}, \psi^0_{0, \lambda_0})$ is indeed the optimal  minimizer. 

\subsubsection*{2nd-order conditions under cross-entropy risk:}

The existence of a valid solution to the quadratic equation 
\begin{align*}
	\lambda \kappa_0(x,w) \psi^2_{0, \lambda}(x, w) - \{1 + \lambda \kappa_0(x,w)\} \psi_{0, \lambda}(x, w) + \psi_0(x, w) = 0 \ , 
\end{align*}%
is discussed in Appendix~\ref{app:proofs_corollary}. The unique solution is given by: 
\begin{align*}
	\psi_{0, \lambda}(x, w) = \frac{ \{ \lambda + \kappa_0(x, w)^{-1} \} - (2x-1) \big\{ ( \lambda + \kappa_0(x, w)^{-1} )^2 - 4\lambda \kappa_{0}(x, w)^{-1}\psi_0(x, w)   \big\}^{1/2}  }{2\lambda} \ . 
\end{align*}

Verifying that this solution is indeed a minimum requires confirming the criterion \eqref{eq:borderedH_det_single}. Given the canonical gradients $D_{R, P_0}(\psi)$ and $D_{\Theta, P_0}(\psi)$, we can write: 
\begin{align*}
	\mathcal{L}_{P_0}(\psi, \lambda) &= \int (\psi_0(x,w) \log \psi(x,w) - (1-\psi_0(x,w)) \log (1-\psi(x,w))  dP_{0}(x,w) \\
	&\hspace{3cm} + \lambda \int \{\psi(1, w) - \psi(0, w)\} dP_0(w) \ ,   \\ 
	\dot{\mathcal{L}}_{P_0}(\psi, \lambda) &=  \frac{1-\psi_0(x, w)}{1- \psi(x, w)} - \frac{\psi_0(x, w)}{\psi(x, w)} + \lambda \kappa_0(x,w) \ ,  \\
	\ddot{\mathcal{L}}_{P_0}(\psi, \lambda) &= \frac{1 - \psi_0(x, w)}{\{1 - \psi(x, w)\}^2} + \frac{\psi_0(x, w)}{\psi^2(x, w)} \ . 
\end{align*}
For any given $\psi = (\psi^1, \psi^0) \in {\bf \Psi}$ and $\lambda \in \openr $, the criterion \eqref{eq:borderedH_det_single} always hold, which concludes that $\psi_{0, \lambda_0} = (\psi^1_{0, \lambda_0}, \psi^0_{0, \lambda_0})$ is the optimal minimizer of the penalized cross-entropy risk.  

\subsection{ATE: estimation}
\label{app:ate_esti}

Let $\big|\big| f \big|\big| \coloneqq \left\{ \int f^2(o) \ dP(o) \right\}^{1/2}$ denote the $L^2(P_0)$-norm of the $P_0$-measurable function $f$, and let $\mathcal{X}$ and $\mathcal{W}$ denote the domains of $X$ and $W$. We assume the following $L^2(P_0)$-consistency for our nuisance estimates:
\begin{align}
	&|| \pi_n - \pi || = \smallO(n^{-{1}/{a}}) \ , 
	\quad 
	|| \psi_n - \psi_0 || = \smallO(n^{-{1}/{b}}) \ . 
	\label{eq:rate_ate}
\end{align}%

\subsubsection*{Estimation under MSE risk:}

We assume the following regularity conditions: 
\begin{equation}\label{eq:regularity_ate_mse}
	\begin{aligned}
		&\inf_{x \in \mathcal{X}, w \in \mathcal{W}} \pi_0(x \mid w) > 0 \ , \qquad 
		&&\inf_{x \in \mathcal{X}, w \in \mathcal{W}} \pi_n(x \mid w) > 0  \ , \\
		&\sup_{x \in \mathcal{X}, w \in \mathcal{W}} \psi_0(x, w) < +\infty  \ , \qquad 
		&&\sup_{x \in \mathcal{X}, w \in \mathcal{W}} \psi_n(x, w) < +\infty \ .
	\end{aligned}%
\end{equation}%
Using the triangle and Cauchy-Schwarz inequalities, we can write: 
\begin{align*}
	\big|\big| \psi_{n,\lambda_n} - \psi_{0, \lambda_0} \big|\big| 
	&= \big|\big| \psi_n - 0.5 \lambda_n \kappa_n  - \psi_0  + 0.5 \lambda_0 \kappa_0 \big|\big| \\
	&\leq \big|\big| \psi_n - \psi_0 \big|\big| + 0.5 \big|\big| \lambda_n \kappa_n   - \lambda_0 \kappa_0 \big|\big| \\
	&= \big|\big| \psi_n - \psi_0 \big|\big| + 0.5 \big|\big| \lambda_n \kappa_n   - \lambda_0 \kappa_0 \pm \lambda_0 \kappa_n \big|\big| \\
	&\leq \big|\big| \psi_n - \psi_0 \big|\big|  
	+ 0.5 \big| \lambda_0 \big| \times \big|\big| \kappa_n - \kappa_0 \big|\big| 
	+ 0.5 \big| \lambda_n - \lambda_0 \big| \times  \big| \big| \kappa_n \big| \big| \ . 
\end{align*}

Given the regularity conditions in \eqref{eq:regularity_ate_mse}, we have $0.5 \big| \lambda_0 \big| \times \big|\big| \kappa_n - \kappa_0 \big|\big| \leq \mathfrak{C}_1 \big|\big| \pi_n - \pi_0 \big|\big|$, where $\mathfrak{C}_1$ is a finite positive constant. Using the triangle and Cauchy-Schwarz inequalities again, and given the law of large numbers, as $n \rightarrow \infty$, we can write: 
{\small 
	\begin{align*}
		0.5 \big| \lambda_n - \lambda_0 \big|
		&\leq  \bigg| \int \Big[ \frac{1}{\pi_0( 1 \mid w) } + \frac{1}{ \pi_0( 0 \mid w)} \Big] dP_0(w) \bigg|^{-1} \\ 
		&\hspace{1cm} \times \ \Big| \int \big( \psi_0(1,w) - \psi_n(1,w) \big) -  \big( \psi_0(0,w) - \psi_n(0,w) \big) dP_0(w) \\
		&\hspace{0.25cm} + \Big| \int \left[\psi_n(1,w) - \psi_n(0,w) \right] dP_0(w) \Big| \\
		&\hspace{1cm}\times \ \bigg| \frac{1}{\int \Big[ \frac{1}{\pi_0( 1 \mid w) } + \frac{1}{ \pi_0( 0 \mid w)} \Big] dP_0(w)}  - \frac{1}{\int \left[ \frac{1}{\pi_n(1 \mid w)} + \frac{1}{\pi_n(0 \mid w)}\right]}  \bigg| \ .  
	\end{align*}
}%
Given the regularity conditions in \eqref{eq:regularity_ate_mse}, we can conclude $0.5 \big| \lambda_n - \lambda_0 \big| \times \big|\big| \kappa_n \big|\big| \leq \mathfrak{C}_2 \big|\big| \psi_n - \psi_0 \big|\big|$, where $\mathfrak{C}_2$ is another finite positive constant. Putting all these together, we have: 
\begin{align*}
	\big|\big| \psi_{n,\lambda_n} - \psi_{0, \lambda_0} \big|\big| 
	\leq \mathfrak{C}_1 \big|\big| \pi_n - \pi_0 \big|\big| 
	+ (1+\mathfrak{C}_2)  \big| \big| \psi_n - \psi_0 \big| \big| \ . 
\end{align*}%
Given the $L^2(P_0)$-consistency rates in \eqref{eq:rate_ate}, we get that  $ \big|\big| \psi_{n,\lambda_n} - \psi_{0, \lambda_0} \big|\big| = o_P(n^{\max\{-{1}/{a}, - {1}/{b}\}})$.

\subsubsection*{Estimation under cross-entropy risk:}

In addition to the regularity conditions in \eqref{eq:regularity_ate_mse}, we assume $\lambda_n$ and $\lambda_0$ are bounded. Using the triangle and Cauchy-Schwarz inequalities, we can write: 
{\small
	\begin{align*}
		\big|\big| \psi_{n,\lambda_n} &- \psi_{0, \lambda_0} \big|\big| \\
		&= \Big|\Big| \frac{ \{ \lambda_n + \kappa_n(x, w)^{-1} \} - (2x-1) \big\{ ( \lambda_n + \kappa_n(x, w)^{-1} )^2 - 4\lambda_n \kappa_n(x, w)^{-1}\psi_n(x, w)   \big\}^{1/2}  }{2\lambda_n} \ \\
		&\hspace{0.5cm} - \frac{ \{ \lambda_0 + \kappa_{0}(x, w)^{-1} \} - (2x-1) \big\{ ( \lambda_0 + \kappa_{0}(x, w)^{-1} )^2 - 4\lambda_0 \kappa_{0}(x, w)^{-1}\psi_0(x, w)   \big\}^{1/2}  }{2\lambda_0}  \Big|\Big| \\
		&= \frac{1}{2} \times \Big|\Big| \frac{1}{\pi_n \lambda_n} - \frac{1}{\pi_0 \lambda_0} -  \big\{  (1+ \frac{2x-1}{\pi_n \lambda_n})^2 - 4 \frac{(2x-1)}{\pi_n\lambda_n} \psi_n \}^{1/2} \\
		&\hspace{1.5cm} + \big\{  (1+ \frac{2x-1}{\pi_0 \lambda_0})^2 - 4 \frac{(2x-1)}{\pi_0\lambda_0} \psi_0 \}^{1/2} \Big| \Big| \\
		&\leq \frac{1}{2} \times \Big|\Big| \frac{1}{\pi_n \lambda_n} - \frac{1}{\pi_0 \lambda_0} \Big|\Big| \\
		&\hspace{0.5cm} + \frac{1}{2} \times \Big|\Big| \big\{  (1+ \frac{2x-1}{\pi_n \lambda_n})^2 - 4 \frac{(2x-1)}{\pi_n\lambda_n} \psi_n \}^{1/2} - \big\{  (1+ \frac{2x-1}{\pi_0 \lambda_0})^2 - 4 \frac{(2x-1)}{\pi_0\lambda_0} \psi_0 \}^{1/2}  \Big|\Big| \\
		&\leq \mathfrak{C}_1 \big|\big|  \pi_n - \pi_0 \big|\big| + \mathfrak{C}_2 \big|\big|  \psi_n - \psi_0 \big|\big| \ , 
	\end{align*}
}%
where $\mathfrak{C}_1$ and $\mathfrak{C}_2$ are finite positive constants. Given the $L^2(P_0)$-consistency rates in \eqref{eq:rate_ate}, we get that  $ \big|\big| \psi_{n,\lambda_n} - \psi_{0, \lambda_0} \big|\big| = o_P(n^{\max\{-{1}/{a}, - {1}/{b}\}})$.

\clearpage
\section{Constraint: natural direct effect (NDE)}
\label{app:nde}

\subsection{NDE: canonical gradients}
\label{app:nde_gradients}

We can write the NDE under $\psi$ as: 
\begin{align*}
	\Theta_{P_0}(\psi) &= \E_{P_0}\left[ \displaystyle \frac{2X-1}{\pi_0(X \mid W)} \gamma_0(M \mid X, W) \psi(X, M, W) \ \right] \ ,  
\end{align*}
where $\gamma_0(m \mid x, w) = f_{0, M}(m \mid 0, w)/f_{0, M}(m \mid x, w)$. According to the general constraint form in \eqref{eq:general_constraint}: 
\begin{align*}
	\kappa_0(x, m, w) = \frac{2x-1}{\pi_0(x \mid w)} \gamma_0(m \mid x, w) \ , \quad \mbox{and} \quad \zeta_0(\psi) = \psi \ .
\end{align*}

\noindent Therefore, the canonical gradients, provided by Theorem~\ref{thm:gradients}, are as follows: 
\begin{itemize}
	\item MSE risk gradient: $D_{R, P_0}(\psi)(o) = 2\{\psi(x,m,w) - \psi_0(x,m,w)\}$, 
	\item Cross-entropy risk gradient: $D_{R, P_0}(\psi)(o) = \displaystyle \frac{\psi(x, m, w)-\psi_0(x, m, w)}{\psi(x, m, w) ( 1- \psi(x, m, w))}$, 
	\item Constraint gradient: $D_{\Theta, P_0}(\psi)(o) = \kappa_0(x,m,w)$. 
\end{itemize}

\subsection{NDE: second-order conditions}
\label{app:nde_2nd}

For verifying second-order conditions, we follow the discussion in Appendix~\ref{app:second-order}. 

\subsubsection*{2nd-order conditions under MSE risk:} 

Given the canonical gradients, we can write:
\begin{align*}
	{\calL}_{P_0}(\psi, \lambda) &= \int \left\{\psi_0(x,m,w) - \psi(x,m,w)\right\}^2 dP_0(x, m, w) \\ 
	&\hspace{1.5cm} + \lambda \int \sum_{m}\left\{\psi(1, m, w) - \psi(0, m, w)\right\} f_{0, M}(m \mid 0, w) dP_{0,W}(w)\\ 
	\dot{\mathcal{L}}_{P_0}(\psi, \lambda) &=  2\left\{\psi(x,m, w) - \psi_0(x,m, w)\right\} + \lambda \kappa_0(x, m, w) \ ,  \\ 
	\ddot{\mathcal{L}}_{P_0}(\psi, \lambda) &= 2 \ . 
\end{align*}
For any $\psi = (\psi^1, \psi^0) \in {\bf \Psi}$ and $\lambda \in \openr$, the criterion \eqref{eq:borderedH_det_single} simplifies to $- 2\{\kappa_0(1, m, w)^2 + \kappa_0(0, m, w)^2\} < 0$, concluding that $\psi_{0, \lambda_0}$ is indeed the optimal minimizer. 

\subsubsection*{2nd-order conditions under cross-entropy risk:} 

According to Appendix~\ref{app:proofs_corollary}, $\psi_{0, \lambda}$ is given by:  
{\small 
	\begin{align*}
		\frac{ \lambda + \kappa_{0}(x, m, w)^{-1} - (2x-1)  \big\{ ( \lambda + \kappa_{0}(x, m, w)^{-1} )^2 - 4\lambda \kappa_{0}(x, m, w)^{-1}\psi_0(x, m, w) \big\}^{1/2}  }{2\lambda} \ .  
	\end{align*}
}%
Following the similar steps as in the ATE case, $\psi_{0, \lambda_0}$ is indeed the optimal minimizer of the penalized cross-entropy risk. 

\subsection{NDE: estimation} 
\label{app:nde_esti}

Let $\big|\big| f \big|\big| \coloneqq \left\{ \int f^2(o) \ dP(o) \right\}^{1/2}$ denote the $L^2(P_0)$-norm of the $P_0$-measurable function $f$, and let $\mathcal{X}$, $\mathcal{M}$, and $\mathcal{W}$ denote the domains of $X$, $M$, and $W$. We assume the following $L^2(P_0)$-consistency  for our nuisance estimates
\begin{align}
	&|| \pi_n - \pi || = \smallO(n^{-{1}/{a}}), \   
	|| \psi_n - \psi_0 || = \smallO(n^{-{1}/{b}}), \  
	|| f_{n, M} - f_{0, M} || = \smallO(n^{-{1}/{c}}) \ .  
	\label{eq:rate_nde}
\end{align}%

\subsubsection*{Estimation under MSE risk:}

We assume the following regularity conditions: 
\begin{equation}\label{eq:regularity_nde_mse}
	\begin{aligned}
		&\inf_{x \in \mathcal{X}, w \in \mathcal{W}} \pi_0(x \mid w) > 0 \ , \quad 
		&&\inf_{x \in \mathcal{X}, w \in \mathcal{W}} \pi_n(x \mid w) > 0  \ , \\
		&\sup_{x \in \mathcal{X}, w \in \mathcal{W}, m \in \mathcal{M}} \psi_0(x, m, w) < +\infty  \ , 
		\quad 
		&&\sup_{x \in \mathcal{X}, w \in \mathcal{W}, m \in \mathcal{M}} \psi_n(x, m, w) < +\infty \  , 
		\\
		&\inf_{x \in \mathcal{X}, w \in \mathcal{W}, m \in \mathcal{M}} f_{0, M}(m \mid x, w) > 0 \ , 
		\quad 
		&&\sup_{x \in \mathcal{X}, w \in \mathcal{W}, m \in \mathcal{M}} f_{0, M}(m \mid x, w) < +\infty  \ , 
		\\
		&\inf_{x \in \mathcal{X}, w \in \mathcal{W}, m \in \mathcal{M}} f_{n, M}(m \mid x, w) > 0 \ , 
		\quad 
		&&\sup_{x \in \mathcal{X}, w \in \mathcal{W}, m \in \mathcal{M}} f_{n, M}(m \mid x, w) < +\infty  \ . 
	\end{aligned}%
\end{equation}%
Following similar steps as in Appendix~\ref{app:ate_esti}, we can write: 
\begin{align*}
	\big|\big| \psi_{n,\lambda_n} &- \psi_{0, \lambda_0} \big|\big| \leq \big|\big| \psi_n - \psi_0 \big|\big|  
	+ 0.5 \big| \lambda_0 \big| \times \big|\big| \kappa_n - \kappa_0 \big|\big| 
	+ 0.5 \big| \lambda_n - \lambda_0 \big| \times  \big| \big| \kappa_n \big| \big| \ . 
\end{align*}
Given the regularity conditions in \eqref{eq:regularity_nde_mse}, we have $0.5 \big| \lambda_0 \big| \times \big|\big| \kappa_n - \kappa_0 \big|\big| \leq \mathfrak{C}_1 \big|\big| \pi_n - \pi_0 \big|\big| + \mathfrak{C}_2 \big|\big| f_{n, M} - f_{0, M} \big|\big|$ and $0.5 \big| \lambda_n - \lambda_0 \big| \times \big|\big| \kappa_n \big|\big| \leq \mathfrak{C}_3 \big|\big| \psi_n - \psi_0 \big|\big| $, where $\mathfrak{C}_1, \mathfrak{C}_2$, and $\mathfrak{C}_3$ are finite positive constants. 
Given the $L^2(P_0)$-consistency rates in \eqref{eq:rate_nde}, we get that  $ \big|\big| \psi_{n,\lambda_n} - \psi_{0, \lambda_0} \big|\big| = o_P(n^{\max\{-{1}/{a}, - {1}/{b}, -{1}/{c}\}})$.

\subsubsection*{Estimation under cross-entropy risk:}

In addition to the regularity conditions in \eqref{eq:regularity_nde_mse}, we assume $\lambda_n$ and $\lambda_0$ are bounded. Following similar steps as in Appendix~\ref{app:ate_esti}, we can write: 
{\small
	\begin{align*}
		\big|\big| \psi_{n,\lambda_n} &- \psi_{0, \lambda_0} \big|\big| 
		\leq \mathfrak{C}_1 \big|\big|  \pi_n - \pi_0 \big|\big| + \mathfrak{C}_2 \big|\big|  \psi_n - \psi_0 \big|\big| + \mathfrak{C}_3 \big|\big|  f_{n, M} - f_{0, M} \big|\big| \ , 
	\end{align*}
}%
where $\mathfrak{C}_1$, $\mathfrak{C}_2$, and $\mathfrak{C}_3$ are finite positive constants. Given the $L^2(P_0)$-consistency rates in \eqref{eq:rate_nde}, we get that  $ \big|\big| \psi_{n,\lambda_n} - \psi_{0, \lambda_0} \big|\big| = o_P(n^{\max\{-{1}/{a}, - {1}/{b}, -1/c\}})$.

\clearpage
\section{Constraint: equalized risk in cases (ERIC)}
\label{app:er_cases}

\subsection{ERIC: canonical gradients}
\label{app:er_cases_gradients}

The ERIC constraint under $\psi$ can be written down as: 
\begin{align*}
	\Theta_{P_0}(\psi) &= \E_{P_0}\left[ -\frac{(2X-1)}{p_{0,1}(X)} \psi_0(X,W) \log\{\psi(X,W)\} \right] \ . 
\end{align*}
According to the general constraint form in \eqref{eq:general_constraint}: 
\begin{align*}
	\kappa_0(x,w) =  -\frac{(2x-1)}{p_{0,1}(x)}\psi_0(x,w) \ , \quad \mbox{and} \quad \zeta_0(\psi) = \log(\psi) \ .
\end{align*}
For simplicity, we let $C_0(x) = (2x-1)/p_{0,1}(x)$. 

\noindent Therefore, the canonical gradients, provided by Theorem~\ref{thm:gradients}, are as follows: 
\begin{itemize}
	\item Cross-entropy risk gradient: $D_{R, P_0}(\psi)(o) = \displaystyle \frac{\psi(x, w)-\psi_0(x, w)}{\psi(x, w) ( 1- \psi(x, w))}$, 
	\item Constraint gradient: $D_{\Theta, P_0}(\psi)(o) = \displaystyle \frac{\kappa_0(x,w)}{\psi(x, w)} = - C_0(x) \frac{\psi_0(x,w)}{\psi(x,w)}$.  
\end{itemize}

For pedagogical clarity, we replicate the above results by presenting an explicit derivation of the constraint gradient in two ways: treating $\psi$ as a conditional mean and as a conditional probability mass function. We ultimately arrive at the exact same characterization for the constraint-specific path. However, the details are worth explicating as this reveals additional insight into the choice of Hilbert space where paths are embedded. 

\subsubsection{Path through the conditional mean}

Consider paths through $\psi(x,w) \in {\bf \Psi}$ at $\delta=0$ of the form 
\[
\left\{ \psi_{\delta, h} = \text{expit}\left[ \text{logit}\left\{\psi(x,w)\right\} + \delta h(x, w)  \right]: \delta \in \openr \right\} \ , 
\] 
with direction $\frac{d}{d\delta} \psi_{\delta,h}|_{\delta = 0} =  h(x, w) \psi(x,w) (1-\psi(x,w))$. We can explore all possible local directions at $\psi$ within the parameter space $\bf \Psi$ by letting $h(x,w)$ vary over real-valued function of $(x,w)$. Thus, $\tangent_{P}(\psi)$ could be taken as $L^2(P_{X, W}),$ that is the space of real-valued functions with bounded second moment defined on the support of $(X, W)$ implied by $P$. 

Let $L(\psi)(o) = -y\log \psi (z) - (1-y)\log \{1-\psi(z)\}$ be the negative log loss. For any $h \in \tangent_P(\psi)$, we have
\begin{align*}
	\frac{d}{d \delta} P_0L(\psi_{\delta, h})
	\Big|_{\delta = 0} 
	&= \int \left\{-y(1-\psi(x,w)) + (1-y)\psi(x,w) \right\} h(z) \ dP_0(o) \\
	&= \int \left\{-\psi_0(x,w)(1-\psi(x,w)) + (1-\psi_0(x,w))\psi(x,w) \right\} h(z) \ dP_0(o) \\
	&= \int \left\{\psi(x,w) -\psi_0(x,w)  \right\} h(z) \ dP_0(o) \ . 
\end{align*}
Thus, $D_{R, P_0}(\psi)(o) = \psi(x,w) -\psi_0(x,w)$.  Furthermore, for any $h \in \tangent_P(\psi)$, we have
\begin{align*}
	\frac{d}{d \delta} &\Theta_{P_0}(\psi_{\delta, h}) \Big|_{\delta = 0}  \\
	&= \frac{d}{d \delta} \Big\{\E_{P_0}(L(\psi_{\delta, h}) \mid X = 1, Y = 1) -
	\E_{P_0} (L(\psi_{\delta, h}) \mid X = 0, Y = 1) \Big\} \Big|_{\delta = 0} \\
	&=- \E_{P_0} \left\{ h(O) (1- \psi(1,W)) \mid X = 1, Y = 1 \right\} \\ 
	&\hspace{2cm} + \E_{P_0}\left\{ h(O)(1-\psi(0, W)) \mid X = 0, Y = 1\right\} \\
	&= \E_{P_0} \left[ \left\{  \frac{\mathbb{I}(X=0, Y=1)}{P_{0}(X=0, Y=1)} -   \frac{\mathbb{I}(X=1, Y=1)}{P_{0}(X=1, Y=1)}  \right\} {h(O)}{(1-\psi(X,W))}   \right] \\
	&= \int - \frac{2x-1}{p_{0,1}(x)} {\psi_0(x,w)} \ h(o) \ (1-\psi(x,w)) \ dP_{\psi, 0}(o) \ .
\end{align*}
Thus, $D_{\Theta, P_0}(\psi)(o) = - C_0(x) {\psi_0(x,w)}{(1-\psi(x,w))}$. Condition~\ref{eq:pathcondition-a} implies:
\begin{align*}
	\psi(x,w) -\psi_0(x,w) - \lambda C_0(x) {\psi_0(x,w)}{(1-\psi(x,w))} = 0 \\ 
	\rightarrow \psi_{0, \lambda}(x, w) = \psi_0(x, w) \ \frac{1 + \lambda\ C_0(x) }{1 + \lambda \ C_0(x) \ \psi_0(x, w)} \ . 
\end{align*} 

We note that although the path direction $\left.\frac{d}{d\delta}\psi_{\delta,h}\right|_{\delta=0}$ is indexed by $\psi$, one may suppress this dependence and carry out the derivations using a direction indexed solely by $h$. In this case, the risk and constraint gradients are both scaled by $(\psi(1 - \psi))^{-1}$. What matters for the constraint-specific path is that these gradients remain proportionally scaled, since the path is defined by the first-order condition $\text{(risk gradient)} + \lambda \times \text{(constraint gradient)} = 0$. 

\subsubsection{Path through the conditional probability mass function}

We now consider paths through conditional probability mass function $P(Y \mid X, W)$, rather than the conditional mean $P(Y = 1 \mid X, W)$. 

Let $\psi_0(y \mid x, w) = P_0(Y = y \mid X = x, W = w)$ be the unconstrained parameter defined as a conditional binary probability mass function. Note that $\psi_0 =  \argmin_{\psi \in {\bf \Psi}} P_0 L(\psi)$, where $L(\psi)(o) = -\log \psi (y \mid x, w)$ is the negative log-likelihood loss. The path $\{\psi_{\delta, h}(o) = \psi(o) \{1 + \delta h(o)\}: \delta \in \openr\}$ with direction $\frac{d}{d\delta} \psi_{\delta,h} |_{\delta = 0} = h(o)\psi(o)$ is through $\psi$ at $\delta = 0$. For this path to be within the parameter space $\bf \Psi$, $h(o)$ must be mean zero under $\psi$, $\sum_{y=0}^1 h(o)\ \psi(y \mid x, w) = 0$ for each $(x,w)$. Thus, exploring all possible local directions at $\psi$ within the parameter space $\bf \Psi$ would require that $h(o) \in L^2_0(P_{\psi, 0}),$ the space of bounded real-valued functions of $O$ with mean zero and finite second moment under the distribution $P_{\psi, 0} = \psi(y \mid x,w) P_{0,X,W}$. Thus, $\tangent_{P}(\psi)$ can be defined as $L^2_0(\psi)$, a subspace of $L^2_0(P_{\psi, 0})$ containing real-valued functions of $O$ defined on the support of $\psi$ with conditional mean zero given $(X,W)$ under $\psi$. We define the inner product on $L^2_0(P_{\psi, 0})$ as $\langle f, g \rangle = \int f(o)g(o)dP_{\psi,0}(o)$.  

To derive the canonical gradients $D_{R, P_0}(\psi)$ and $D_{\Theta, P_0}(\psi)$, we note that for any $h \in \tangent_P(\psi)$, we have
\begin{align*}
	\frac{d}{d \delta} P_0L(\psi_{\delta, h}) \Big|_{\delta = 0}  
	&= - P_0 \big(h(O)\big)  
	= - \int h(o) \ dP_0(o) \\
	&= - \int h(o) \times \frac{\psi_0(y \mid x, w)}{\psi(y \mid x, w)} \ dP_{\psi,0}(o) \\ 
	&= \int \frac{\psi(y \mid x, w) - \psi_0(y \mid x, w)}{\psi(y \mid x, w)} \ h(o) \ dP_{\psi, 0}(o) \ .
\end{align*}%
The last equality holds since $\int h(o) dP_{\psi, 0}(o) = 0$. Thus, $D_{R, P_0} (\psi) = (\psi - \psi_0) / \psi$. We also have  
\begin{align*}
	\frac{d}{d \delta} &\Theta_{P_0}(\psi_{\delta, h}) \Big|_{\delta = 0}  \\
	&= \frac{d}{d \delta} \Big\{\E_{P_0}(L(\psi_{\delta, h}) \mid X = 1, Y = 1) -
	\E_{P_0} (L(\psi_{\delta, h}) \mid X = 0, Y = 1) \Big\} \Big|_{\delta = 0} \\
	&=- \E_{P_0} (h(O)\mid X = 1, Y = 1) + \E_{P_0}(h(O) \mid X = 0, Y = 1) \\
	&= \E_{P_0} \bigg[ \Big\{  \frac{\mathbb{I}(X=0, Y=1)}{P_{0}(X=0, Y=1)} -   \frac{\mathbb{I}(X=1, Y=1)}{P_{0}(X=1, Y=1)}  \Big\} h(O)   \bigg] \\
	&= \int - \frac{y(2x-1)}{P_0(x,y)} \ \frac{\psi_0(y \mid x, w)}{\psi(y \mid x, w)}  \ h(o) \ dP_{\psi, 0}(o) \\
	&=  \int - \frac{2x-1}{P_0(x,Y=1)} \ \frac{\psi_0(1 \mid x, w)}{\psi(1 \mid x, w)} \ (y - \psi(1 \mid x, w)) \ h(o) \ dP_{\psi, 0}(o) \ . 
\end{align*}
The last equality holds since \[
\sum_{y=0}^1 \frac{y(2x-1)}{P_0(X = x, Y = y)} \ \frac{\psi_0(y \mid x, w)}{\psi(y \mid x, w)}  \ \psi(y \mid x, w) = \frac{2x-1}{p_{0,1}(x)} \psi_0(1 \mid x, w) \ .\] 
Thus, \[D_{\Theta, P_0}(\psi)(o) = - C_0(x) \displaystyle \frac{\psi_0(1 \mid x, w)}{\psi(1 \mid x,w)} \ \{y - \psi(1 \mid x, w)\} \ . \] 
Condition (\ref{eq:pathcondition-a}) yields: 
\begin{align*}
	\psi_{0, \lambda}(y \mid x, w) = \psi_0(y \mid x, w) \ \frac{1 + \lambda\ y \ C_0(x) }{1 + \lambda \ C_0(x) \ \psi_0(1 \mid x, w)} \ , 
\end{align*}
which agrees with the characterization of the constraint-specific path in \eqref{eq:ex_erica}.

\subsection{ERIC: second-order conditions}
\label{app:er_cases_2nd}

For verifying second-order conditions, we follow the discussion in Appendix~\ref{app:second-order}. 

First note that $\psi_{0, \lambda}$ in \eqref{eq:ex_erica} should be a valid probability mass function. Thus, for all $(x,w)$, we should have 
\begin{align*}
	0 \ \le \psi_0(x, w) \ \frac{1 + \lambda C_0(x) }{1 + \lambda C_0(x) \psi_0(x, w)} \ \le \ 1 \ . 
\end{align*} 
This is guaranteed only if $-P_0(X=1, Y=1) < \lambda < P_0(X=0, Y=1)$. 

Given the canonical gradients $D_{R, P_0}(\psi)$ and $D_{\Theta, P_0}(\psi)$, we have:
\begin{align*}
	\mathcal{L}_{P_0}(\psi, \lambda) &=  \int \left[ \psi_0(x,w) \log \psi(x,w) - \left\{1-\psi_0(x,w)\right\} \log \left\{1-\psi(x,w)\right\} \right]  dP_{0}(x,w) \\
	&\hspace{0.25cm} + \lambda \Big\{ \E_P[L(\psi)(O) \mid X=1, Y=1]  - \E_P[L(\psi)(O) \mid X=0, Y=1] \Big\} \ , \\ 
	\dot{\mathcal{L}}_{P_0}(\psi, \lambda) &=  \frac{1-\psi_0(x, w)}{1- \psi(x, w)} - \frac{\psi_0(x, w)}{\psi(x, w)}  - \lambda C_0(x) \frac{\psi_0(x,w)}{\psi(x,w)} \ , \\
	\ddot{\mathcal{L}}_{P_0}(\psi, \lambda) &=  \frac{1 - \psi_0(x, w)}{(1-\psi(x,w))^2} + \frac{\psi_0(x,w)}{\psi(x,w)^2} \big( 1+ \lambda C_0(x) \big) \ .
\end{align*} %
For any given $\psi = (\psi^1, \psi^0) \in {\bf \Psi}$ and $\lambda \in \openr$, the criterion \eqref{eq:borderedH_det_single} simplifies to:  
\begin{align*}
	&- D^2_{\Theta, P}(\psi^1) \times \ddot{\calL}_P(\psi^0, \lambda)
	- D^2_{\Theta, P}(\psi^0) \times \ddot{\calL}_P(\psi^1, \lambda)  \\
	&\hspace{1cm} = - C^2_0(1) \frac{\psi_0(1, w)^2}{\psi(1, w)^2} \frac{1 - \psi_0(0, w)}{(1-\psi(0,w))^2} 
	\\
	&\hspace{1.5cm} - C^2_0(0) \frac{\psi_0(0, w)^2}{\psi(0, w)^2} \frac{1 - \psi_0(1, w)}{(1-\psi(1,w))^2} 
	\\
	&\hspace{1.5cm}- C^2_0(1) \frac{\psi^2_0(1, w)}{\psi^2(1, w)} \times  \frac{\psi_0(0, w)}{\psi^2(0, w)} \left\{ 1 + \lambda C_0(0) \right\} \\
	&\hspace{1.5cm} - C^2_0(0) \frac{\psi^2_0(0, w)}{\psi^2(0, w)} \times  \frac{\psi_0(1, w)}{\psi^2(1, w)} \left\{ 1 + \lambda C_0(1) \right\}  \ . 
\end{align*}
Given this expression, it is straightforward to show that condition \eqref{eq:borderedH_det_single} is satisfied so long as $-p_{0,1}(1) < \lambda < p_{0,1}(0)$. This concludes that $\psi_{0, \lambda_0}$ is the optimal minimizer of the penalized cross-entropy risk.  

\subsection{ERIC: estimation} 
\label{app:er_cases_est}

Let $\big|\big| f \big|\big| \coloneqq \left\{ \int f^2(o) \ dP(o) \right\}^{1/2}$ denote the $L^2(P_0)$-norm of the $P_0$-measurable function $f$, and let $\mathcal{X}$ and $\mathcal{W}$ denote the domain of $X$ and $W$. 

We assume the following regularity conditions: 
\begin{equation}\label{eq:regularity_er_cases_cross}
	\begin{aligned}
		&\inf_{x \in \mathcal{X}} p_{0,1}(x) > 0 \ , \quad 
		&&\inf_{x \in \mathcal{X}} p_{n,1}(x) > 0  \ , \\
		&\sup_{x \in \mathcal{X}, w \in \mathcal{W}} \psi_0(x, w) < +\infty  \ , 
		\quad 
		&&\sup_{x \in \mathcal{X}, w \in \mathcal{W}} \psi_n(x, w) < +\infty \  . 
	\end{aligned}%
\end{equation}%
Additionally, given the restrictions on the range of $\lambda_0$, provided in Appendix~\ref{app:er_cases_2nd}, $\lambda_0$ and $\lambda_n$ are both bounded within $\min \{-p_{0,1}(1), -p_{n, 1}(1)\}$ and $\max \{p_{0,1}(0), p_{n, 1}(0)\}$. 

Given the above regularity conditions, we can write: 
\begin{align*}
	\big|\big| \psi_{n,\lambda_n} - \psi_{0, \lambda_0} \big|\big| 
	&= \big|\big|  \frac{\psi_n(x, w) + \lambda_n C_n(x) \psi_n(x, w) }{1 + \lambda_n C_n(x) \psi_n(x, w)} -  \frac{\psi_0(x, w) + \lambda_0 C_0(x) \psi_0(x, w) }{1 + \lambda C_0(x) \psi_0(x, w)}  \big|\big|   \\
	&= \big|\big| \frac{\psi_n - 1}{1 + \lambda_n C_n \psi_n} - \frac{\psi_0 - 1}{1 + \lambda_0 C_0 \psi_0} \big| \big| \ \\
	&\leq \big| \big| \psi_n - \psi_0 \big| \big| \ . 
\end{align*}

\clearpage
\section{Additional causal constraints}
\label{app:cate}

\subsection{Conditional average treatment effect}
\label{app:cate_2dim}

Suppose $Z = (X, W, L)$ where $(W, L)$ are baseline covariates. Let $O = (Z, Y) \sim P_0 \in \mathcal{M}$, and $L(\psi)(o) = \{y - \psi(l, w, x)\}^2$ be the squared-error loss. We consider a constraint where the treatment effect of $X$ on $Y$ is null in all sub-populations defined by levels of the covariate $L$. Let: 
\begin{align}
	\Theta_{P_0,\ell}(\psi) &= \int \left\{ \psi(1, w, L=\ell) - \psi(0, w, L=\ell) \right\} dP_0(w \mid L = \ell )  \ . 
	\label{eq:cate}
\end{align}
Under sufficient causal identification assumptions, $\Theta_{P_0,\ell}(\psi)$ defines the \textit{conditional average treatment effect} (CATE) of $X$ on $Y$, stratified by $\ell$. For simplicity, assume $L$ is binary. We define $\Theta_{P_0}(\psi)$ as a two-dimensional vector of constraints: 
\begin{align*}
	\Theta_{P_0}(\psi) = (\Theta_{P_0, \ell=1}(\psi), \Theta_{P_0, \ell=0}(\psi))^\top \ . 
\end{align*}

We can write $\Theta_{P_0, \ell}(\psi)$ in \eqref{eq:cate} as follows: 
\begin{align*}
	\Theta_{P_0, \ell}(\psi) &= \E_{P_0}\left[ \displaystyle \frac{1}{P_0(L)} \frac{2X-1}{\pi_0(X \mid W, L)} \psi(X, W, L) \right] \ . 
\end{align*}%
Therefore, according to the general constraint form in \eqref{eq:general_constraint}: 
\begin{align*}
	\kappa_0(x, w, \ell) = \frac{1}{P_0(\ell)} \frac{2x-1}{\pi_0(x \mid w, \ell)}  \quad  \mbox{and} \quad \zeta_0(\psi) = \psi \ .
\end{align*}

\noindent Consequently, the canonical gradients, provided by Theorem~\ref{thm:gradients}, are as follows:
\begin{itemize}
	\item MSE risk gradient: $D_{R, P_0}(\psi)(o) = 2\{\psi(x,w, \ell) - \psi_0(x,w, \ell)\}$, 
	\item The two-dimensional constraint gradient: \\ $D_{\Theta, P_0}(\psi)(o) = \big(\kappa_0(x,w, \ell=1), \kappa_0(x,w, \ell=0)\big)^\top$. 
\end{itemize}

According to Appendix~\ref{app:methods-multi}, for $\lambda \in \openr^2$, the path $\{\psi_{0, \lambda} : \lambda \in \openr^2\}$ must satisfy 
\begin{align}
	D_{R, P_0}(\psi_{0, \lambda})(o) + \lambda^\top D_{\Theta, P_0}(\psi_{0, \lambda})(o) = 0  \ , 
\end{align}%
which yields: 
\begin{align*}
	2\{\psi_{0, \lambda}(x,w, \ell) - \psi_0(x,w,\ell)\} + \frac{1}{P_0(\ell)} \frac{2x - 1}{\pi_0(x \mid w, \ell)} \left\{ \lambda_1 \ell + \lambda_0 (1-\ell) \right\} = 0 \ . 
\end{align*}%
The above implies:  
\begin{align*}
	2\{\psi_{0, \lambda_z}(x,w,\ell) - \psi_0(x,w,\ell)\} &+ \lambda_\ell  \frac{1}{P_0(\ell)} \frac{2x - 1}{\pi_0(x \mid w, \ell)}  = 0  \ , \\ 
	\rightarrow \  \psi_{0, \lambda_z}(x,w,\ell) &= \psi_0(x,w,\ell) - \lambda_\ell  \frac{1}{2P_0(\ell)} \frac{2x - 1}{\pi_0(x \mid w, \ell)}   \ , \\
	\rightarrow \  \psi_{0, \lambda_\ell}(x,w,\ell) &= \psi_0(x,w,\ell) - 0.5 \lambda_\ell  \kappa_0(x,w, \ell)   \ . 
\end{align*}

We can also get a closed-form solution for $\lambda_\ell$ as: 
{\small 
	\begin{align} 
		\lambda_\ell =  \frac{2 \ \Theta_{P_0, \ell}(\psi_0)}{ P_0(\ell)^{-1} \displaystyle \int \left\{ \pi_0(X=1 \mid w, \ell)^{-1} + \pi_0(X=0 \mid w, \ell)^{-1} \right\} dP_{0}(w \mid \ell)} \ . 
	\end{align}%
}
Similarly, the second‑order conditions and estimation discussions follow what we covered for the ATE and NDE examples.

\subsection{Average treatment effect on the treated/control} 

Certain applications require controlling the causal effect within a specific treatment group. For example, clinical trials often demand that the estimated response in each arm meets predefined benchmarks. We accommodate this by imposing restrictions on the effect size within the subpopulation that received the treatment (or the control/placebo). 

Suppose $Z = (X, W)$, and $O = (Z, Y) \sim P_0 \in \mathcal{M}$. Let: 
\begin{align}
	\Theta_{P_0,x}(\psi) &= \int \left\{ \psi(1, w) - \psi(0, w) \right\} dP_0(w \mid X = x)  \ . 
	\label{eq:att}
\end{align}
Under sufficient causal identification assumptions, $\Theta_{P_0,x=1}(\psi)$ defines the \textit{average treatment effect among the treated} (ATT) and $\Theta_{P_0,x=0}(\psi)$ defined  the \textit{average treatment effect among the control} (ATC).  

\vspace{0.35cm}
\noindent \textbf{ATT constraint:} We can rewrite the ATT as: 
\begin{align*}
	\Theta_{P_0,x=1}(\psi)&= \E_{P_0}\left[ \displaystyle  \frac{2X-1}{P_0(X=1)} \frac{\pi_0(X=1 \mid W)}{\pi_0(X \mid W)}  \psi(X, W) \right] \ . 
\end{align*}%
Therefore, according to the general constraint form in \eqref{eq:general_constraint}: 
\begin{align*}
	\kappa_0(x, w) = \frac{2x-1}{P_0(X=1)} \frac{\pi_0(X=1 \mid w)}{\pi_0(x \mid w)}   \quad  \mbox{and} \quad \zeta_0(\psi) = \psi \ .
\end{align*}
Consequently, by Theorem~\ref{thm:gradients}, the ATT constraint gradient is $\kappa_0(x, w)$. 

\vspace{0.35cm}
\noindent \textbf{ATC constraint:} We can rewrite the ATC as: 
\begin{align*}
	\Theta_{P_0,x=0}(\psi)&= \E_{P_0}\left[ \displaystyle   \frac{2X-1}{P_0(X=0)} \frac{\pi_0(X=0 \mid W)}{\pi_0(X \mid W)} \psi(X, W) \right] \ . 
\end{align*}%
Therefore, according to the general constraint form in \eqref{eq:general_constraint}: 
\begin{align*}
	\kappa_0(x, w) = \frac{2x-1}{P_0(X=0)} \frac{\pi_0(X=0 \mid w)}{\pi_0(x \mid w)}  \quad  \mbox{and} \quad \zeta_0(\psi) = \psi \ .
\end{align*}
Consequently, by Theorem~\ref{thm:gradients}, the ATC constraint gradient is $\kappa_0(x, w)$.

The closed-form characterization for $\psi^*_n$, the second-order optimality conditions, and the estimation strategy for both ATT and ATC constraints follow directly from the same approach we used for the earlier causal examples.


\clearpage
\section{Additional risk constraints}
\label{app:er_overall} 

\subsection{Overall equalized MSE risk}
\label{app:er_overall_mse} 

Let $L(\psi)(o) = \{y - \psi(x, w)\}^2$ be squared-error loss and define $\Theta_{P}(\psi) = \E_{P}\{L(\psi)(O) \mid X = 1\} - \E_P\{L(\psi)(O) \mid X = 0\}$ as the overall equalized mean squared error between the two groups defined by sensitive characteristic $X.$ The constraint on $\psi$ defined via $\Theta_{P_0}(\psi) = 0$ is similar to the \textit{demographic parity} notion of fairness.

By Theorem~\ref{thm:gradients}, the gradient of the risk is $D_{R, P_0}(\psi)(o) = 2\{\psi(x,w) - \psi_0(x,w)\}$.  We can rewrite the constraint as: 
\begin{align*}
	\Theta_P(\psi) &= \E_{P}\left[ -\frac{2X-1}{p_0(X)} \left[ \sigma^2_0(X,W) + \{ \psi_0(X,W) - \psi(X,W) \}^2 \right] \right]\ . 
\end{align*}
According to the general constraint form in \eqref{eq:general_constraint}: 
\begin{align*}
	\kappa_0(Z) =  -\frac{2X-1}{p_0(X)} \ , \ \mbox{and} \ \zeta_0(\psi) = \sigma_0^2 + \{ \psi_0 - \psi\}^2 \ ,
\end{align*}
where $\sigma^2_0(X,W)$ is the conditional variance of $Y$ given $X$ and $W$. By Theorem~\ref{thm:gradients}, $D_{\Theta, P_0}(\psi) = -2\kappa_0 (\psi_0(x,w) - \psi(x,w)).$ 

For pedagogical clarity, we replicate the above results by presenting an explicit derivation of the constraint gradient. Let  $C_0(x) = (2x-1)/p_0(x)$. For any $h \in \tangent_P(\psi)$, 
\begin{align*}
	\frac{d}{d \delta} &\Theta_{P_0}(\psi_{\delta, h}) \Big|_{\delta = 0}  
	= \frac{d}{d \delta} \left[\E_{P_0}\{L(\psi_{\delta, h}) \mid X = 1\} -
	\E_{P_0} \{L(\psi_{\delta, h}) \mid X = 0\} \right] \Big|_{\delta = 0} \\
	&=\E_{P_0} \left[ 2\{Y - \psi(X,W)\} h(X,W)\mid X = 1\right] - \E_{P_0}\left[ 2\left\{Y - \psi(X,W)\right\} h(X,W) \mid X = 0\right] \\
	&= \E_{P_0} \left[ 2 \{\psi_0(X, W) - \psi(X,W)\} \left\{  \frac{\mathbb{I}(X=1)}{p_{0}(1)} -   \frac{\mathbb{I}(X=0)}{p_{0}(0)}  \right\} h(X,W)  \right] \\
	&=  \int 2 \frac{2x-1}{p_0(x)} \{\psi_0(x,w) - \psi(x,w)\} h(x,w) \ dP_0(x,w) \ .
\end{align*}
Thus, $D_{\Theta, P_0}(\psi)(o) = 2 C_0(x) \{\psi_0(x,w) - \psi(x,w)\}$. 

Condition~\eqref{eq:pathcondition-a} implies
\begin{align}
	\{1 + \lambda C_0(x)\} \{\psi_0(x,w) - \psi_{0, \lambda}(x, w) \}  = 0 \ , 
	\label{eq:ex_overall_erica}
\end{align}
where $C_0(x) = (2x-1)/p_0(x)$. This equation implies $\psi_{0, \lambda} = \psi_0$ is a stationary point in the Lagrangian optimization problem \eqref{eq:lagrangedef}. Yet, $\psi_0$ does not necessarily constitute a valid solution $\Theta_{P_0}(\psi_0) = 0$. If it were a solution, then $\lambda_0 = 0$ and the problem reduces to unconstrained optimization. Otherwise equation~\eqref{eq:ex_overall_erica} and the constraint equation imply
\begin{align}
	&\hspace{1.5cm} \left\{ 1 + \frac{\lambda}{p_{0}(1)} \right\} \left\{ \psi_0(1,w) - \psi(1,w) \right\} =0  \ , \notag \\
	&\hspace{1.5cm} \left\{ 1 - \frac{\lambda}{p_{0}(0)}\right\} \left\{\psi_0(0,w) - \psi(0,w)\right\} =0  \ ,  \label{eq:ex_overall_erica_3eqs} \\
	&\int \left[ \frac{\pi_0(1 \mid w)}{p_{0}(1)} \left\{ \psi_0(1,w) - \psi(1,w) \right\}^2 - \frac{\pi_0(0 \mid w)}{p_{0}(0)} \left\{\psi_0(0,w) - \psi(0,w)\right\}^2 \right] dP(w) \notag  \\ 
	&\hspace{2cm}= \int \left\{\frac{\pi_0(0 \mid w)}{p_{0}(0)} \sigma^2_0(0, w) - \frac{\pi_0(1 \mid w)}{p_{0}(1)} \sigma^2_0(1, w) \right\} dP(w)  \ ,  \label{eq:eqrisk_soln_integral_equation}
\end{align}
where $\sigma^2_0(x, w)$ denotes the conditional variance of $Y$ given $X = x, W = w$. This system of equations can be solved by considering the two cases defined by whether the right hand side of \eqref{eq:eqrisk_soln_integral_equation} is positive versus negative. 

If the right-hand side of \eqref{eq:eqrisk_soln_integral_equation} is positive, then the system of equations implies an optimal solution can be achieved by setting $\psi(0,w) = \psi_0(0, w)$. This in turn implies that $\lambda_0 = - p_{0}(1)$ and that $\psi_0^*(1,w)$ is any root in $\psi$ of the equation
\begin{align}
	\int \left( \frac{\pi_0(1 \mid w)}{p_{0}(1)} \left[\left\{\psi_0(1,w) - \psi(1,w)\right\}^2 + \sigma_0(1, w)\right] - \frac{\pi_0(0 \mid w)}{p_{0}(0)} \sigma^2_0(0, w) \right) dP(w) \ . \label{eq:eqrisk_soln_integral_equation2}
\end{align}
There will in general be infinitely many such roots. However, we can show that $R_{P_0}(\psi) = \E[ \{ 1+\lambda_0 C_0(X) \} \{ Y - \psi_0(X,W)\} ]^2$, which does not depend on $\psi(1,\cdot)$. 
Thus, any root in the above equation will have the same penalized risk and therefore represents a valid solution to the optimization problem. A closed form expression for a solution $\psi_0^*(1,w)$ can be obtained by setting the integrand of \eqref{eq:eqrisk_soln_integral_equation2} to 0 and solving,
\begin{align*}
	\psi_0^*(1,w) = \psi_0(1,w) + \left\{\frac{p_{0}(1)}{p_{0}(0)} \frac{\pi_0(0 \mid w)}{\pi_0(1 \mid w)} \sigma^2_0(0, w) - \sigma^2_0(1, w) \right\}^{1/2} \ . 
\end{align*}

If the right-hand side of \eqref{eq:eqrisk_soln_integral_equation} is negative, the above arguments can be made almost identically. Such an analysis concludes that $\lambda_0 = p_0(0)$, $\psi_{0}^*(1,w)=\psi_0(1,w)$, and one solution for $\psi_0^*(0,w)$ is \begin{align*}
	\psi_0^*(0,w) = \psi_0(0,w) + \left\{\frac{p_{0}(0)}{p_{0}(1)} \frac{\pi_0(1 \mid w)}{\pi_0(0 \mid w)} \sigma^2_0(1, w) - \sigma^2_0(0, w) \right\}^{1/2} \ . 
\end{align*}

Thus to generate estimates of $\psi_{0}^*$, we could obtain estimates of $\eta_n = (\psi_n, \pi_n, p_{n}, \sigma^2_n)$ of $\eta_0=(\psi_{0}, \pi_0, p_0, \sigma^2_0)$. Here $\psi_n$ and $\pi_n$ can be obtained via any form of unconstrained learning, $p_n(1) = n^{-1} \sum_{i=1}^n X_i$ and $p_n(0) = 1 - p_n(1)$. The estimate $\sigma^2_n$ can be obtained via any approach appropriate for conditional variance estimation. We define the plug-in estimator of the constraint for a particular $\psi$ as \[
\Theta_n(\psi) = \frac{1}{n}\sum\limits_{i=1}^n \frac{2X_i-1}{p_{n}(X_i)} L(\psi)(O_i) \ .
\]
Next, we select a threshold $\epsilon$ and check whether $|\Theta_n(\psi_n)| < \epsilon$. If so, we report $\psi_n^* = \psi_n$. Otherwise, we compute the sign of \[
\frac{1}{n} \sum_{i=1}^n \left\{\frac{\pi_n(0 \mid W_i)}{p_{n}(0)} \sigma^2_n(Y \mid 0, W_i) - \frac{\pi_n(1 \mid W_i)}{p_{n}(1)} \sigma^2_n(Y \mid 1, W_i) \right\} \ .
\]
If positive, then we set $\psi_{n}^*(0,w) = \psi_n(0,w)$, and 
\begin{align*}
	\psi_{n}^*(1,w) = \psi_n(1,w) + \left[\frac{p_n(1)}{p_n(0)} \frac{\pi_n(0 \mid w)}{\pi_n(1 \mid w)} \sigma^2_n(Y \mid 0, w) - \sigma^2_n(Y \mid 1, w) \right]^{1/2} \ . 
\end{align*}
If negative, then we set $\psi_n^*(1,w) = \psi_n(1,w)$, and 
\begin{align*}
	\psi_{n}^*(0,w) = \psi_n(0,w) + \left[\frac{p_n(0)}{p_n(1)} \frac{\pi_n(1 \mid w)}{\pi_n(0 \mid w)} \sigma^2_n(Y \mid 1, w) - \sigma^2_n(Y \mid 0, w) \right]^{1/2} \ .
\end{align*}

\vspace{0.15cm}
\noindent {\bf Second-order conditions.} 
Given $D_{R, P_0}(\psi)$ and $D_{\Theta, P_0}(\psi)$, we have:
\begin{align*}
	\mathcal{L}_{P_0}(\psi, \lambda) &=  \int \{\psi_0(x,w) - \psi(x,w)\}^2 \{1 + \lambda C_0(x)\}  dP_{0}(x,w) \ ,  \\ 
	\dot{\mathcal{L}}_{P_0}(\psi, \lambda) &=  - 2 \{\psi_0(x,w) - \psi(x,w)\} \{1 + \lambda C_0(x)\} \ , \\
	\ddot{\mathcal{L}}_{P_0}(\psi, \lambda) &=  2 \ .
\end{align*} %
For any given $\psi = (\psi^1, \psi^0) \in {\bf \Psi}$ and $\lambda \in \openr$, the criterion \eqref{eq:borderedH_det_single} simplifies to:  
\begin{align*}
	&- D^2_{\Theta, P}(\psi^1) \times \ddot{\calL}_P(\psi^0, \lambda)
	- D^2_{\Theta, P}(\psi^0) \times \ddot{\calL}_P(\psi^1, \lambda)  \\
	&\hspace{1cm} = - 2 \frac{(\psi^1 - \psi_0^1)^2}{p_{0}(x)^2} -  2 \frac{(\psi^0 - \psi_0^0)^2}{p_{0}(0)^2} < 0  \ . 
\end{align*}
This concludes that $\psi_{0, \lambda_0}$ is the optimal minimizer of the penalized MSE risk.

\subsection{Overall equalized cross-entropy risk}
\label{app:er_overall_cross}

Suppose $O = (W, X, Y) \sim P_0 \in \mathcal{M}$. We are interested in the mappings $\psi$ on the support of $(X, W)$. We consider the cross-entropy risk and let the constraint be the equalized risk, $\Theta_{P}(\psi) = \E_{P}\{L(\psi)(O) \mid X = 1\} - \E_P\{L(\psi)(O) \mid X = 0\}$. 

By Theorem~\ref{thm:gradients}, $D_{R, P_0}(\psi) = (1-\psi_0)/(1- \psi) - \psi_0/\psi$. We can rewrite the constraint as: 
\begin{align*}
	\Theta_P(\psi) &= \E_{P}\left[ \frac{1-2X}{p_0(X)} \left[\psi_0(X,W)\log \psi(X,W) + \{1 - \psi_0(X,W)\} \log \{1-\psi(X,W)\}\right]\right] \ . 
\end{align*}
According to the general constraint form in \eqref{eq:general_constraint}: 
\begin{align*}
	\kappa_0(x) =  -\frac{2x-1}{p_0(x)} \ , \ \mbox{and} \ \zeta_0(\psi) = \psi_0\log \psi + (1 - \psi_0) \log (1-\psi) \ . 
\end{align*} 
By Theorem~\ref{thm:gradients}, $D_{\Theta, P_0}(\psi) = - \kappa_0(x) D_{R, P_0}(\psi).$

For pedagogical clarity, we replicate the above results by presenting an explicit derivation of the constraint gradient. Let  $C_0(x) = - \kappa_0(x) = (2x-1)/p_0(x)$. For any $h \in \tangent_P(\psi)$, 
\begin{align*}
	\frac{d}{d \delta} &\Theta_{P_0}(\psi_{\delta, h}) \Big|_{\delta = 0}  \\
	&=	\frac{d}{d \delta} \Big\{\E_{P_0}(L(\psi_{\delta, h}) \mid X = 1) -
	\E_{P_0} (L(\psi_{\delta, h}) \mid X = 0) \Big\} \Big|_{\delta = 0} \\
	&= \E_{P_0} \bigg[ \Big\{  \frac{\mathbb{I}(X=1)}{P_{0}(X=1)} -   \frac{\mathbb{I}(X=0)}{P_{0}(X=0)}  \Big\} \Big\{ \frac{1-Y}{1-\psi(X,W)} - \frac{Y}{\psi(X,W)} \Big\} h(O)   \bigg] \\
	&= \int \frac{2x-1}{P_0(x)} \Big\{ \frac{1-\psi_0(x,w)}{1-\psi(x,w)}  - \frac{\psi_0(x,w)}{\psi(x,w)} \Big\} \ h(o) \ dP_{0}(x,w) \ . 
\end{align*}
Thus, $D_{\Theta, P_0}(\psi)(o) = \displaystyle C_0(x) \Big\{ \frac{1-\psi_0(x,w)}{1-\psi(x,w)}  - \frac{\psi_0(x,w)}{\psi(x,w)} \Big\} $. Condition \eqref{eq:pathcondition-a} implies
\begin{align*}
	\{1 + \lambda C_0(x)\}  \left\{\frac{1-\psi_0(x,w)}{1-\psi_{0, \lambda}(x, w)} -  \frac{\psi_0(x,w)}{\psi_{0, \lambda}(x, w)} \right\}  = 0 \ .  
\end{align*} 
Along with the constraint equation, the above implies: 
\begin{align}
	&\hspace{2cm} \left\{1 + \frac{\lambda}{p_{0}(x)}\right\}  \left\{\frac{1-\psi_0(1,w)}{1-\psi_{0, \lambda}(1, w)} -  \frac{\psi_0(1,w)}{\psi_{0, \lambda}(1, w)} \right\}  = 0  \ ,  \notag \\
	&\hspace{2cm} \left\{1 - \frac{\lambda}{p_{0}(0)} \right\} \left\{\frac{1-\psi_0(0,w)}{1-\psi_{0, \lambda}(0, w)} -  \frac{\psi_0(0,w)}{\psi_{0, \lambda}(0, w)} \right\}  = 0 \ , \label{eq:erica_cross_3eqs} \\ 
	&\int \frac{\pi_0(1 \mid w)}{p_{0}(x)} \left\{ - \psi_0(1,w) \log \psi(1,w) - (1-\psi_0(1,w)) \log (1- \psi(1,w)) \right\} dP(w)  \notag  \\
	&\hspace{0.5cm} = \int \frac{\pi_0(0 \mid w)}{p_{0}(0)} \left\{ - \psi_0(0,w) \log \psi(0,w) - (1-\psi_0(0,w)) \log (1- \psi(0,w)) \right\} dP(w)  \notag  \ . 
\end{align}%
The above system of equations yield two solutions: 
\begin{enumerate}
	\item \  $\psi(0, w) = \psi_0(0, w),$ with $\lambda_0 = -p_{0}(x)$ and $\psi(1,w)$ chosen such that 
	{\small 
		\begin{align*}
			&\int \frac{\pi_0(1 \mid w)}{p_{0}(x)} \left\{ - \psi_0(1,w) \log \psi(1,w) - (1-\psi_0(1,w)) \log (1- \psi(1,w)) \right\} dP(w)  \notag  \\
			&\hspace{0.5cm} = \int \frac{\pi_0(0 \mid w)}{p_{0}(0)} \left\{ - \psi_0(0,w) \log \psi_0(0,w) - (1-\psi_0(0,w)) \log (1- \psi_0(0,w)) \right\} dP(w) \ , 
		\end{align*}
	}
	\item  \ $\psi(1, w) = \psi_0(1, w),$ with $\lambda_0 = p_{0}(0)$ and $\psi(0,w)$ chosen such that 
	{\small 
		\begin{align*}
			&\int \frac{\pi_0(0 \mid w)}{p_{0}(0)} \left\{ - \psi_0(0,w) \log \psi(0,w) - (1-\psi_0(0,w)) \log (1- \psi(0,w)) \right\} dP(w)  \notag  \\
			&\hspace{0.5cm} = \int \frac{\pi_0(1 \mid w)}{p_{0}(x)} \left\{ - \psi_0(1,w) \log \psi_0(1,w) - (1-\psi_0(1,w)) \log (1- \psi_0(1,w)) \right\} dP(w)  \ . 
		\end{align*}
	}
\end{enumerate}
Note that since the penalized risk at $\lambda_0$
{\small 
	\begin{align*}
		&R_{P_0}(\psi) + \lambda_0 \Theta_{P_0}(\psi) 
		\\
		&\hspace{0.5cm}= \E[ (1 + \lambda_0 C_0(X)) (- \psi_0(X,W) \log \psi(X,W) - (1-\psi_0(X,W)) \log (1- \psi(X,W)) ] \\
		&\hspace{0.5cm}= \E\bigg[ (1 + \lambda_0 C_0(X)) \Big\{ - \psi_0(X,W) \log \frac{\psi(X,W)}{\psi_0(X,W)} - (1-\psi_0(X,W)) \log \frac{1- \psi(X,W)}{1-\psi_0(X,W)} \Big\} \bigg]   \\ 
		&\hspace{1cm} + \E\bigg[ (1 + \lambda_0 C_0(X)) \Big\{ - \psi_0(X,W) \log \psi_0(X,W) - (1-\psi_0(X,W)) \log (1-\psi_0(X,W)) \Big\} \bigg] \\
		&\hspace{0.5cm}= \E\bigg[ (1 + \lambda_0 C_0(X)) \Big\{ - \psi_0(X,W) \log \psi_0(X,W) - (1-\psi_0(X,W)) \log (1-\psi_0(X,W)) \Big\} \bigg] \ . 
	\end{align*}
}%
does not depend on $\psi(x,w)$. Thus, any $\psi(x,w)$, as long as it satisfies \eqref{eq:erica_cross_3eqs}, should be a minimizer to the penalized risk. 

\vspace{0.35cm} 
\noindent \textbf{Remark.} 
An observation on the general form of the constraint given in \eqref{eq:general_constraint} is that it helps explain the non-uniqueness of the solution for the equalized risk and anticipate other settings where this non-uniqueness may arise. In particular, in the equalized risk examples, we have that $\dot{\zeta}(\psi) = D_{R,P_0}(\psi)$. More generally, we can imagine a situation where $\dot{\zeta}(\psi) = a D_{R,P_0}(\psi)$ for some real-valued constant $a$. In this case, equation \eqref{eq:pathcondition-a} implies that $D_{R, P_0}(\psi) ( 1 + \lambda a \kappa_0 ) = 0$. This equation can be trivially solved for all $\lambda$ by letting $\psi = \psi_0$, since $\psi_0$ is a minimizer of the unconstrained risk $R_{P_0}$, which in turn implies that $D_{R, P_0}(\psi_0) = 0$. Thus, deriving a valid solution to the Lagrangian system of equations necessarily requires explicit consideration for the trivial case that $\psi_0$ satisfies $\Theta_{P_0}(\psi_0) = 0$. In our other specific examples, solving \eqref{eq:pathcondition-a} yielded a closed for solution for $\psi_{0,\lambda}$ that was equal to $\psi_0$ only if $\lambda = 0$.

The lack of uniqueness in the equalized risk setting can also be explained by the fact that if $\dot{\zeta}(\psi) = a D_{R,P_0}(\psi)$, then the gradients of the constraint and the risk are proportional. Thus, we can infer that the risk and the constraint have the same level sets, since gradients are orthogonal to the level set. The implication is that \emph{any} $\psi$ that satisfies the constraint will necessarily have locally minimal risk in $\bm{\Psi}$. This provides a geometric view of the reasoning behind the existence of multiple solutions in this problem.

\vspace{0.35cm}
\noindent {\bf Second-order conditions.}  
Given $D_{R, P_0}(\psi)$ and $D_{\Theta, P_0}(\psi)$, we have:
\begin{align*}
	\mathcal{L}_{P_0}(\psi, \lambda) &=  \int \left\{- \psi_0(x,w) \log \psi(x,w) - (1-\psi_0(x,w)) \log (1-\psi(x,w)) \right\} \\ 
	&\hspace{2cm} \hspace{2cm} \times (1 + \lambda C_0(x))  \ dP_{0}(x,w) \ ,  \\ 
	\dot{\mathcal{L}}_{P_0}(\psi, \lambda) &=  \left\{\frac{1-\psi_0(x,w)}{1-\psi(x, w)} -  \frac{\psi_0(x,w)}{\psi(x, w)} \right\}  (1 + \lambda C_0(x)) \ , \\
	\ddot{\mathcal{L}}_{P_0}(\psi, \lambda) &= \frac{1-\psi_0(x,w)}{(1-\psi(x, w))^2} +  \frac{\psi_0(x,w)}{\psi(x, w)^2}  \ .
\end{align*} %
For any given $\psi = (\psi^1, \psi^0) \in {\bf \Psi}$ and $\lambda \in \openr$, the criterion \eqref{eq:borderedH_det_single} would always be satisfied. This concludes that $\psi_{0, \lambda_0}$ is the optimal minimizer of the penalized cross-entropy risk.

\subsection{Equalized cross-entropy risk in the cases and controls}
\label{app:er_cases_controls}

Suppose $O = (W, X, Y) \sim P_0 \in \mathcal{M}$. We are interested in the mappings $\psi$ on the support of $(X, W)$. Let $\Theta_{P_0}(\psi) = (\Theta_{P_0, 1} (\psi), \Theta_{P_0, 0} (\psi))$ where $\Theta_{P_0, y}(\psi) = \E_{P_0}(L(\psi) \mid Y = y, X = 1) - \E_{P_0}(L (\psi) \mid Y = y, X = 0)$ and $L(\psi)$ is the negative log-likelihood loss. $\Theta_{P_0}(\psi)$ encodes a two-dimensional constraint corresponding to the equalized risk in both cases ($Y=1$) and controls ($Y=0$).  

The canonical gradient of the cross-entropy risk is given by Theorem~\ref{thm:gradients}, and the gradient of the constraint $\Theta_{P_0, y}(\psi)$ is $- (2x-1)/P_{0}(x, y) \{ \psi_0(x,w)/\psi(x,w)\}$ (see Supplemental Appendix~\ref{app:er_cases_gradients}). Condition \eqref{appeq:lfmpathconditionb} yields the following:
\begin{align*}
	\frac{\psi_{0, \lambda}(y \mid x, w) - \psi_0(y \mid x, w)}{\psi_{0, \lambda}(y \mid x, w)} -  \ \lambda^{\top}
	\left(
	\begin{array}{cc} 
		C_{0}(x, 1) \ \displaystyle \frac{\psi_0(1 \mid x, w)}{\psi_{0, \lambda}(1 \mid x, w)} (\mathbb{I}(Y = 1)- \psi_{0, \lambda}(1 \mid x, w))  \\ 
		C_{0}(x, 0) \ \displaystyle \frac{\psi_0(0 \mid x, w)}{\psi_{0, \lambda}(0 \mid x, w)}  (\mathbb{I}(Y  = 0) -\psi_{0, \lambda}(0 \mid x, w))
	\end{array}
	\right)
	= 0 \ , 
\end{align*}
where $C_{0}(x, y) = (2x-1)/P_0(x, y)$ and $\lambda^\top = (\lambda_1, \lambda_2)$. A closed-form solution for $\psi_{0, \lambda}$ is obtained as follows 
\begin{align*}
	\psi_{0, \lambda}(y \mid x, w) = \psi_0(y \mid x, w)  \frac{1 +  \lambda_1 C_{0}(x, 1) y + \lambda_2  C_{0}(x, 0) (1-y) }{ 1  + \lambda_1 C_{0}(x, 1) \psi_0(1 \mid x, w) + \lambda_2 C_{0}(x, 0) \psi_0(0 \mid x, w) } \ .
\end{align*}
To determine $\psi_{0, \lambda_0}$, given a solution for $\psi_{0, \lambda},$ it suffices to find $\lambda_0 = (\lambda_{1, 0}, \lambda_{2, 0})$ that satisfies $\Theta_{P_0, 1}(\psi_{0, \lambda_0}) = 0$ and $\Theta_{P_0, 0}(\psi_{0, \lambda_0}) = 0$, via a two-dimensional grid search.

\clearpage
\section{Example: constrained density estimation}
\label{app:moment_rest}

Some statistical models require that estimated probability densities satisfy predefined moment conditions. For instance, in economic policy analysis, we may model income distributions while ensuring that the estimated mean income aligns with a regulatory target. Similarly, in physics and engineering, probability distributions governing energy states often need to satisfy constraints on expected energy levels. 

Let $\psi_0(x) = P_0(X = x)$ be the unconstrained parameter defined as a conditional density function over $X$. Let $\psi_0 =  \argmin_{\psi \in {\bf \Psi}} P_0 L(\psi)$, where $L(\psi)(x) = -\log \psi (x)$ is the negative log-likelihood loss.  Consider $\Theta_0(\psi)$ to be a constraint target parameter of interest $\Theta_0: {\cal M} \mapsto \mathbb{R}$ at $\psi$, where $\cal M$ is the nonparametric model. 

Consider paths $\{\psi_{\delta, h}(x) = \psi(x) \{1 + \delta h(x)\}: \delta \in \openr^\text{restricted}\}$ with direction $\frac{d}{d\delta} \psi_{\delta,h} |_{\delta = 0} = h(x)\psi(x)$ is through $\psi$ at $\delta = 0$. For this path to be within the parameter space $\bf \Psi$, the range of $\delta$ is restricted, and $h(x)$ must be mean zero under $\psi$, $\int h(x) d\psi(x)= 0$. Thus, exploring all possible local directions at $\psi$ within the parameter space $\bf \Psi$ would require that $h(x) \in L^2_0(\psi),$ the space of bounded real-valued functions of $X$ with mean zero and finite second moment under the distribution $\psi(X)$. Thus, $\tangent_{P}(\psi)$ is also defined as $L^2_0(\psi)$. We define the inner product on $L^2_0(\psi)$ as $\langle f, g \rangle = \int f(x)g(x)d\psi(x)$.

\vspace{0.25cm}
\noindent \textbf{Remark.} 
We could also consider paths through $\psi$ at $\delta = 0$ of the form 
\begin{align*}
	\psi_{\delta,h}(x) = \frac{\psi(x)\,\exp\bigl\{\delta\,h(x)\bigr\}}{\displaystyle\int \psi(x)\,\exp\bigl\{\delta\,h(x)\bigr\}\,dx} \ , \quad \delta \in \openr \ , 
\end{align*}
with direction $\frac{d}{d\delta} \psi_{\delta,h}\Big|_{\delta = 0} = h(x) \, \psi(x)$. The canonical gradients do not change. 

For any $h \in \tangent_P(\psi)$, we have
\begin{align*}
	\left. \frac{d}{d \delta} P_0L(\psi_{\delta, h}) \right |_{\delta = 0}  
	&= - P_0 \big(h(X)\big)  \\ 
	&= - \int h(x) \ d\psi_0(x) \\
	&= - \int h(x) \times \frac{\psi_0(x)}{\psi(x)} \ d\psi(x) \\ 
	&= \int \frac{\psi(x) - \psi_0(x)}{\psi(x)} \ h(x) \ d\psi(x) \ .
\end{align*}%
The last equality holds since $\int h(x) d\psi(x) = 0$. Thus, $D_{R, P_0}(\psi) = (\psi - \psi_0) / \psi$. 

\vspace{0.35cm}
\noindent \textbf{Constraint on the $k$-th moment:} Let $\Theta_{P_0}(\psi) = \int x^k d\psi(x)$. 
For any $h \in \tangent_P(\psi)$,  
\begin{align*}
	\frac{d}{d \delta} \Theta_{P_0}(\psi_{\delta, h}) \Big|_{\delta = 0}  
	&=	\frac{d}{d \delta} \int x^k d\psi_{\delta, h}(x) \Big|_{\delta = 0}  \\ 
	&= \int x^k h(x) d\psi(x) \\ 
	&= \int (x^k - \Theta_{P_0}(\psi) ) h(x) d\psi(x) \ . 
\end{align*} 
Thus, $D_{\Theta, P_0}(\psi)(x) = x^k - \Theta_{P_0}(\psi)$.

\vspace{0.45cm}
\noindent The constraint-specific condition \eqref{eq:pathcondition-a} yields: 
\begin{align*}
	\psi_{0, \lambda}(x) = \frac{\psi_0(x)}{1 +  \lambda \left(x^k - \Theta(\psi_{0, \lambda}) \right)   } \ . 
\end{align*}




\vspace{0.35cm}
\noindent \textbf{A general class of constraint on the density:} Let $\Theta_{P_0}(\psi) = \int f(x) d\psi(x),$ for a pre-specified function $f.$ For any $h \in \tangent_P(\psi)$,  
\begin{align*}
	\frac{d}{d \delta} \Theta_{P_0}(\psi_{\delta, h}) \Big|_{\delta = 0}  
	&=	\frac{d}{d \delta} \int f(x) d\psi_{\delta, h}(x) \Big|_{\delta = 0}  \\ 
	&= \int f(x) h(x) d\psi(x) \\ 
	&= \int (f(x) - \Theta_{P_0}(\psi) ) h(x) d\psi(x) \ . 
\end{align*} 
Thus, $D_{\Theta, P_0}(\psi)(x) = f(x) - \Theta_{P_0}(\psi)$. 

\vspace{0.45cm}
\noindent The alternative characterization in \eqref{eq:pathcondition_b} yields:  
\begin{align*}
	\frac{d \psi_{0, \lambda}}{d \lambda} 
	= \mathcal{A}^{-1}_{\lambda, \psi_{0, \lambda}}\Big( D_{\Theta, P_0}(\psi_{0, \lambda}) \psi_{0, \lambda} \Big) \ . 
\end{align*}%
where 
\begin{align*}
	\mathcal{A}_{\lambda, \psi}(h) = \big(1 + \lambda D_{\Theta, P_0}(\psi) \big) h 
	+ \lambda \psi \frac{d}{d \psi} D_{\Theta, P_0}(\psi)(h) \ . 
\end{align*}


\noindent Given $\frac{d }{d \psi}D_{\Theta, P_0}(\psi)(h) = - \int  f(x) h(x) dx$, we have:  
\begin{align*}
	\mathcal{A}_{\lambda, \psi}(h) = \tilde{f}_{\psi, \lambda} h 
	+ \lambda \psi  \alpha(h) = m_\psi  \ , 
\end{align*}
where 
\begin{align*}
	\tilde{f}_{\psi, \lambda} \coloneqq 1 + \lambda \big(f - \Theta_{P_0}(\psi) \big) \ , \quad 
	\alpha(h) \coloneqq \int  f(x) h(x) dx \ , \quad m_\psi \coloneqq D_{\Theta, P_0}(\psi) \psi  \ . 
\end{align*}

We want the inverse of this operator, denoted by $\mathcal{A}^{-1}_{\lambda, \psi}(m_\psi)$. We can write $\mathcal{A}_{\lambda, \psi}(h) = \tilde{f}_{\psi, \lambda} \times \tilde{\mathcal{A}}_{\lambda, \psi}(h) = m_\psi$, where  $\tilde{\mathcal{A}}_{\lambda, \psi}(h) \coloneqq  h + \lambda \psi  \frac{\alpha(h)}{\tilde{f}_{\psi, \lambda}}.$ Therefore,  $\mathcal{A}^{-1}_{\lambda, \psi}(m_\psi) = h = \tilde{\mathcal{A}}^{-1}_{\lambda, \psi}(m_\psi /\tilde{f}_{\psi, \lambda})$. Let  $\tilde{\mathcal{A}}_{\lambda, \psi}(h) = m^*_\psi \coloneqq m_\psi /\tilde{f}_{\psi, \lambda}$. Applying the operator $\alpha(.)$ to both sides, we get: $\alpha(h) + \lambda \alpha(h) \alpha(\psi \tilde{f}^{-1}_{\psi, \lambda}) = \alpha(m^*_\psi).$ So $\alpha(h) = \alpha(m^*_\psi)/(1 + \lambda \alpha(\psi \tilde{f}^{-1}_{\psi, \lambda}) )$. Therefore, \\ 
$h = m^*_\psi - \lambda \psi \tilde{f}^{-1}_{\psi, \lambda} \frac{ \alpha(m^*_\psi)}{1 + \lambda \alpha(\psi \tilde{f}^{-1}_{\psi, \lambda}) }$. Finally,  
\begin{align*}
	\mathcal{A}^{-1}_{\lambda, \psi}(m_\psi) = m_\psi \tilde{f}^{-1}_{\psi, \lambda} - \lambda \psi \tilde{f}^{-1}_{\psi, \lambda} \frac{ \alpha(m_\psi \tilde{f}^{-1}_{\psi, \lambda})}{1 + \lambda \alpha(\psi \tilde{f}^{-1}_{\psi, \lambda}) } \ . 
\end{align*}
Putting everything together, we arrive at:  
\begin{align}
	\frac{d \psi_{0, \lambda}}{d \lambda} 
	&= m_{\psi_{0, \lambda}} \tilde{f}^{-1}_{\psi_{0, \lambda}, \lambda} - \lambda \psi_{0, \lambda} \tilde{f}^{-1}_{\psi_{0, \lambda}, \lambda} \frac{ \alpha(m_{\psi_{0, \lambda}} \tilde{f}^{-1}_{\psi_{0, \lambda}, \lambda})}{1 + \lambda \alpha(\psi_{0, \lambda} \tilde{f}^{-1}_{\psi_{0, \lambda}, \lambda}) } \ ,
\end{align}
where
\begin{align*}
	m_{\psi_{0, \lambda}} &= \big(f - \Theta_{P_0}(\psi_{0, \lambda}) \big) \psi_{0, \lambda} \ , \\
	\tilde{f}_{\psi_{0, \lambda}, \lambda} &= 1 + \lambda \big(f - \Theta_{P_0}(\psi_{0, \lambda}) \big) \ , \\
	\alpha(m_{\psi_{0, \lambda}} \tilde{f}^{-1}_{\psi_{0, \lambda}, \lambda}) &= \int f(x) \frac{ f(x) - \Theta_{P_0}(\psi_{0, \lambda}) }{1 + \lambda \big(f(x) - \Theta_{P_0}(\psi_{0, \lambda}) \big)}  \psi_{0, \lambda} dx \ , \\
	\alpha(\psi_{0, \lambda} \tilde{f}^{-1}_{\psi_{0, \lambda}, \lambda}) &= \int \frac{f(x) }{1 + \lambda \big(f - \Theta_{P_0}(\psi_{0, \lambda}) \big)} \psi_{0, \lambda} dx \ . 
\end{align*}
This representation allows for recursively constructed estimators of $\psi_{0,\lambda}$, while $\lambda_0$ can be determined via a grid search over empirical minimization of the constraint. 

\vspace{1.5cm}
\section{Example: bias correction in misspecified models}
\label{app:misspecified}

Let $P_0 \in \mathcal{M}$, where $\mathcal{M}$ is a nonparametric model. 
Let $\psi_{0,\text{np}}$ be the density $p_0$ of the true $P_0$, i.e.,  the nonparametric minimum defined as $\psi_{0,\text{np}} = \arg \min_{\psi} P_0 L(\psi)$, where $L(\psi) = -\log \psi$. 

Let $\bf \Psi$ denote a parameter space that might not contain the nonparametric minimum $\psi_{0,\text{np}}$. In other words, $\bf \Psi$ is a misspecified parameter space. Let $\psi_0$ denote the projection of the true density $p_0 = \psi_{0,\text{np}}$ onto this misspecified model; i.e.,  $\psi_0 = \arg \min_{\psi \in \bf \Psi} P_0 L(\psi)$.  

Suppose $\mu : \mathcal{M} \to \mathbb{R}$ is a nonparametrically defined target parameter of interest. Let $D^*_\mu(\psi_{0,\text{np}})$ be the canonical gradient of $\mu$ at $P_0$. By design,  $P_0 D^*_\mu(\psi_{0,\text{np}}) = 0.$ Suppose $P_0 D^*_\mu(\psi) = 0$ implies $\mu(\psi) = \mu(p_0)$. Due to the misspecification of $\bf \Psi$, we might not have $P_0 D^*_\mu(\psi_0) = 0$, meaning $\psi_0$ is biased for $\mu(p_0)$. Our objective is to learn a function-valued parameter in the misspecified model $\bf \Psi$, constrained to be unbiased for $\mu(p_0)$. 

Let $\Theta_0(\psi) = P_0 D^*_\mu(\psi).$ We define our constrained functional parameter, $\psi^*_0$, as follows:
\begin{equation}
	\psi_0^* = \argmin_{\psi \in {\bf \Psi}, \ \Theta_0(\psi) = 0} P_0 L(\psi) \ . 
\end{equation}

As a concrete example, consider the following. 

\noindent Let ${\bf \Psi} = \{\psi_\beta : \beta\}$ be a parametric regression model for $\psi_{0,\text{np}} = E_{P_0}(Y\mid X,W)$. 

\noindent Define $\psi_0 = \psi_{\beta_0}$, where: $ \beta_0 = \arg \min_\beta P_0 L(\psi_\beta)$, and $L(\psi_\beta) = (Y - \psi_\beta(X,W))^2$. 

\noindent Let $\mu(P) = E_P [E_P (Y \mid X = 1,W)]$. The canonical gradient of $\mu$ at $P$ is given by:
{\small 
	\begin{equation*}
		D^*_\mu(P) = \frac{X}{\pi(X \mid  W)} \left( Y - E_P(Y \mid X, W) \right) + E(Y \mid X = 1, W) - \mu(P) \ , 
	\end{equation*}
}
where $\pi(x \mid w) = P(x \mid w)$. 

\noindent Define the constrained functional parameter as: $\Theta_0(\beta) = P_0 D^*_\mu(\psi_\beta, \pi_0)$, and the constrained functional parameter as:
\begin{equation}
	\psi_0^* = \psi_{\beta_0^*}, \quad \text{where } \beta_0^* = \argmin_{\beta, \Theta_0(\beta) = 0} P_0 L(\psi_\beta).
\end{equation}

\noindent The solution can be found along the path $\psi_{\beta_0, \lambda}$, where $\lambda_0$ solves: $\Theta_0(\beta_0, \lambda) = 0.$ 

\vspace{0.15cm}
Consider paths through $\psi \in {\bf \Psi}$ of the form $\{\psi_{\delta, h}(x, w) = \psi(x,w) + \delta h(x,w): \delta \in \openr\}$. These paths are indexed by direction $\frac{d}{d\delta} \psi_{\delta,h} |_{\delta = 0} = h(x,w)$, which we can allow to vary over $\tangent_{P}(\psi) = L^2(P_{X,W})$, the space of bounded real-valued functions of $(x,w)$ defined on the support of $(X, W)$ implied by $P$. 

The canonical gradient of the MSE risk is $D_{R, P_0}(\psi_\beta) = 2(\psi_{\beta} - \psi_{\beta_0})$, and the canonical gradient of the constraint $\Theta_{0}(\beta)$ can be derived as follows: 
\begin{align*}
	\frac{d}{d \delta} \Theta_{0}(\psi_{\delta, h}) \Big|_{\delta = 0}  
	&=	\frac{d}{d \delta} \int \Big\{ \frac{x}{\pi_0(1 \mid w)} \left( y - \psi_{\delta, h}(x, w) \right) + \psi_{\delta, h}(1, w) \Big\} dP_0(o) \Big|_{\delta = 0} \\ 
	&\hspace{0.25cm} - \frac{d}{d \delta} \int \psi_{\delta, h}(1, w) dP_0(w) \Big|_{\delta = 0}  \\ 
	&= \int \Big\{ \frac{x}{\pi_0(1 \mid w)} \left( y - h(x, w) \right) + h(1, w) \Big\} dP_0(o) - \int h(1, w) dP_0(w) \\ 
	&= \int  \frac{x}{\pi_0(1 \mid w)} \left( y - h(x, w) \right) dP_0(o) \\ 
	&= \int  \frac{x}{\pi_0(1 \mid w)} \Big\{ \frac{\psi_0(x, w)}{h(x,w)} -  1 \Big\} h(x, w) dP_0(o) \ . 
\end{align*} 

Therefore, $D_{\Theta, P_0}(\psi_\beta)(x, w) =  \frac{a}{\pi_0(1 \mid w)} \Big\{ \frac{\psi_{\beta_0}(x, w)}{h_\beta(x,w)} -  1 \Big\}$, where $h_\beta(x,w) = \frac{d}{d\beta} \psi_{\beta}(x,w)$.  

\vspace{0.35cm}
The constraint-specific condition \ref{eq:pathcondition-a} yields: 
\begin{align*}
	2(\psi_{\beta}(x,w) - \psi_{\beta_0}(x,w)) + \lambda  \frac{x}{\pi_0(1 \mid w)} \Big\{ \frac{\psi_{\beta_0}(x, w)}{h_\beta(x,w)} -  1 \Big\} = 0 \ , 
\end{align*}
which simplifies to: 
\begin{align*}
	\psi_{\beta}(x,w) = \psi_{\beta_0}(x, w) - 0.5 \lambda \frac{x}{\pi_0(1 \mid w)} \Big\{ \frac{\psi_{\beta_0}(x, w)}{h_\beta(x,w)} -  1 \Big\} \ . 
\end{align*}


\clearpage
\section{Additional simulation details}
\label{app:sims_details}
\subsection{Results of primary simulation for NDE}

\begin{figure}
	\centering
	\includegraphics[width=6.5cm, height=5.5cm]{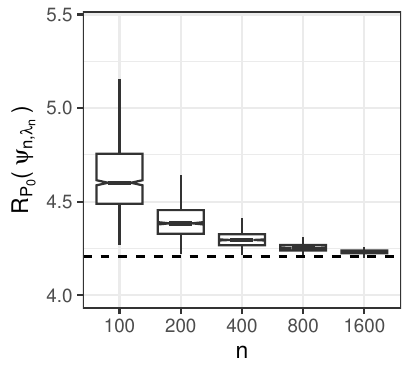}
	\includegraphics[width=6.5cm, height=5.5cm]{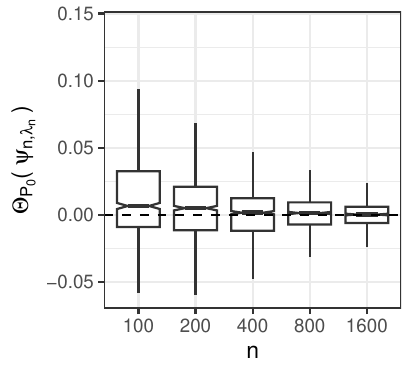}
	\includegraphics[width=6.5cm, height=5.5cm]{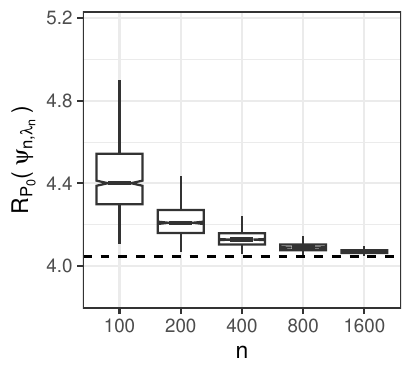}
	\includegraphics[width=6.5cm, height=5.5cm]{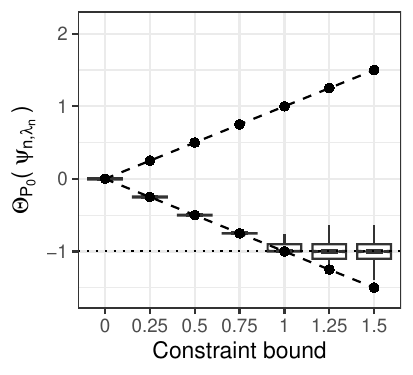}
	\caption{\textbf{Natural direct effect constraint and mean squared error.} \underline{Top left:} Distribution of risk of $\psi_{n,\lambda_n}$ over 1000 realizations for each sample size for the equality constraint $\Theta_{P_0}(\psi) = 0$. The dashed line indicates the optimal risk $R_{P_0}(\psi_0^*)$. \underline{Top right:} Distribution of the true constraint over 1000 realizations for each sample size. The dashed line indicates the equality constraint value of zero. The constraint value under the unconstrained $\psi_0$, $\Theta_{P_0}(\psi_0) = -1$ and is not shown due to the scale of the figure. \underline{Bottom left:} Distribution of risk of $\psi_{n,\lambda_n}$ over 1000 realizations for each sample size for the inequality constraint $|\Theta_{P_0}(\psi)| \le 0.5$. The dashed line indicates the optimal risk $R_{P_0}(\psi_0^*)$ under this constraint. \underline{Bottom right:} Distribution of the true constraint for estimators built using the equality constraint (constraint bound = 0) and inequality constraints with varying bounds at $n = 800$. The dotted line shows the value of the constraint under $\psi_0$, $\Theta_{P_0}(\psi_0)$. The dashed lines shows the positive and negative bounds on the constraint.
	}
	\label{fig:nde_sim}
\end{figure}

Results from the primary simulation for the NDE are shown in Figure \ref{fig:nde_sim}. As with the ATE, we found that the estimators behaved as expected by theory, with estimators achieving optimal constrained risk in large samples and appropriately controlling the constraint based on the user-specified level. We also found dramatically improved performance of the proposed method relative to constrained MLE both in terms of risk, as well as in terms of appropriate control of the constraint.

\begin{figure}
	\centering
	\includegraphics[width=0.45\linewidth]{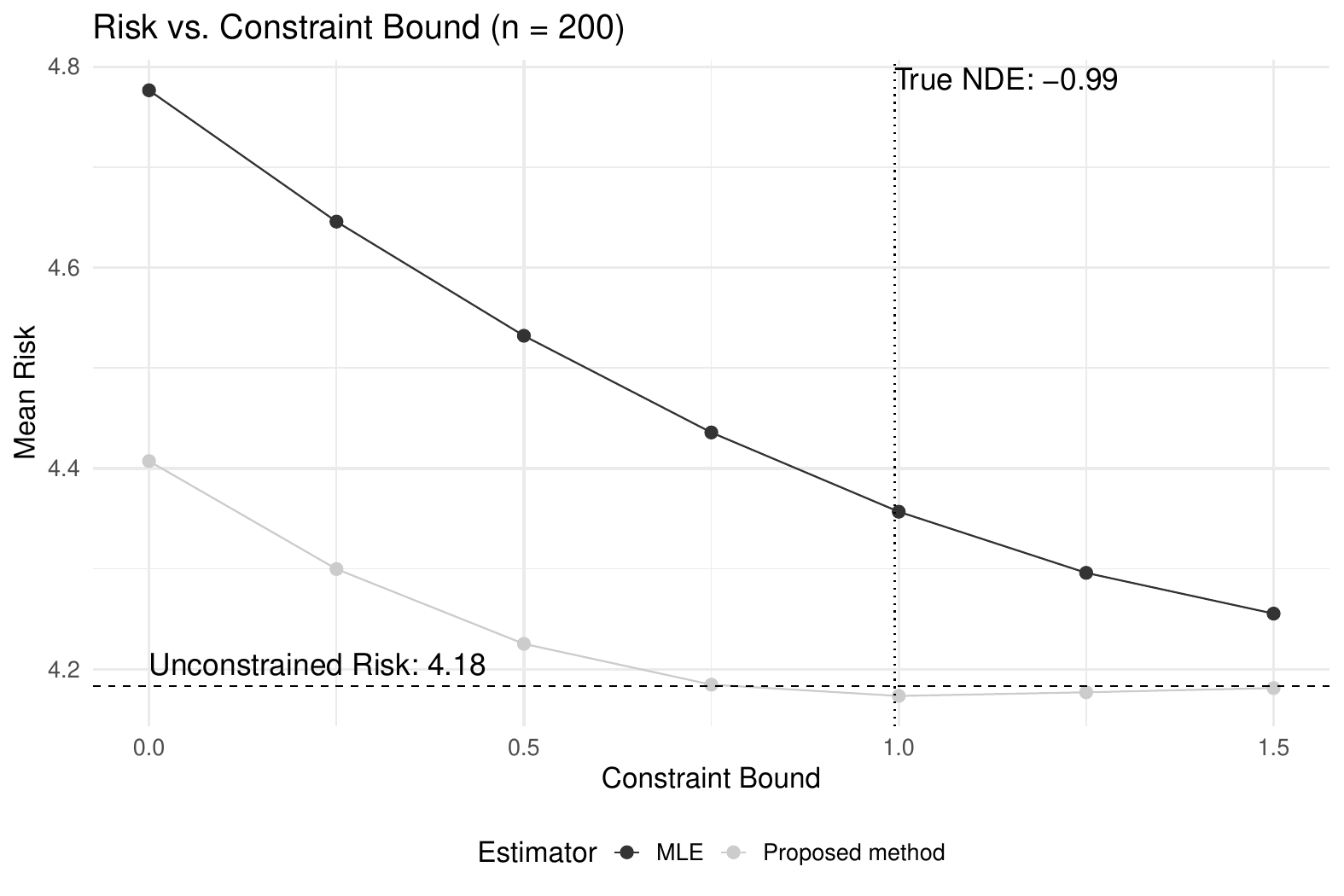}
	\includegraphics[width=0.45\linewidth]{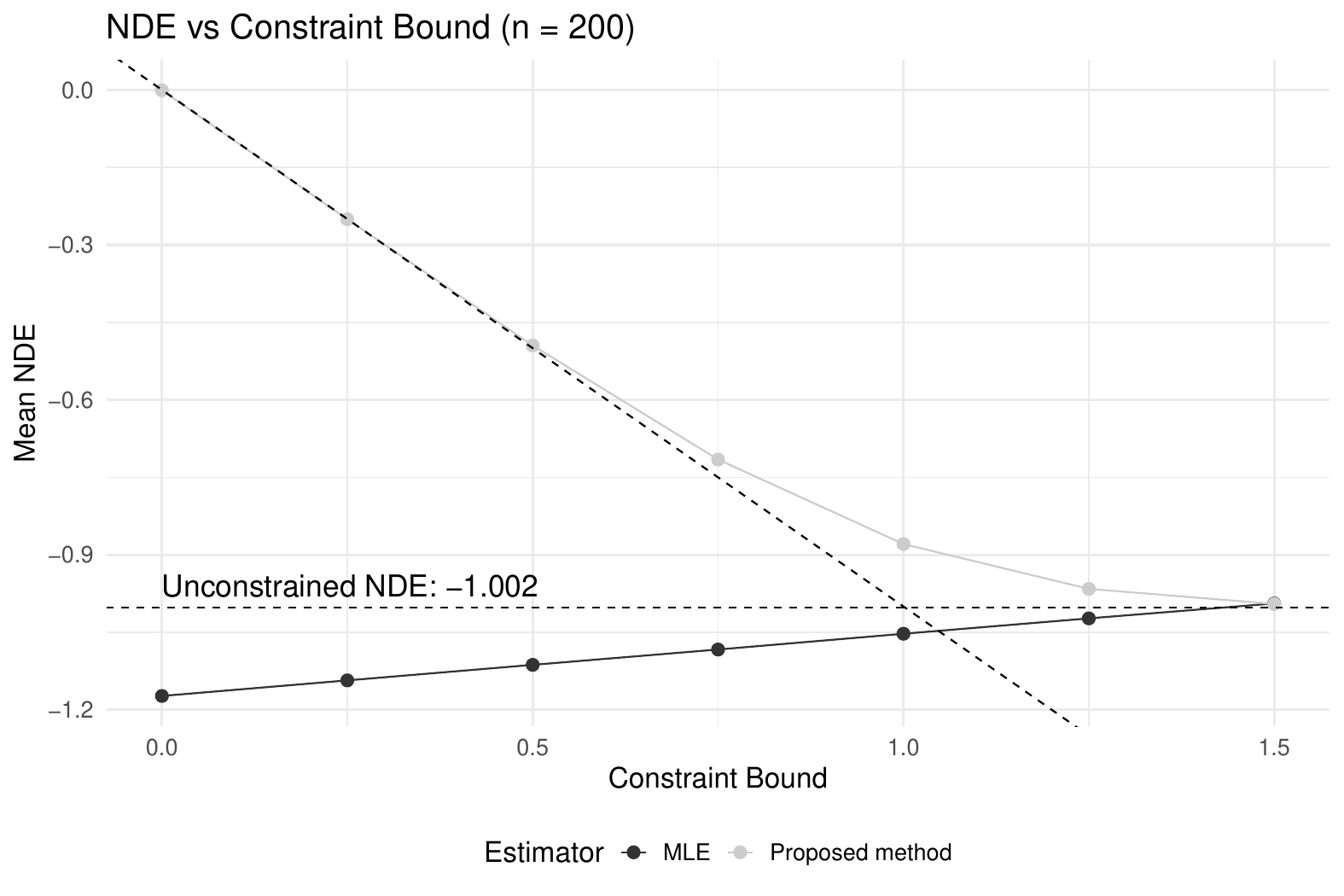}
	\includegraphics[width=0.45\linewidth]{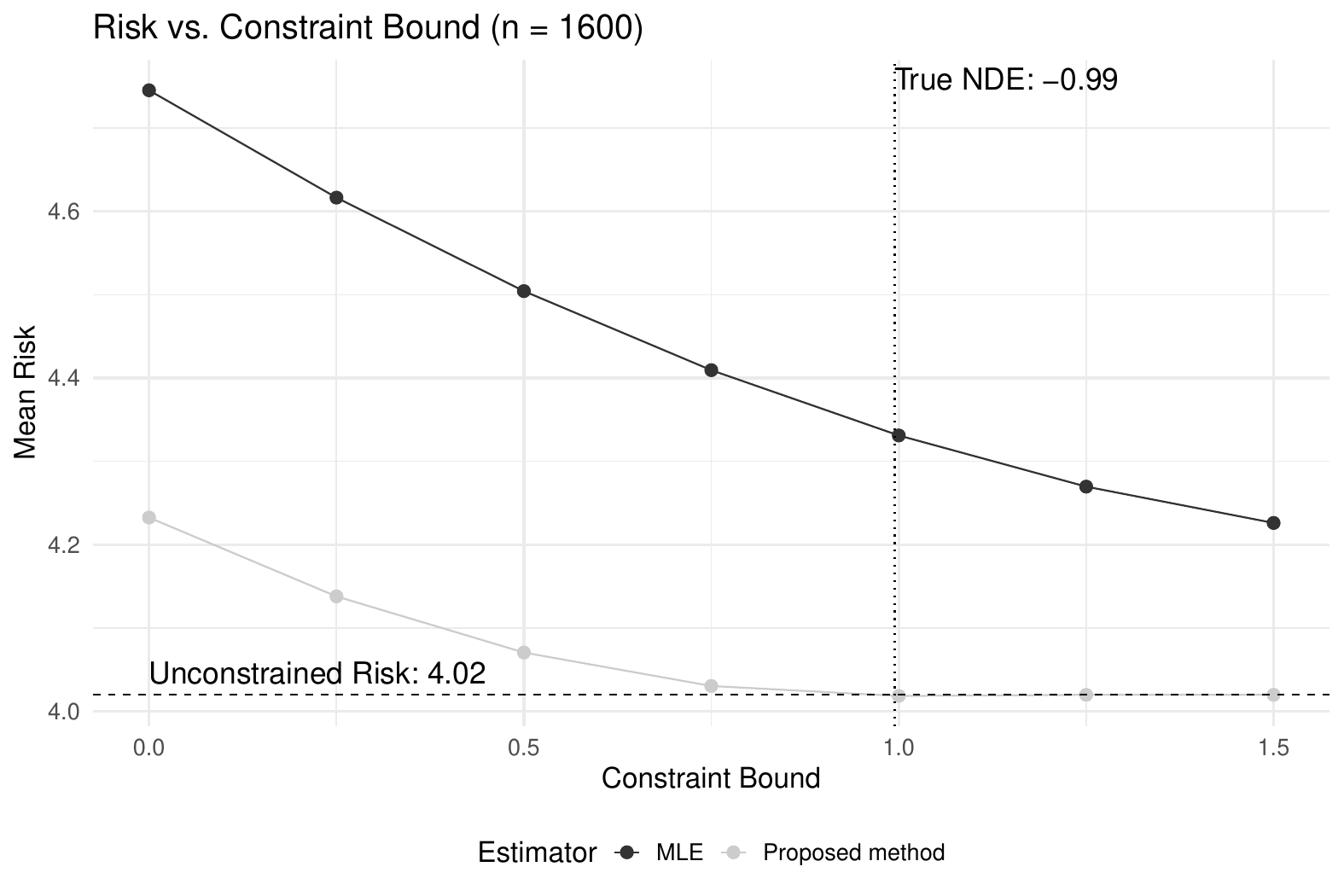}
	\includegraphics[width=0.45\linewidth]{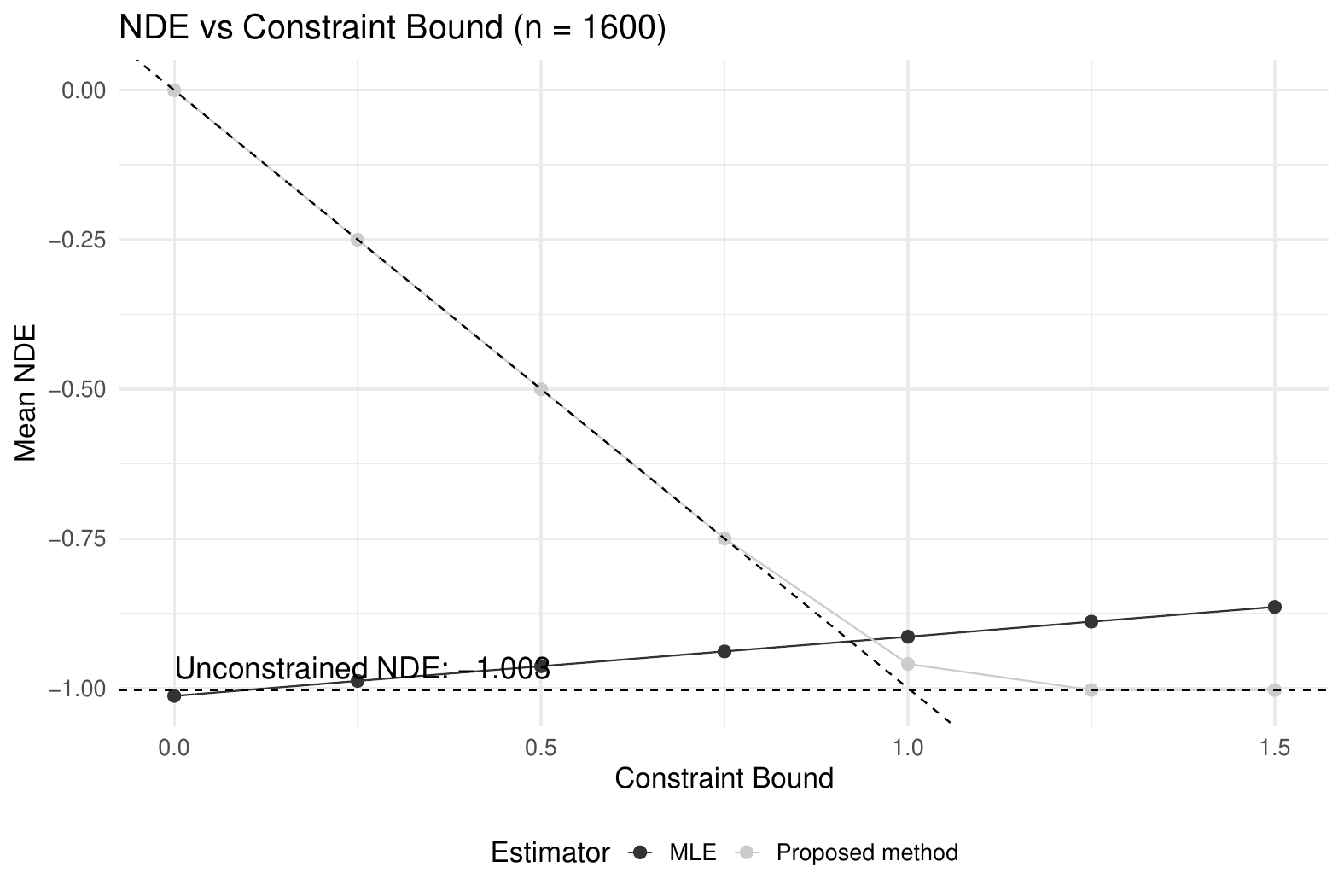}
	\caption{\textbf{Comparison of proposed method vs. constrained MLE.} Average risk (left) and constraint (right) of $\psi_{n,\lambda_n}$ (gray) vs. constrained MLE (black) at two sample sizes. The horizontal dashed line displays the risk of  the unconstrained estimate of $\psi_0$.}
	\label{fig:sim_comp_to_mle}
\end{figure}

\subsection{Results under misspecification of functional parameters}

In the context of causal inference-motivated fair machine learning, efficient estimators of $\Theta_{P_0}(\psi_0)$ often enjoy doubly or multiply robust properties. However, the robustness results for our estimator $\psi_{n, \lambda_n}$ of $\psi_{0, \lambda_0}$ are more complex. In this section, we explore the impact of misspecification of the model for one of the functional parameters required on our procedure. This misspecification was achieved through adding non-linear and interaction terms to the regression formulas used to generate the mediator $M$ and/or the outcome $Y$. However, in the estimation stage, these non-linear terms and interactions were not properly modeled and instead models with only main terms were used. We considered two different simulations, the first extended our base simulation from the main body and focused only on using the plug-in estimator, while exploring the impact of misspecification on both the ATE and NDE. The second considered the impact of estimating the constraint with various different estimators that are available in the causal literature.

\subsubsection{Plug-in estimation of constraint only}

Specifically, we used the same joint distribution of $W$ as described in the main body. In our ``base scenario'', we generated $X$ from a conditional Bernoulli distribution with $\pi_0(1 \mid W) = \mbox{expit}(W_1 - W_3/2 + W_6 / 10)$, $\gamma_0(1 \mid X, W) = \mbox{expit}(-1 - X - W_1 + W_2 / 2 - W_5/5)$, and the conditional mean of the outcome given $(M,X,W)$ was $-2X - M + 2W_1 - W_3 - W_4 + 2W_5$. We then considered three separate ``misspecified scenarios'', wherein we modified the distribution of one of $\pi_0$, $\gamma_0$, and $\E_{P_0}(Y \mid W, M, X)$, respectively, from the ``base scenario''. In the setting where $\pi_0$ was modified to be misspecified, we set $\pi_0(1 \mid W) = \mbox{expit}(W_1 W_2 + W_4^2/50 - W_3/2 + W_6/10)$. When $\gamma_0$ was modified, we set $\gamma_0(1 \mid X, W) = \mbox{expit}(-1 - X - W_1 W_2 + W_4^2/50 + W_2/2 - W_5/2)$. When the conditional mean outcome was modified, we set it to be $-2X - M + 2 W_1 W_2 + W_4^2/2 - W_3 - W_4 + 2W_5$. We only considered the first and third setting for the ATE constraint, because estimation of $\gamma_0$ plays no role in this constraint. In the first and third settings, the true ATE is around -1.86. We considered the NDE constraint in all three settings. In each case the true NDE is -2.

We repeated this simulation study for misspecification of $\psi_0$ (for ATE and NDE constraints), $\pi_0$ (for ATE and NDE constraints), and $\gamma_0$ (for NDE constraint only). We then repeated all simulation studies in a setting where the true value of the constraint under the unconstrained $\psi_0$, $\Theta_{P_0}(\psi_0) = 0$. This allowed us to explore the extent to which misspecification may result in worse control of the constraint than using the unconstrained $\psi_0$. 

Results were similar across different patterns of misspecification and across the two constraints. Thus, we present full results for misspecification of $\psi_0$ and the ATE constraint. Abbreviated results for other patterns of misspecification are shown.

\begin{figure}
	\centering
	\includegraphics[scale=1]{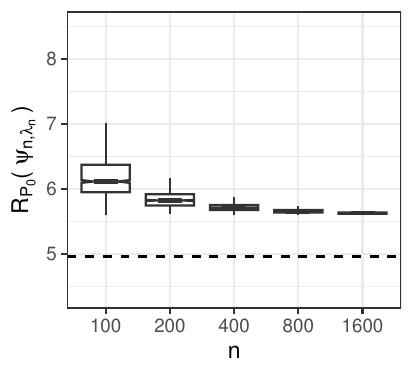}
	\includegraphics[scale=1]{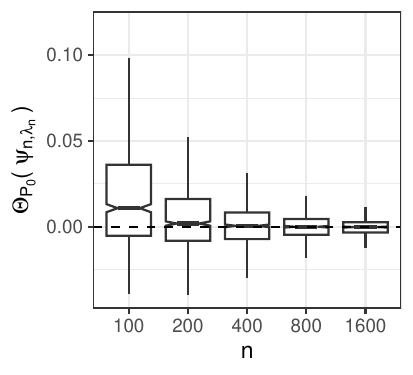}
	\includegraphics[scale=1]{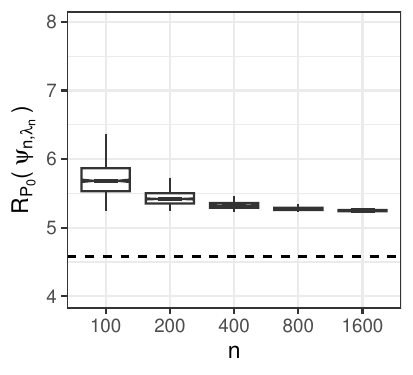}
	\includegraphics[scale=1]{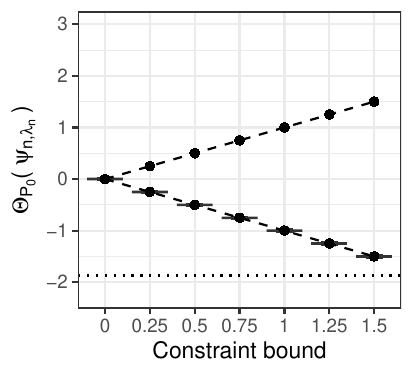}
	\caption{\textbf{Average treatment effect constraint and mean squared error when $\psi_0$ is inconsistently estimated.} \underline{Top left:} Distribution of risk of $\psi_{n,\lambda_n}$ over 1000 realizations for each sample size for the equality constraint $\Theta_{P_0}(\psi) = 0$. The dashed line indicates the optimal risk $R_{P_0}(\psi_0^*)$. \underline{Top right:} Distribution of the true constraint over 1000 realizations for each sample size. The dashed line indicates the equality constraint value of zero. The constraint value under the unconstrained $\psi_0$, $\Theta_{P_0}(\psi_0) = -1.87$ and is not shown due to the scale of the figure. \underline{Bottom left:} Distribution of risk of $\psi_{n,\lambda_n}$ over 1000 realizations for each sample size for the inequality constraint $|\Theta_{P_0}(\psi)| \le 0.5$. The dashed line indicates the optimal risk $R_{P_0}(\psi_0^*)$ under this constraint. \underline{Bottom right:} Distribution of the true constraint for estimators built using the equality constraint (constraint bound = 0) and inequality constraints with varying bounds at $n = 800$. The dotted line shows $\Theta_{P_0}(\psi_0)$. The dashed lines shows the positive and negative bounds on the constraint.}
	\label{fig:ate_misspec_sim}
\end{figure}

For both the ATE, when $\psi_0$ was inconsistently estimated, the constraint $\Theta_{P_0}(\psi_{n,\lambda_n})$ was well-controlled both for equality and inequality constraints (Figure \ref{fig:ate_misspec_sim}, right column). However, due to misspecification of functional parameters, we found sub-optimal risk of $\psi_{n,\lambda_n}$ (left column), confirming the lack of double-robustness of our procedure with respect to risk minimization.

Results for other patterns of misspecification are shown in Figures \ref{fig:ate_misspec_sim}-\ref{fig:ate_sim_misspec_pM}.

\begin{figure}
	\centering
	\includegraphics[scale=1]{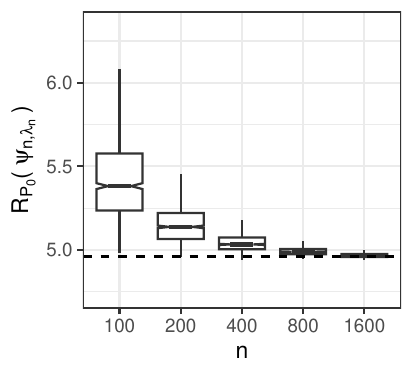}
	\includegraphics[scale=1]{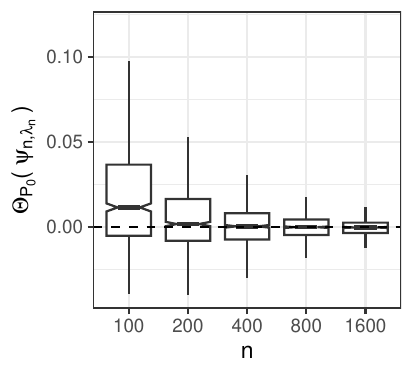}
	\vspace{-0.35cm}
	\caption{\textbf{Average treatment effect constraint in the ``base scenario'' where all functional parameters are consistently estimated.} \underline{Left:} Distribution of risk (mean squared error) of $\psi_{n,\lambda_n}$ over 1000 realizations for each sample size. The dashed line indicates the risk of $\psi_{0,\lambda_0}$. \underline{Right:} Distribution of the true constraint over 1000 realizations for each sample size. The dashed line indicates the desired constraint value of zero. 
	}
	\label{fig:ate_sim_misspec_base_case}
\end{figure}

\begin{figure}
	\centering
	\includegraphics[scale=1]{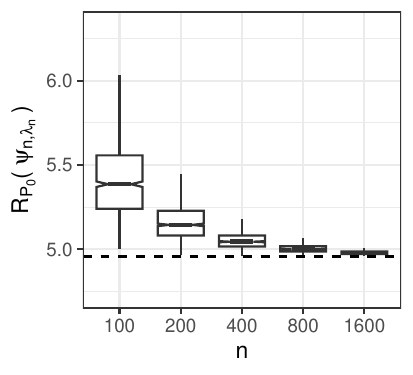}
	\includegraphics[scale=1]{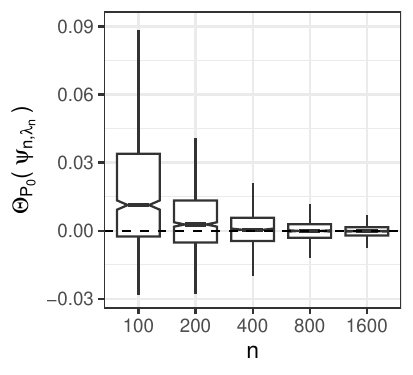}
	\vspace{-0.35cm}
	\caption{\textbf{Average treatment effect constraint in the scenario where $P_0(X \mid W)$ is inconsistently estimated.} Left: Distribution of risk (mean squared error) of $\psi_{n,\lambda_n}$ over 1000 realizations for each sample size. The dashed line indicates the risk of $\psi_{0,\lambda_0}$. Right: Distribution of the true constraint over 1000 realizations for each sample size. The dashed line indicates the desired constraint value of zero. 
	}
	\label{fig:ate_sim_misspec_ps}
\end{figure}

\begin{figure}
	\centering
	\includegraphics[scale=1]{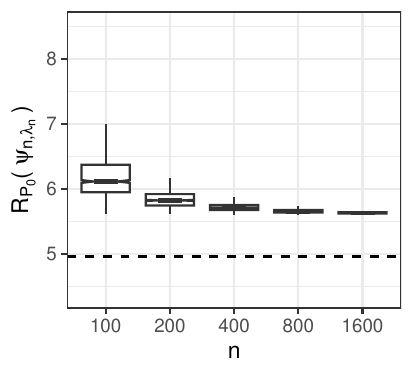}
	\includegraphics[scale=1]{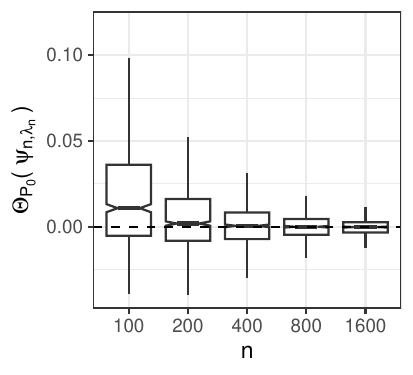}
	\vspace{-0.35cm}
	\caption{\textbf{Average treatment effect constraint in the scenario where the conditional mean outcome is inconsistently estimated.} \underline{Left:} Distribution of risk (mean squared error) of $\psi_{n,\lambda_n}$ over 1000 realizations for each sample size. The dashed line indicates the risk of $\psi_{0,\lambda_0}$. \underline{Right:} Distribution of the true constraint over 1000 realizations for each sample size. The dashed line indicates the desired constraint value of zero. 
	}
	\label{fig:ate_sim_misspec_or}
\end{figure}

\begin{figure}
	\centering
	\includegraphics[scale=1]{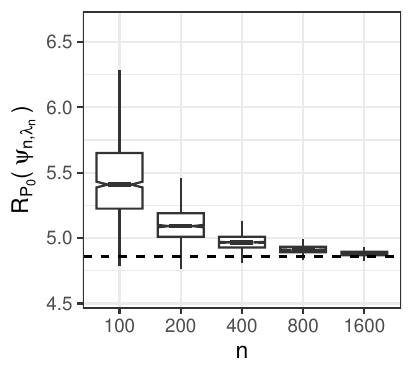}
	\includegraphics[scale=1]{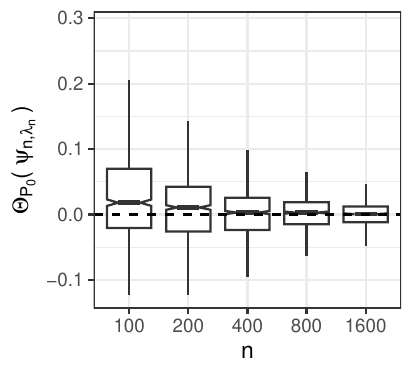}
	\vspace{-0.35cm}
	\caption{\textbf{Natural direct effect constraint in the ``base scenario'' where all functional parameters are consistently estimated.} \underline{Left:} Distribution of risk (mean squared error) of $\psi_{n,\lambda_n}$ over 1000 realizations for each sample size. The dashed line indicates the risk of $\psi_{0,\lambda_0}$. \underline{Right:} Distribution of the true constraint over 1000 realizations for each sample size. The dashed line indicates the desired constraint value of zero. 
	}
	\label{fig:nde_sim_misspec_base_case}
\end{figure}

\begin{figure}
	\centering
	\includegraphics[scale=1]{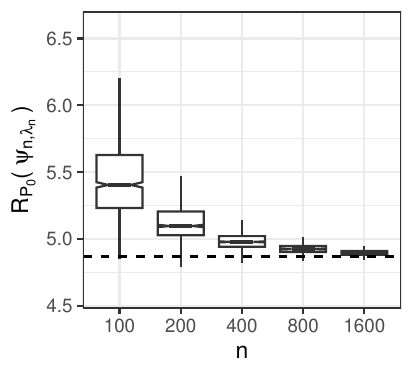}
	\includegraphics[scale=1]{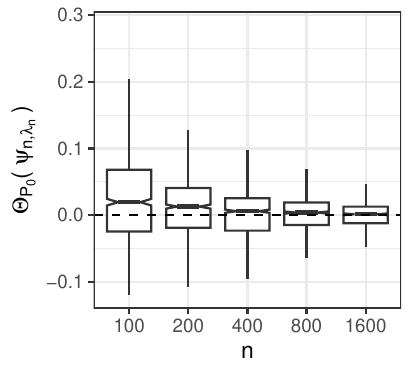}
	\vspace{-0.35cm}
	\caption{\textbf{Natural direct effect constraint in the scenario where $P_0(X \mid W)$ is inconsistently estimated.} \underline{Left:} Distribution of risk (mean squared error) of $\psi_{n,\lambda_n}$ over 1000 realizations for each sample size. The dashed line indicates the risk of $\psi_{0,\lambda_0}$. \underline{Right:} Distribution of the true constraint over 1000 realizations for each sample size. The dashed line indicates the desired constraint value of zero. 
	}
	\label{fig:nde_sim_misspec_ps}
\end{figure}

\begin{figure}
	\centering
	\includegraphics[scale=1]{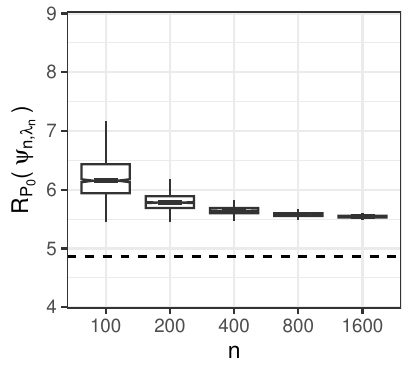}
	\includegraphics[scale=1]{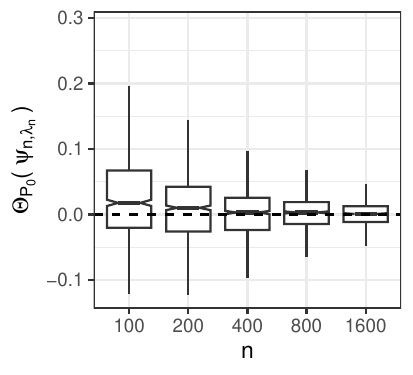}
	\vspace{-0.35cm}
	\caption{\textbf{Natural direct effect constraint in the scenario where the conditional mean outcome is inconsistently estimated.} \underline{Left:} Distribution of risk (mean squared error) of $\psi_{n,\lambda_n}$ over 1000 realizations for each sample size. The dashed line indicates the risk of $\psi_{0,\lambda_0}$. \underline{Right:} Distribution of the true constraint over 1000 realizations for each sample size. The dashed line indicates the desired constraint value of zero. 
	}
	\label{fig:nde_sim_misspec_or}
\end{figure}

\begin{figure}
	\centering
	\includegraphics[scale=1]{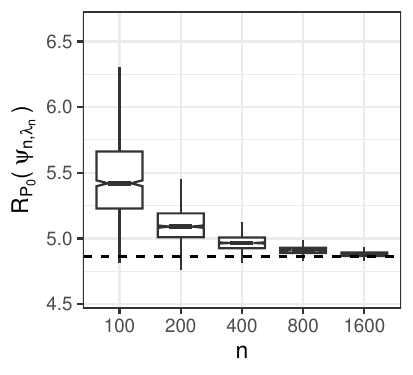}
	\includegraphics[scale=1]{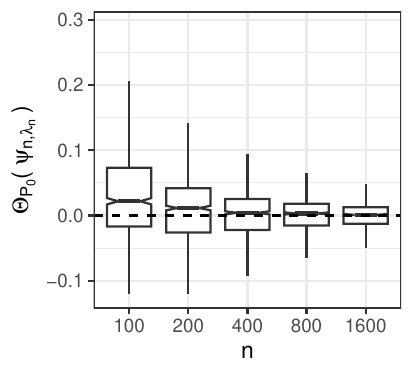}
	\vspace{-0.35cm}
	\caption{\textbf{Natural direct effect constraint in the scenario where $P_0(M \mid X, W)$ is inconsistently estimated.} \underline{Left:} Distribution of risk (mean squared error) of $\psi_{n,\lambda_n}$ over 1000 realizations for each sample size. The dashed line indicates the risk of $\psi_{0,\lambda_0}$. \underline{Right:} Distribution of the true constraint over 1000 realizations for each sample size. The dashed line indicates the desired constraint value of zero. 
	}
	\label{fig:ate_sim_misspec_pM}
\end{figure}

We repeated the above simulation but changed the ``base scenario'' so that both the ATE and NDE constraints evaluated to zero. This was achieved by setting the conditional mean of the outcome to $2W_1 - W_3 - W_4 + 2W_5$. In the scenario where the conditional mean of the outcome is misspeficied, we set the true value of the conditional mean to $2W_1 W_2 + W_4^2/2 - W_3 - W_4 + W_5$. Results are shown in Figures \ref{fig:ate_sim_misspec_base_case} - \ref{fig:ate_sim_misspec_pM}.

\begin{figure}
	\centering
	\includegraphics[scale=1]{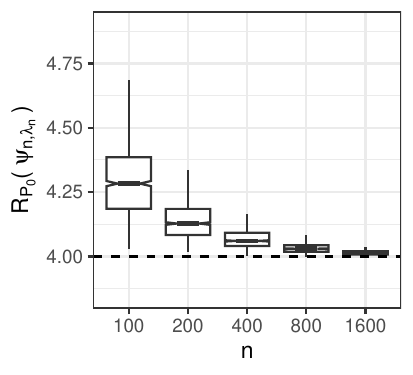}
	\includegraphics[scale=1]{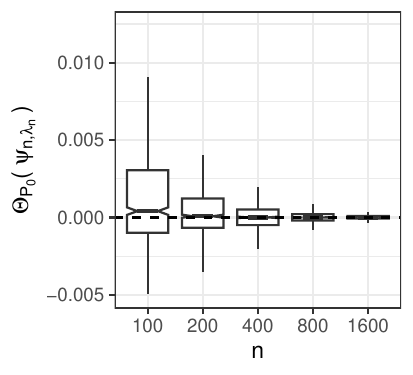}
	\vspace{-0.35cm}
	\caption{\textbf{Average treatment effect constraint in the ``base scenario'' with zero-valued constraint and where all functional parameters are consistently estimated.} \underline{Left:} Distribution of risk (mean squared error) of $\psi_{n,\lambda_n}$ over 1000 realizations for each sample size. The dashed line indicates the risk of $\psi_{0,\lambda_0}$. \underline{Right:} Distribution of the true constraint over 1000 realizations for each sample size. The dashed line indicates the desired constraint value of zero. 
	}
	\label{fig:ate_sim_misspec_base_case}
\end{figure}

\begin{figure}
	\centering
	\includegraphics[scale=1]{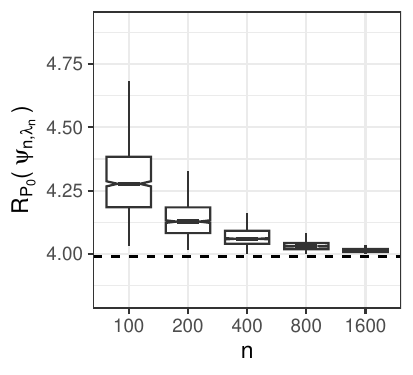}
	\includegraphics[scale=1]{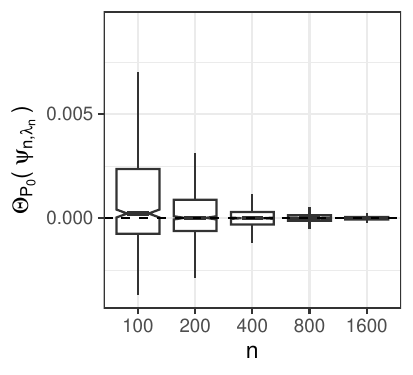}
	\vspace{-0.35cm}
	\caption{\textbf{Average treatment effect constraint in the scenario with zero-valued constraint and where $P_0(X \mid W)$ is inconsistently estimated.} \underline{Left:} Distribution of risk (mean squared error) of $\psi_{n,\lambda_n}$ over 1000 realizations for each sample size. The dashed line indicates the risk of $\psi_{0,\lambda_0}$. \underline{Right:} Distribution of the true constraint over 1000 realizations for each sample size. The dashed line indicates the desired constraint value of zero. 
	}
	\label{fig:ate_sim_misspec_ps}
\end{figure}

\begin{figure}
	\centering
	\includegraphics[scale=1]{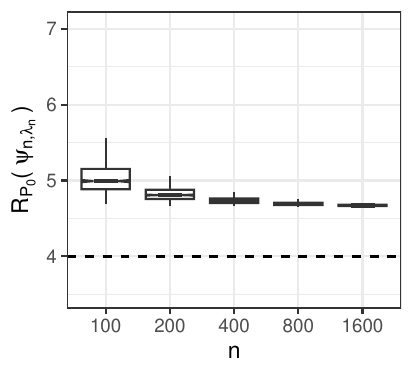}
	\includegraphics[scale=1]{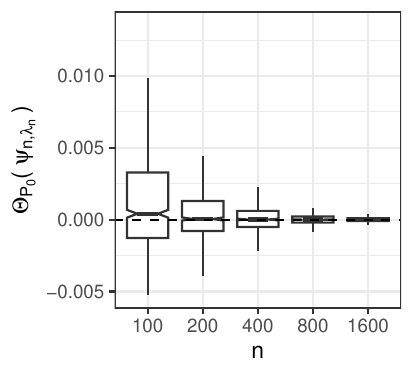}
	\vspace{-0.35cm}
	\caption{\textbf{Average treatment effect constraint in the scenario with zero-valued constraint and where the conditional mean outcome is inconsistently estimated.} \underline{Left:} Distribution of risk (mean squared error) of $\psi_{n,\lambda_n}$ over 1000 realizations for each sample size. The dashed line indicates the risk of $\psi_{0,\lambda_0}$. \underline{Right:} Distribution of the true constraint over 1000 realizations for each sample size. The dashed line indicates the desired constraint value of zero. 
	}
	\label{fig:ate_sim_misspec_or}
\end{figure}

\begin{figure}
	\centering
	\includegraphics[scale=1]{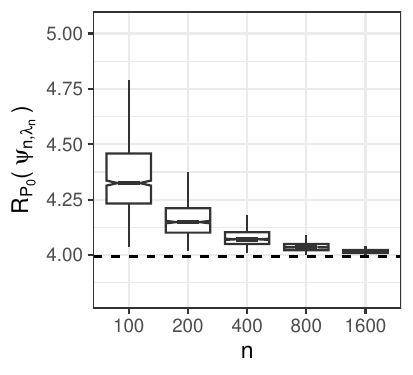}
	\includegraphics[scale=1]{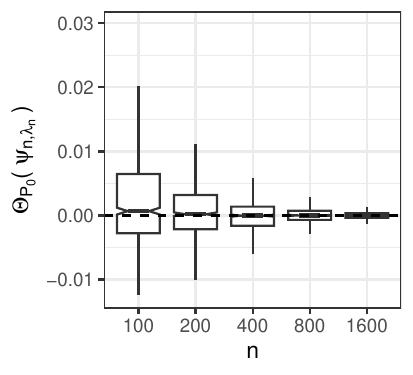}
	\vspace{-0.35cm}
	\caption{\textbf{Natural direct effect constraint in the ``base scenario'' with zero-valued constraint and where all functional parameters are consistently estimated.} \underline{Left:} Distribution of risk (mean squared error) of $\psi_{n,\lambda_n}$ over 1000 realizations for each sample size. The dashed line indicates the risk of $\psi_{0,\lambda_0}$. \underline{Right:} Distribution of the true constraint over 1000 realizations for each sample size. The dashed line indicates the desired constraint value of zero. 
	}
	\label{fig:nde_sim_misspec_base_case}
\end{figure}

\begin{figure}
	\centering
	\includegraphics[scale=1]{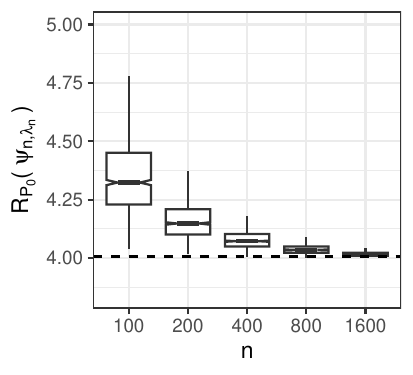}
	\includegraphics[scale=1]{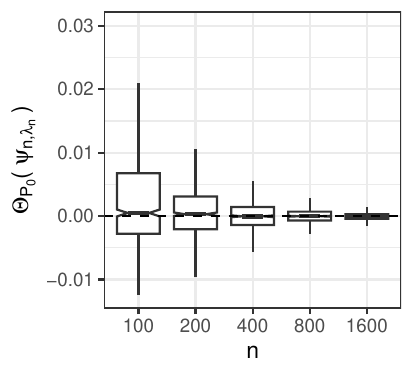}
	\vspace{-0.35cm}
	\caption{\textbf{Natural direct effect constraint in the scenario with zero-valued constraint and where $P_0(X \mid W)$ is inconsistently estimated.} \underline{Left:} Distribution of risk (mean squared error) of $\psi_{n,\lambda_n}$ over 1000 realizations for each sample size. The dashed line indicates the risk of $\psi_{0,\lambda_0}$. \underline{Right:} Distribution of the true constraint over 1000 realizations for each sample size. The dashed line indicates the desired constraint value of zero. 
	}
	\label{fig:nde_sim_misspec_ps}
\end{figure}

\begin{figure}
	\centering
	\includegraphics[scale=1]{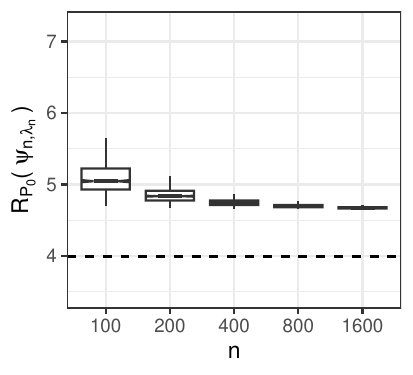}
	\includegraphics[scale=1]{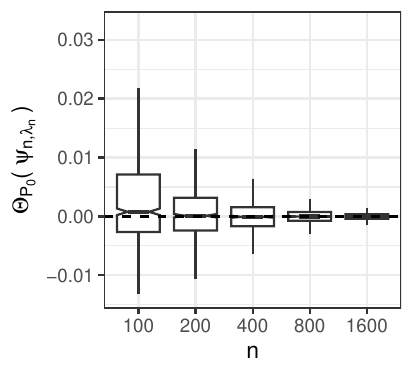}
	\vspace{-0.35cm}
	\caption{\textbf{Natural direct effect constraint in the scenario with zero-valued constraint and where the conditional mean outcome is inconsistently estimated.} \underline{Left:} Distribution of risk (mean squared error) of $\psi_{n,\lambda_n}$ over 1000 realizations for each sample size. The dashed line indicates the risk of $\psi_{0,\lambda_0}$. \underline{Right:} Distribution of the true constraint over 1000 realizations for each sample size. The dashed line indicates the desired constraint value of zero. 
	}
	\label{fig:nde_sim_misspec_or}
\end{figure}

\begin{figure}
	\centering
	\includegraphics[scale=1]{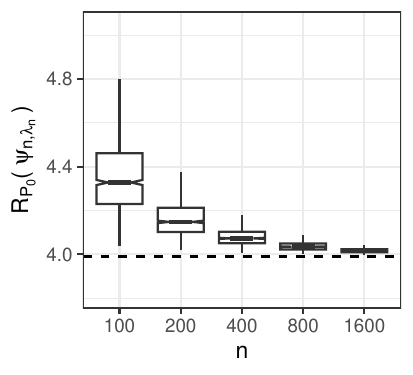}
	\includegraphics[scale=1]{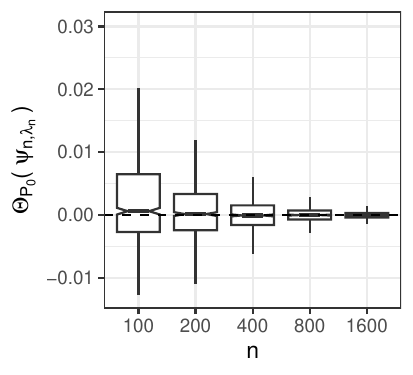}
	\vspace{-0.35cm}
	\caption{\textbf{Natural direct effect constraint in the scenario with zero-valued constraint and where $P_0(M \mid X, W)$ is inconsistently estimated.} \underline{Left:} Distribution of risk (mean squared error) of $\psi_{n,\lambda_n}$ over 1000 realizations for each sample size. The dashed line indicates the risk of $\psi_{0,\lambda_0}$. \underline{Right:} Distribution of the true constraint over 1000 realizations for each sample size. The dashed line indicates the desired constraint value of zero. 
	}
	\label{fig:ate_sim_misspec_pM}
\end{figure}

\subsubsection{Using robust estimators of the constraint}

In this simulation, we wanted to characterize the impact of the choice of estimator of the constraint on the performance of $\psi_{n, \lambda_n}$. We considered the NDE, where several estimators of the NDE are available in the causal inference literature, including plug-in estimation (PLUG-IN), inverse probability of treatment weighted estimation (IPTW), and augmented inverse probability of treatment weight estimation (AIPTW). The first IPTW estimator we considered is defined as \[
\Theta_{n,\text{IPTW}} = \frac{1}{n} \sum_{i=1}^{n} \frac{2X_i - 1}{\pi_n(X_i \mid W_i)} \frac{\gamma_n(0 \mid X_i, W_i)}{\gamma_n(M_i \mid X_i, W_i)} Y_i \ ,
\]
but we also considered an estimator defined as \[
\Theta_{n,\text{IPTW-ALT}} = \frac{1}{n} \sum_{i=1}^{n} \frac{2X_i - 1}{\pi_n(X_i \mid W_i)} \frac{\gamma_n(0 \mid X_i, W_i)}{\gamma_n(M_i \mid X_i, W_i)} \psi_n(X_i, M_i, W_i) \ .
\]
The AIPTW estimator can be constructed by defining \[
\bar{\psi}_{n,x}(W_i) = \sum_{m=0}^1 \psi_n(x, m, W_i) \gamma_n(m \mid x, W_i) \ , 
\]
and subsequently computing \begin{align*}
	\phi_{n,x}(O_i) &= \frac{I(X_i = x)}{\pi_n(X_i \mid W_i)} \frac{\gamma_n(0 \mid X_i, W_i)}{\gamma_n(M_i \mid X_i, W_i)} \{ Y_i - \psi_n(X_i, M_i, W_i) \} \\
	&\hspace{2em} \frac{I(X_i = 0)}{\pi_n(0 \mid W_i)} \{ \psi_n(x, M_i, W_i) -  \bar{\psi}_{n,x}(W_i) \} \ ,
\end{align*}
for each $i = 1, \dots, n$. The AIPTW estimator is $\Theta_{n, \text{AIPTW}} = n^{-1} \sum_{i=1}^n \{ \phi_{n,1}(O_i) - \phi_{n,0}(O_i) \}$. This estimator enjoys a multiply robust property such that it is consistent for $\Theta_{P_0}(\psi_0)$ if two out of the three of the conditional distribution of $X$, conditional distribution of $M$, and the conditional mean of $Y$ are consistently estimated.

For this simulation, we generated two covariates $W$. $W_1$ was drawn from a Bernoulli(1/2) distribution; $W_2$ was drawn from a Uniform(-2, 2) distribution. We set $\pi_0(1 \mid W) = \mbox{expit}(-W_1 + 2 W_1 W_2)$ and set $\gamma_0(1 \mid X, W) = \mbox{expit}(-1/3 + 1/2 X - W_1 + 2 W_1 W_2)$. Finally, we simulated $Y$ by drawing a random error term from a Normal(0, $2^2$) distribution with conditional mean $\psi_0(X, M, W) = X + M + W_1 W_2$. The true NDE under this data generating process is 1.

We note that each of the nuisance parameters features a cross-product term between $W_1$ and $W_2$. We considered different scenarios where either (i) all nuisances were consistently estimated using maximum likelihood based on correctly specific parametric regression models or (ii) one of the nuisances was inconsistently estimated due to the use of parametric regression formula that omitted the cross-product between $W_1$ and $W_2$. We considered two sample sizes $n = 200$ and $n = 1600$ and estimators that fixed the value of the constraint to 0. For each sample size, we simulated 1000 data sets and report on the value of the risk $R_{P_0}(\psi_{n, \lambda_n})$ and of the constraint $\Theta_{P_0}(\psi_{n, \lambda_n})$ for estimates based on the four different estimation strategies for $\Theta$. 

\begin{figure}[ht]
	\centering
	
	\begin{subfigure}[b]{0.45\textwidth}
		\centering
		\includegraphics[width=\textwidth]{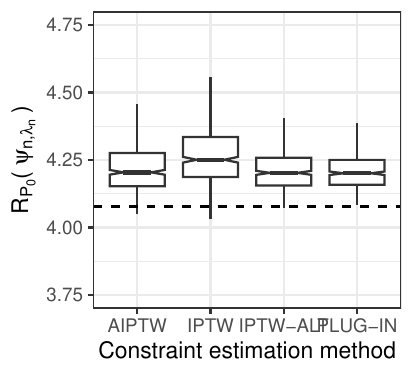}
		\caption{Risk when all nuisances consistently estimated ($n = 200$)}
	\end{subfigure}
	\hfill
	\begin{subfigure}[b]{0.45\textwidth}
		\centering
		\includegraphics[width=\textwidth]{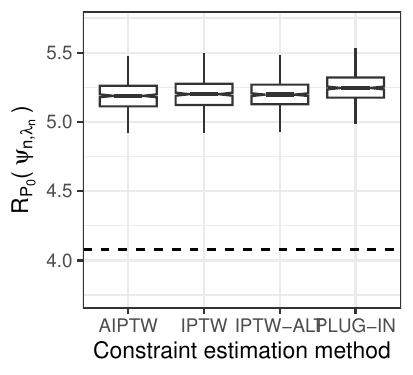}
		\caption{Risk when outcome regression $\psi_0$ inconsistently estimated ($n = 200$)}
	\end{subfigure}
	
	\vspace{0.5cm} 
	
	\begin{subfigure}[b]{0.45\textwidth}
		\centering
		\includegraphics[width=\textwidth]{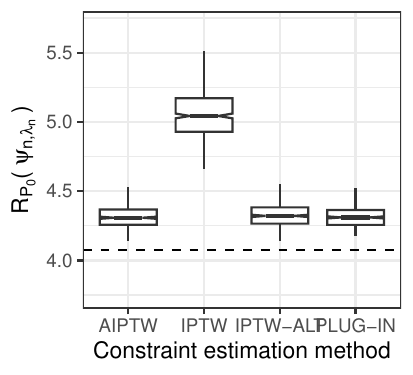}
		\caption{Risk when propensity score $\pi_0$ inconsistently estimated ($n = 200$)}
	\end{subfigure}
	\hfill
	\begin{subfigure}[b]{0.45\textwidth}
		\centering
		\includegraphics[width=\textwidth]{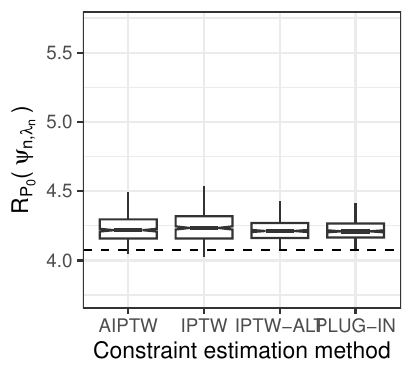}
		\caption{Risk when mediator distribution $\gamma_0$ inconsistently estimated ($n = 200$)}
	\end{subfigure}
	
	\caption{Comparison of risk of constraint estimators constructed using robust and non-robust estimators of $\Theta_{P_0}$. The theoretically optimal value of the risk of constained estimators is shown by the horizontal dashed line.}
	\label{fig:risk_misspec_sim_n200}
\end{figure}

\begin{figure}[ht]
	\centering
	
	\begin{subfigure}[b]{0.45\textwidth}
		\centering
		\includegraphics[width=\textwidth]{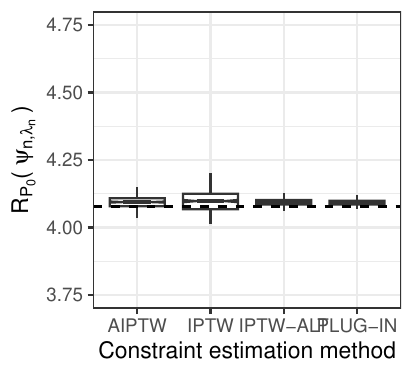}
		\caption{Risk when all nuisances consistently estimated ($n = 1600$)}
	\end{subfigure}
	\hfill
	\begin{subfigure}[b]{0.45\textwidth}
		\centering
		\includegraphics[width=\textwidth]{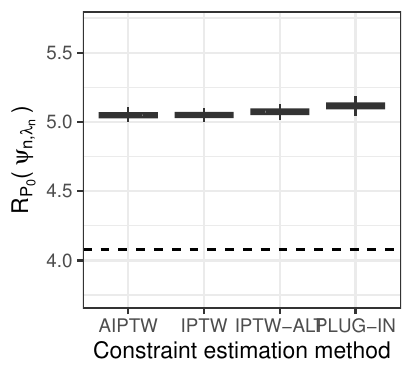}
		\caption{Risk when outcome regression $\psi_0$ inconsistently estimated ($n = 1600$)}
	\end{subfigure}
	
	\vspace{0.5cm} 
	
	\begin{subfigure}[b]{0.45\textwidth}
		\centering
		\includegraphics[width=\textwidth]{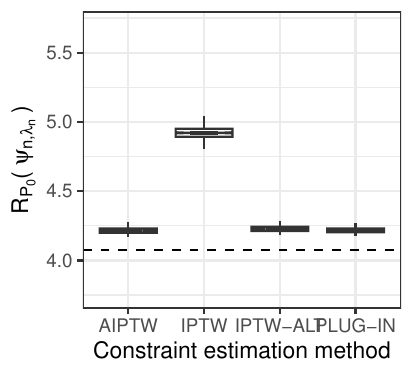}
		\caption{Risk when propensity score $\pi_0$ inconsistently estimated ($n = 1600$)}
	\end{subfigure}
	\hfill
	\begin{subfigure}[b]{0.45\textwidth}
		\centering
		\includegraphics[width=\textwidth]{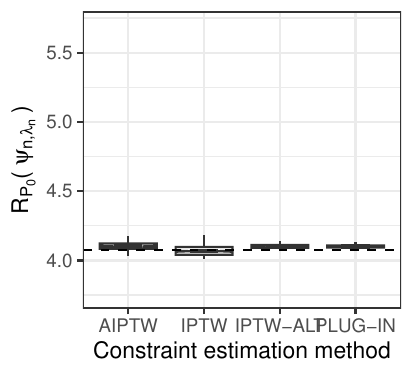}
		\caption{Risk when mediator distribution $\gamma_0$ inconsistently estimated ($n = 1600$)}
	\end{subfigure}
	
	\caption{Comparison of risk of constraint estimators constructed using robust and non-robust estimators of $\Theta_{P_0}$. The theoretically optimal value of the risk of constained estimators is shown by the horizontal dashed line.}
	\label{fig:risk_misspec_sim_n1600}
\end{figure}

Unsurprisingly, the estimator's predictive performance was most heavily impacted by inconsistent estimation of $\psi_0$ (Figures \ref{fig:risk_misspec_sim_n200} and \ref{fig:risk_misspec_sim_n1600}), though, as predicted by our theory, achieving the optimal constrained risk in this problem requires consistent estimation of all nuisance parameters and larger sample size (top left panel). We also found that when the propensity score was inconsistently estimated, the performance of $\psi_{n, \lambda_n}$ when constructed using an IPTW estimate of $\Theta_{P_0}$ was similarly poor, in spite of consistent estimation of $\psi_0$ in this setting (bottom left panel).

\begin{figure}[ht]
	\centering
	
	\begin{subfigure}[b]{0.45\textwidth}
		\centering
		\includegraphics[width=\textwidth]{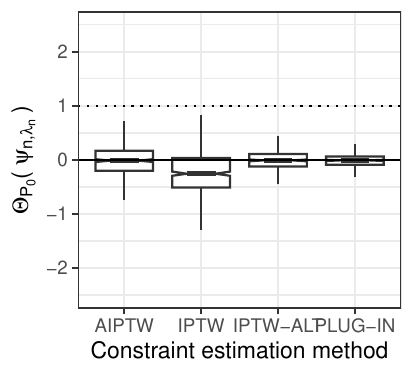}
		\caption{Constraint value when all nuisances consistently estimated ($n = 200$)}
	\end{subfigure}
	\hfill
	\begin{subfigure}[b]{0.45\textwidth}
		\centering
		\includegraphics[width=\textwidth]{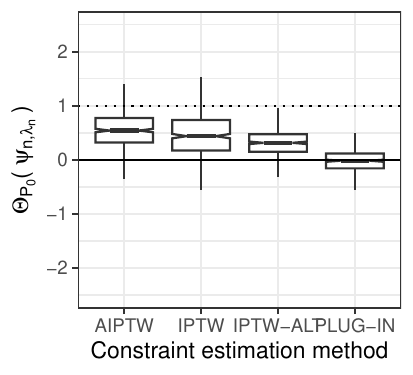}
		\caption{Constraint value when outcome regression $\psi_0$ inconsistently estimated ($n = 200$)}
	\end{subfigure}
	
	\vspace{0.5cm} 
	
	\begin{subfigure}[b]{0.45\textwidth}
		\centering
		\includegraphics[width=\textwidth]{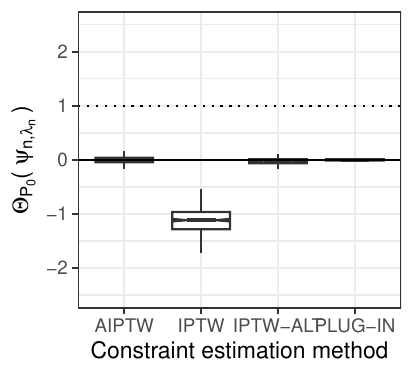}
		\caption{Constraint value when propensity score $\pi_0$ inconsistently estimated ($n = 200$)}
	\end{subfigure}
	\hfill
	\begin{subfigure}[b]{0.45\textwidth}
		\centering
		\includegraphics[width=\textwidth]{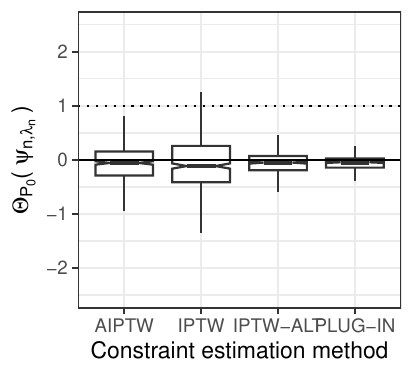}
		\caption{Constraint value when mediator distribution $\gamma_0$ inconsistently estimated ($n = 200$)}
	\end{subfigure}
	
	\caption{Comparison of the ability of estimators to control constraint when constructed using robust and non-robust estimators of $\Theta_{P_0}$. The true value $\Theta_{P_0}(\psi_0)$ is shown by the dotted line. The target constraint value is 0, as indicated by the solid horizontal line.}
	\label{fig:constraint_misspec_sim_n200}
\end{figure}

\begin{figure}[ht]
	\centering
	
	\begin{subfigure}[b]{0.45\textwidth}
		\centering
		\includegraphics[width=\textwidth]{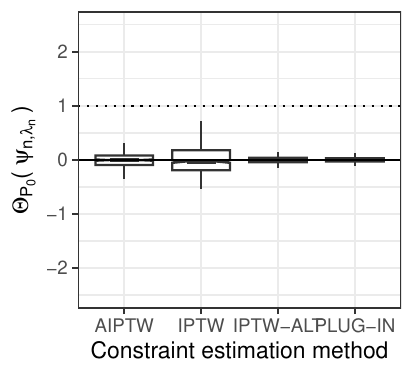}
		\caption{Constraint value when all nuisances consistently estimated ($n = 1600$)}
	\end{subfigure}
	\hfill
	\begin{subfigure}[b]{0.45\textwidth}
		\centering
		\includegraphics[width=\textwidth]{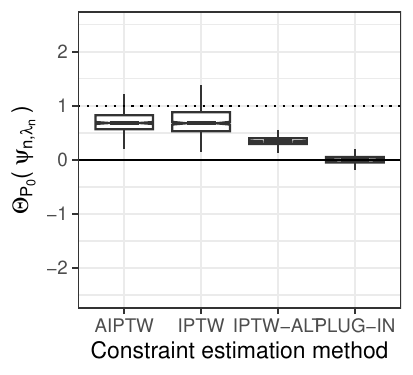}
		\caption{Constraint value when outcome regression $\psi_0$ inconsistently estimated ($n = 1600$)}
	\end{subfigure}
	
	\vspace{0.5cm} 
	
	\begin{subfigure}[b]{0.45\textwidth}
		\centering
		\includegraphics[width=\textwidth]{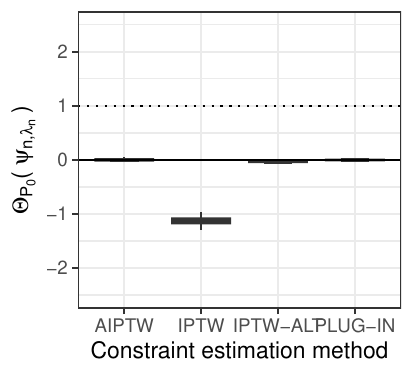}
		\caption{Constraint value when propensity score $\pi_0$ inconsistently estimated ($n = 1600$)}
	\end{subfigure}
	\hfill
	\begin{subfigure}[b]{0.45\textwidth}
		\centering
		\includegraphics[width=\textwidth]{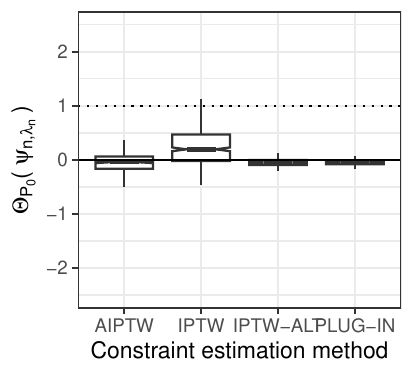}
		\caption{Constraint value when mediator distribution $\gamma_0$ inconsistently estimated ($n = 1600$).}
	\end{subfigure}
	
	\caption{Comparison of the ability of estimators to control constraint when constructed using robust and non-robust estimators of $\Theta_{P_0}$. The true value $\Theta_{P_0}(\psi_0)$ is shown by the dotted line. The target constraint value is 0, as indicated by the solid horizontal line.}
	\label{fig:constraint_misspec_sim_n1600}
\end{figure}

Inconsistent nuisance estimation also generally led to poor control of the constraint, with particularly poor control for estimators built using IPTW (Figures \ref{fig:constraint_misspec_sim_n200} and \ref{fig:constraint_misspec_sim_n1600}). In fact, estimators based on IPTW generally had poor performance even when all nuisance parameters were consistently estimated. Interestingly, the plug-in method yielded tight control of the constraint in the setting when $\psi_0$ was inconsistently estimated. This is again in line with our theory that predicts that whenever the estimate of $\Theta_{P_0}$ is consistent for $\Theta_{P_0}(\tilde{\psi})$ (where $\tilde{\psi}$ is the in-probability limit of $\psi_n$), then it should still be possible to control the constraint in spite of misspecification. For the NDE, the PLUG-IN estimator would indeed be consistent for this target. On the other hand, the putatively robust AIPTW estimator demonstrated poor control of the constraint in this setting. This can be explained by the fact that its multiple robustness property ensure that the estimator is consistent for $\Theta_{P_0}(\psi_0)$ in spite of the misspecification of $\psi_0$. However, in this situation $\Theta_{P_0}(\psi_0) \ne \Theta_{P_0}(\tilde{\psi})$ owing to misspecification of $\psi_0$. In all other scenarios, AIPTW, IPTW-ALT and PLUG-IN generally had similar performance. We leave to future work characterizing alternative robust estimators that can be leveraged to produce more robust constrained learning estimates.

\subsection{Data generating mechanism for high-dimensional simulation}

For the simulation setting described in Section~6.3, we simulated $W$ as a $p$-dimensional vector of standard Normal covariates for $p= 10, 50, 100$. We set $\pi_0(1 \mid W) = \mbox{expit}(W_1 - W_2/2 + W_3/3 - W_4/4 + W_5/5)$, $\gamma_0(1 \mid X, W) = \mbox{expit}(-X + W_1 - W_2/2 + W_3/3 - 4 * W_4/4 + W_5/5)$. The outcome $Y$ was generated from a conditional Normal distribution with mean given $(X,M,W)$ of $X + M + W_1 - W_2/2 + W_3/3 - W_4/4 + W_5/5$ and variance equal to 4. In this setting, the true value of the ATE is about 0.813 and the true value of the NDE is 1.

Each relevant functional parameter was estimated using LASSO where 10-fold cross-validation was used to select the LASSO penalization parameter that minimized mean squared error (the default behavior in standard LASSO software packages). 

\subsection{Details for highly adaptive LASSO simulation}

In this simulation, we simulated a univariate covariate $W$ from a Unif($-\pi, \pi$) distribution. We set $\pi_0(1 \mid W) = \mbox{expit}\{\mbox{sin}(\beta W)\}$. We considered only the ATE constraint and so did not simulate a mediator. The outcome $Y$ was generated from a conditional Normal distribution with mean given $(X,W)$ given by $X + X\mbox{sin}(\beta W) + (1-X)\mbox{cos}(\beta W)$ and unit variance. Note that $\beta$ controls the variation norm of $\pi_0(1 \mid W)$ and of $\psi_0$, with larger values of $\beta$ implying larger variation in the underlying function. We ran simulations setting $\beta = 1, 10$ to compare performance under low and high variation norms respectively. 

\begin{figure}
	\centering
	\includegraphics[scale=1]{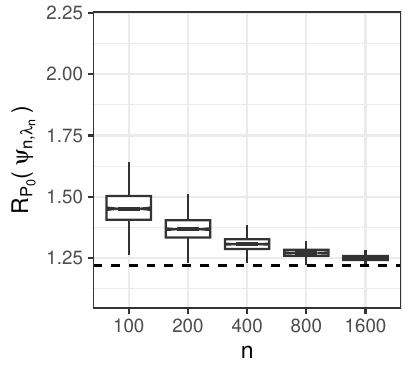}
	\includegraphics[scale=1]{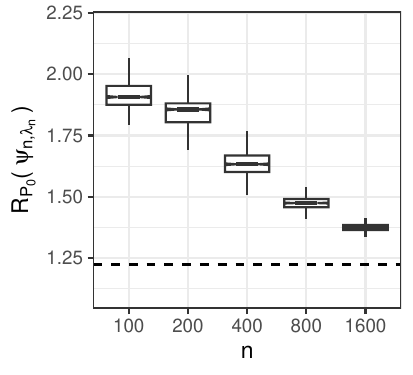}
	\vspace{-0.35cm}
	\caption{\textbf{Average treatment effect constraint using highly adaptive LASSO with varying variation norm.} Distributions of risk (mean squared error) of $\psi_{n,\lambda_n}$ over 1000 realizations for each sample size for low variation norm (left, $\beta = 1$) and high variation norm (right, $\beta = 10$). The dashed line indicates the optimal risk $R_{P_0}(\psi_{0,\lambda_0})$.}
	\label{fig:glmnet_sim_risks}
\end{figure}

\begin{figure}
	\centering
	\includegraphics[scale=1]{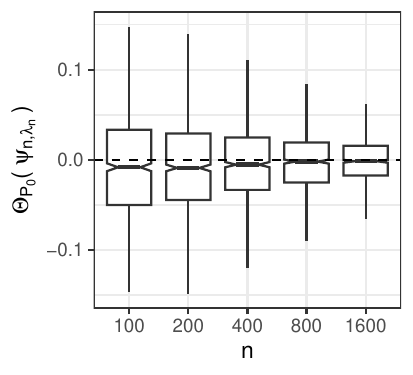}
	\includegraphics[scale=1]{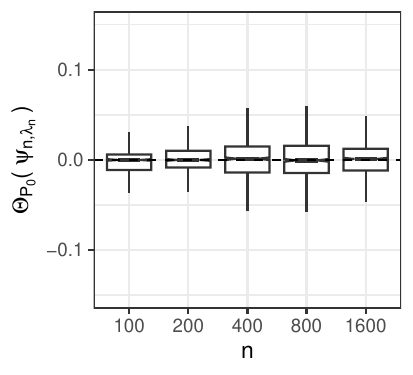}
	\vspace{-0.35cm}
	\caption{\textbf{Average treatment effect constraint using highly adaptive LASSO with varying variation norm.}  Distributions of the constraint $\Theta_{P_0}(\psi_{n,\lambda_n})$ over 1000 realizations for for low variation norm (left, $\beta = 1$) and high variation norm (right, $\beta = 10$). The dashed line indicates the desired constraint value of zero. The constraint value under the unconstrained $\psi_0$, $\Theta_{P_0}(\psi_0) = 1$ and is not shown due to the scale of the figure.}
	\label{fig:glmnet_sim_constr}
\end{figure}

\subsection{Adult-based simulation study}

We designed a simulation study based on the characteristics of our analysis of the Adult data set \citep{adult_2}. We generated $W_1$ (simulated race) by sampling the four race categories with probability equal to the empirical probability in our analysis data set: $P(W_1 = 1) = 0.01; P(W_1 = 2) = 0.03; P(W_1 = 3) = 0.09; P(W_1 = 4) = 0.87$. Similarly, we simulated country of origin by sampling according to empirical distributions. While the original data contained over 40 countries of origin, for simplicity we reduced the number to 11. The probability $P(W_2 = w_2) = 0.01$ for $w_2 = 1, \dots, 10$ and $P(W_2 = 11) = 0.9$, reflecting a similar distribution heavily skewed towards US as the country of origin, as was observed in the original data. Age ($W_3)$ was simulated from a Gamma(7.5, 0.2) distribution, which gave reasonably close approximations to moments and quantiles of the observed age distribution. Sex ($X$) was drawn from a Bernoulli distribution with $\pi_0(1 \mid W) = \mbox{expit}(0.933 + 0.031  I(W_1 = 2) - 0.462  I(W_1 = 3) + 0.316  I(W_1 = 4) - 1.146  I(W_2 = 2) - 0.469  I(W_2 = 3) - 1.223  I(W_2 = 4) - 1.282  I(W_2 = 5) - 1.578  I(W_2 = 6) - 1.161  I(W_2 = 7) - 1.004  I(W_2 = 8) - 1.152  I(W_2 = 9) - 1.124  I(W_2 = 10) - 0.958  I(W_2 = 11) + 0.014  W_3)$. Coefficients for this model were selected based on a fitted logistic regression in the analysis data. 

For simplicity, we considered a constructing a binary mediator $M$ to reflect mediators of sex/income disparities. To do this, we fit a logistic regression in the analysis data set of the binary outcome onto $W, M, X$ and created a mediator score given by the linear predictor of only the mediator variables. Individuals with a score of greater than the median were considered to have ``high mediator score.'' We then regressed this binary high mediator score onto $X$ and $W$ to generate realistic coefficients. This allowed us to simulate a single binary $M$ from a Bernoulli distribution with $\gamma_0(1 \mid X, W) = \mbox{expit}(-1.451 + 0.141  X + 0.844  I(W_1 = 2) - 0.078  I(W_1 = 3) + 0.455  I(W_1 = 4) + 0.786  I(W_2 = 2) + 0.501  I(W_2 = 3) - 0.216  I(W_2 = 4) + 0.164  I(W_2 = 5) - 0.79  I(W_2 = 6) - 0.348  I(W_2 = 7) - 1.067  I(W_2 = 8) + 1.152  I(W_2 = 9) + 1.601  I(W_2 = 10) + 0.407  I(W_2 = 39) + 0.013  W_3)$. Finally, the income outcome was simulated from a Bernoulli distribution with coefficients selected based on a fit to the true data, such that $\psi_0(X, M, W) = \mbox{expit}(-4.897 + 0.706  I(W_1 = 2) + 0.119  I(W_1 = 3) + 0.679  I(W_1 = 4) - 0.079  I(W_2 = 2) - 0.454  I(W_2 = 3) - 2.312  I(W_2 = 4) - 0.552  I(W_2 = 5) - 1.41  I(W_2 = 6) - 0.875  I(W_2 = 7) - 1.213  I(W_2 = 8) + 0.017  I(W_2 = 9) + 0.015  I(W_2 = 10) - 0.405  I(W_2 = 39) + 0.042  W3 + 1.247  X + 1.651  M)$.

We simulated 1000 data sets of size 10000 and compared the performance of our estimator to that of a constrained maximum likelihood estimator \citep{nabi2018fair} in terms of risk and constraint control. As with previous simulation studies, risk and constraint value were evaluated on an independent simulated test data set of size one million. 

\begin{figure}
	\centering
	\includegraphics[width=0.8\linewidth]{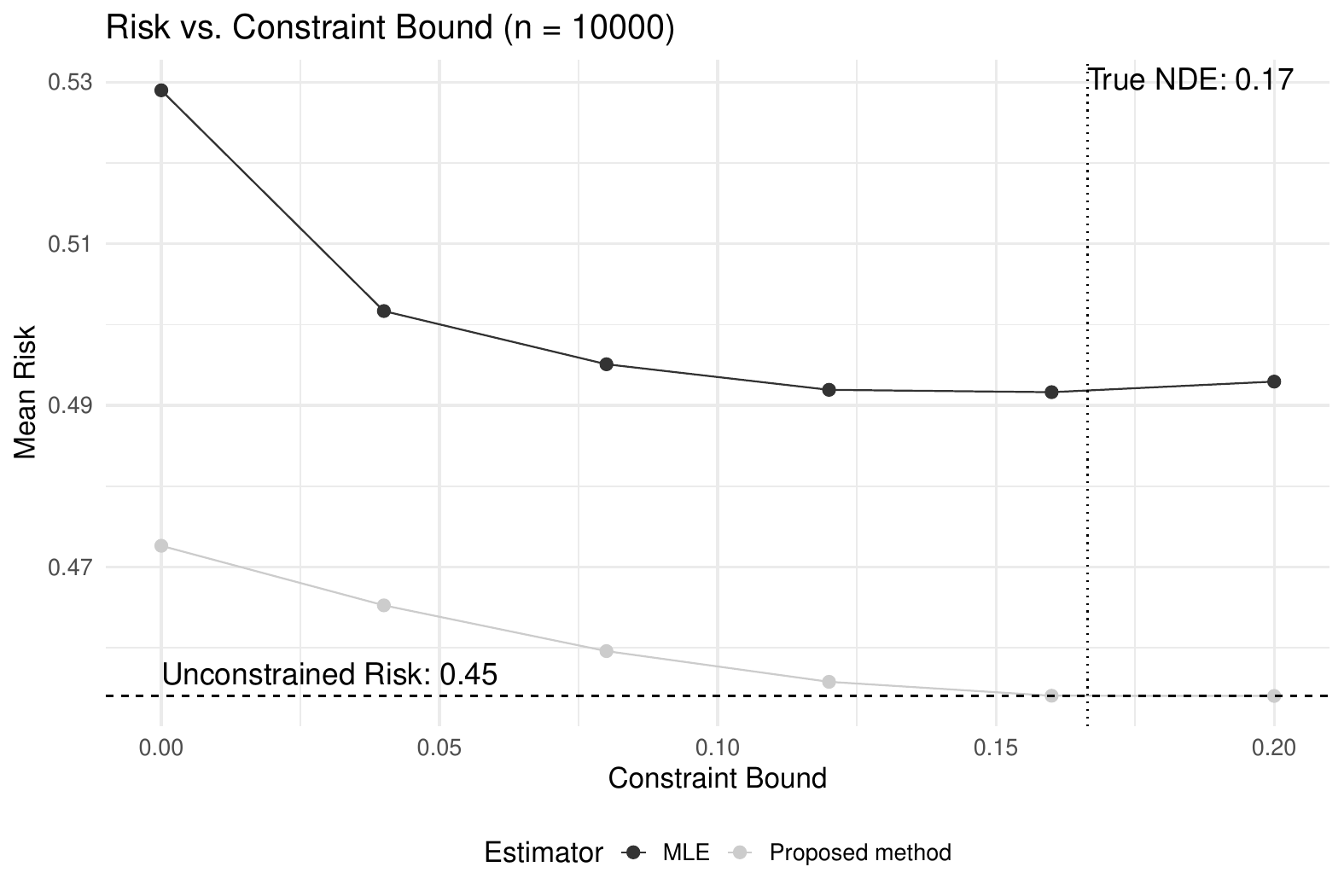}
	\includegraphics[width=0.8\linewidth]{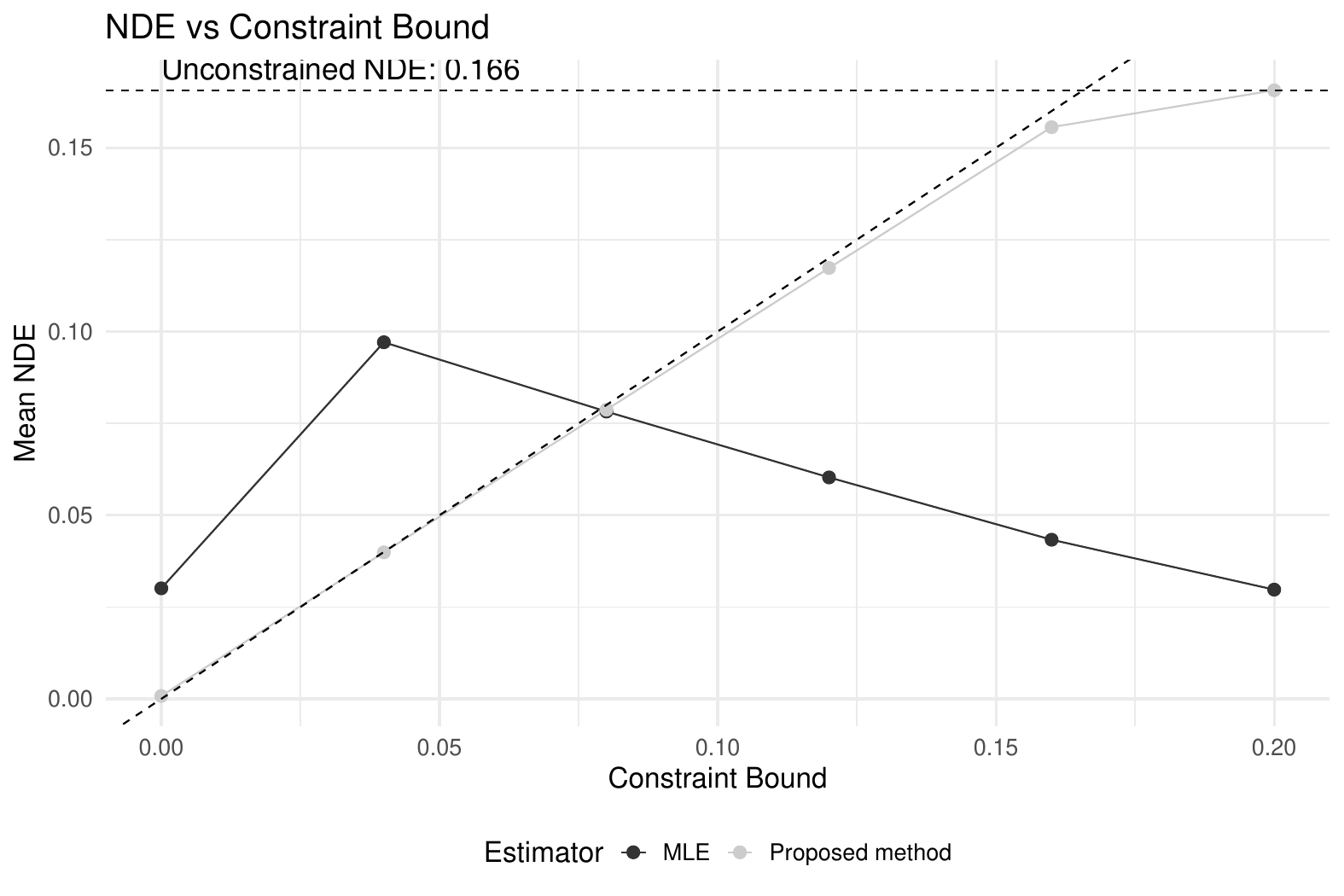}
	\caption{Comparison of risk and constraint control in the Adult data-inspired simulation for a constrained maximum likelihood estimator (black) versus the proposed method (gray).}
	\label{fig:adult-sim-results}
\end{figure}

We found that the proposed method yielded substantially lower estimated risk and substantially more appropriate control of the constraint at the bound (Figure \ref{fig:adult-sim-results}, which is in line with our other simulation results and our analysis of the Adult data.

\clearpage
\section{Additional real-data application}
\label{app:real_data}


We applied our methods to the COMPAS dataset, a risk‑assessment tool used in the US criminal justice system to inform pretrial detention decisions. The data we used in our study has been made accessible by Propublica and is detailed in the study by \cite{propublica}. Our primary focus is on developing a predictive model for two-year recidivism (denoted by $Y$) using variables such as the defendant's age ($W_1$), gender ($W_2$), race ($X$), past convictions ($M$), while ensuring that race has no direct effect on the recidivism prediction. This analysis is restricted to a comparison between African-American ($X=1$) and Caucasian ($X=0$) individuals. We acknowledge the contentious nature of recidivism predictions and clarify that our analysis leverages this dataset as a widely recognized benchmark in the field. 

\begin{figure}
	\centering
	\includegraphics[scale=0.23]{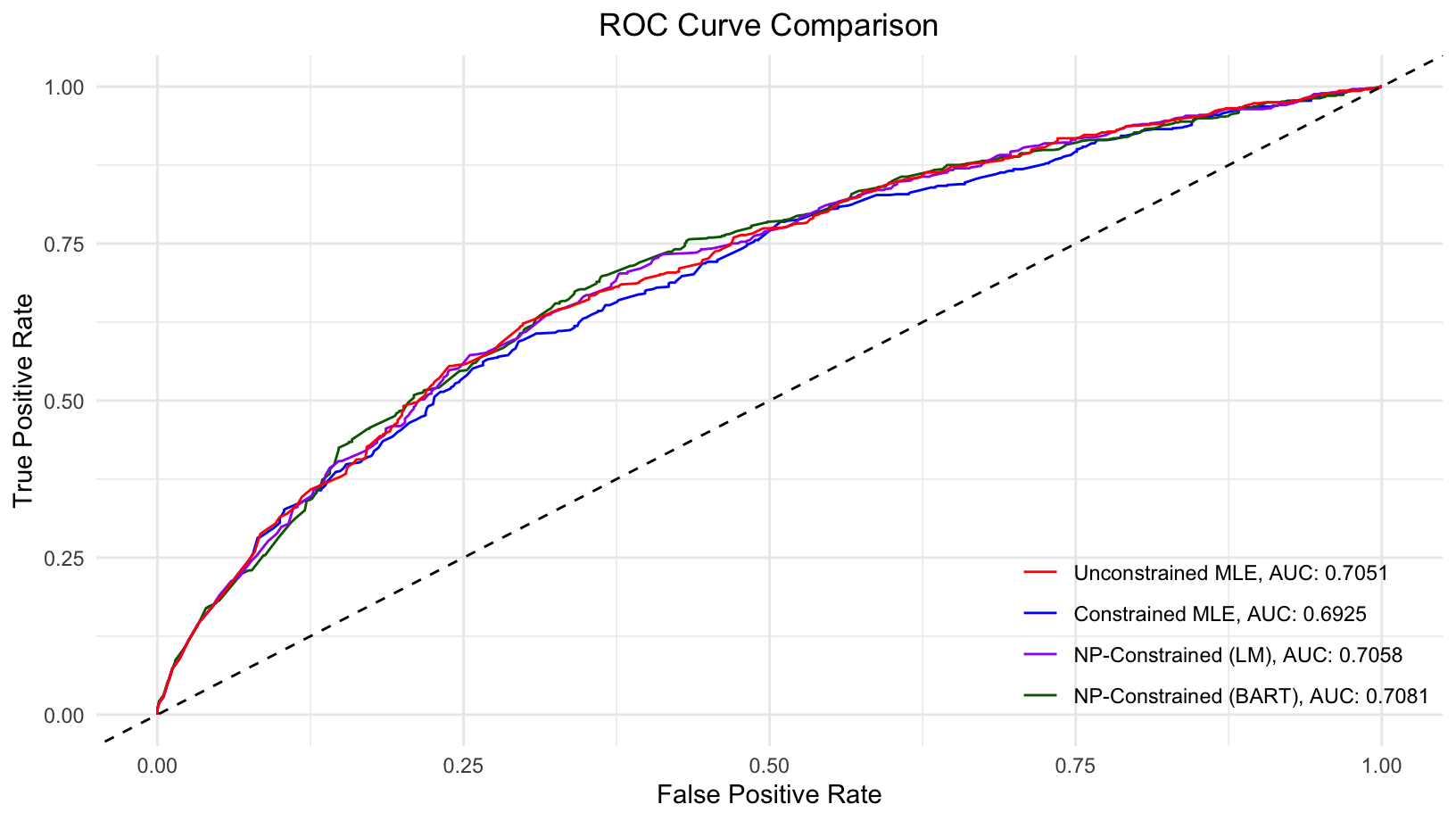}
	\caption{ROC curve comparisons for model performance evaluation. }
	\label{fig:compas_roc}
\end{figure}

We allocated the sample into training and test datasets with a $70\%$ to $30\%$ split. We calculated the direct effect of race on recidivism using the plug-in estimate formula\\ $\frac{1}{n} \sum_{i = 0}^n 
\sum_{m=0}^1 \{\psi_n(1,m,W_i) - \psi_n(0, m, W_i) \} \gamma_n(m \mid 0, W_i)$. Generalized linear models (GLMs) were used to estimate the nuisance functions $\gamma_n$ and $\psi_n$, finding the effect to be $0.044 \pm 0.002$, across $100$ splits. Our approach aimed to eliminate this direct effect, focusing on minimizing the cross-entropy risk. 
We used both GLMs and Bayesian additive regression trees (BART) for estimates of the nuisance functional parameters in our methodology, comparing them to constrained maximum likelihood (MLE) where the direct effect is constrained to $|\Theta_{n}(\psi_n)| \leq 0.001$. The ROC curve comparisons, shown in Figure~\ref{fig:compas_roc}, demonstrate that our approaches, whether using GLM or BART for nuisance parameter estimation, achieve higher AUC compared to the constrained MLE approach. 
Additionally, using the constrained MLE, the direct effect in the test data is estimated to be $0.0013 \pm 0.002$ across 100 splits, while our approach yield an absolute zero for the direct effect. 

\clearpage


\begingroup
\renewcommand{\baselinestretch}{0.92}
\selectfont  
\setlength{\bibsep}{9pt}    
\bibliographystyle{abbrvnat}
\bibliography{refs}
\endgroup

\end{document}